\documentclass{bmvc2k}
\usepackage{subfig}
\usepackage{footnote}
\makesavenoteenv{tabular}
\makesavenoteenv{table}

\title{Light-Weight RefineNet for\\ Real-Time Semantic Segmentation}

\addauthor{Vladimir Nekrasov}{vladimir.nekrasov@adelaide.edu.au}{1}
\addauthor{Chunhua Shen}{chunhua.shen@adelaide.edu.au}{1}
\addauthor{Ian Reid}{ian.reid@adelaide.edu.au}{1}

\addinstitution{
School of Computer Science,\\
The University of Adelaide,\\
Australia
}

\runninghead{Nekrasov \bmvaEtAl}{Light-Weight RefineNet for real-time sem. segm.}

\def\etal{\emph{et al}\bmvaOneDot}

\begin{document}

\makeatletter
\@addtoreset{section}{part}
\makeatother

\renewcommand{\partname}{}

\maketitle
\vspace{-0.25in}
\begin{abstract}
We consider an important task of effective and efficient semantic image segmentation. In particular, we adapt a powerful semantic segmentation architecture, called \textit{RefineNet}~\cite{LinMSR17}, into the more compact one, suitable even for tasks requiring real-time performance on high-resolution inputs. To this end, we identify computationally expensive blocks in the original setup, and propose two modifications aimed to decrease the number of parameters and floating point operations. By doing that, we achieve more than twofold model reduction, while keeping the performance levels almost intact. Our fastest model undergoes a significant speed-up boost from $20$ FPS to $55$ FPS on a generic GPU card on $512\times512$ inputs with solid $81.1\%$ mean iou performance on the test set of PASCAL VOC~\cite{EveringhamGWWZ10}, while our slowest model with $32$ FPS (from original $17$ FPS) shows $82.7\%$ mean iou on the same dataset. Alternatively, we showcase that our approach is easily mixable with light-weight classification networks: we attain $79.2\%$ mean iou on PASCAL VOC using a model that contains only $3.3$M parameters and performs only $9.3$B floating point operations.
\end{abstract}
\vspace{-0.25in}
\section{Introduction}
\label{sec:intro}
The number of empirical successes of deep learning keeps steadily increasing, and new technological and architectural breakthroughs become available practically every month. In particular, deep learning has become the default choice in most areas of computer vision, natural language processing, robotics, audio processing~\cite{KrizhevskySH12,SermanetEZMFL13,GirshickDDM14,SimonyanZ14a,SzegedyLJSRAEVR14,ZeilerF14,Girshick15,KarpathyL15,LillicrapHPHETS15,RenHG015,VinyalsTBE15,XuBKCCSZB15,OordDZSVGKSK16,GuHLL17}. Nevertheless, these breakthroughs often come at a price of expensive computational requirements, which hinders their direct applicability in tasks requiring real-time processing. The good part of this story is that it has been empirically shown by multiple researchers that oftentimes there are lots of redundant operations~\cite{DenilSDRF13,BaC14,DentonZBLF14,JaderbergVZ14,HanPTD15,HintonVD15,UrbanGKAWCMPR16}, and this redundancy can be (and should be) exploited in order to gain speed benefits while keeping performance mostly intact. The bad part of the story is that there are not (yet) general approaches on how to exploit such redundancies in most cases. For example, such methods as knowledge distillation~\cite{BaC14,HintonVD15,BucilaCN06,RomeroBKCGB14} and pruning~\cite{HanPTD15,CunDS89,HassibiS92,HanMD15} require us to have an access to an already pre-trained large model in order to train a smaller model of the same (or near) performance. Obviously, there might be times when this is not feasible. In contrast, designing a novel architecture for a specific scenario of high-resolution inputs \cite{PaszkeCKC16,ZhaoQSSJ17} limits the applicability of the same architecture on datasets with drastically different properties and often requires expensive training from scratch.
	
	Here we are restricting ourselves to the task of semantic segmentation, which has proven to be pivotal on the way of solving scene understanding, and has been successfully exploited in multiple real-world applications, such as medical image segmentation~\cite{CiresanGGS12,RonnebergerFB15}, road scene understanding~\cite{BadrinarayananH15,XuGYD17}, aerial segmentation~\cite{KlucknerMRB09,MnihH10}. These applications tend to rely on real-time processing with high-resolution inputs, which is the Achilles' heel of most modern semantic segmentation networks.
	
	Motivated by the aforementioned observations we aim to tackle the task of real-time semantic segmentation in a different way. In particular, we build our approach upon \textit{RefineNet}~\cite{LinMSR17}, a powerful semantic segmentation architecture that can be seamlessly used with any backbone network, such as ResNet~\cite{HeZRS16}, DenseNet~\cite{HuangLMW17}, NASNet~\cite{ZophVSL17}, or any other. This architecture belongs to the family of the `encoder-decoder' segmentation networks~\cite{PaszkeCKC16,BadrinarayananH15,NohHH15}, where the input image is first progressively downsampled, and later progressively upsampled in order to recover the original input size. We have chosen this particular architecture as it i) shows the best performance among the `encoder-decoder' approaches, and ii) does not operate over large feature maps in the last layers as methods exploiting atrous convolution do~\cite{ChenPKMY14,ChenPK0Y16,ChenPSA17,ZhaoSQWJ17,YuKF17}. The processing of large feature maps significantly hinders real-time performance of such architectures as DeepLab~\cite{ChenPKMY14,ChenPK0Y16,ChenPSA17} and PSPNet~\cite{ZhaoSQWJ17} due to the increase in the number of floating point operations. 
	As a weak point, the `encoder-decoder' approaches tend to have a larger number of parameters due to the decoder part that recovers the high resolution output. In this work, we tackle the real-time performance problem by specifically concentrating on the decoder and empirically showing that both the number of parameters and floating point operations can be drastically reduced without a significant drop in accuracy.
	
	In particular, i) we outline important engineering ameliorations that save the number of parameters by more than $50\%$ in the original RefineNet; then ii) we pinpoint the redundancy of residual blocks in the architecture and show that the performance stays intact with those blocks being removed. We conduct extensive experiments on three competitive benchmarks for semantic segmentation - PASCAL VOC~\cite{EveringhamGWWZ10}, NYUDv2~\cite{SilbermanHKF12} and PASCAL Person-Part~\cite{ChenMLFUY14,ChenYWXY16}, and with five different backbone networks - ResNet-50, ResNet-101, ResNet-152~\cite{HeZRS16}, along with recently released NASNet-Mobile~\cite{ZophVSL17}, and MobileNet-v2~\cite{abs-1801-04381}. The first three backbones are used for the direct comparison between our approach and the original RefineNet, while the last two are used to showcase that our method is orthogonal to the backbone compression, and that we can further benefit from combining these methods.
	
Quantitatively, our fastest ResNet model achieves $55$ FPS on inputs of size $512\times512$ on GTX 1080Ti, while having $81.1\%$ mean intersection over union (mean iou) performance on PASCAL VOC, $41.7\%$ mean iou on NYUDv2, and $64.9\%$ mean iou on Person-Part. In contrast, our best performing model with $82.7\%$ mean iou on VOC, $44.4\%$ mean iou on NYUDv2, and $67.6\%$ mean iou on Person-Part, still maintains solid $32$ FPS.\par
	All the models will be made publicly available to facilitate the usage of effective and efficient semantic segmentation in a multitude of applications.
\vspace{-0.12in}

\section{Related work}
\vskip -0.1in
\textbf{Semantic segmentation.} Early approaches in semantic segmentation relied on handcrafted features, such as HOG~\cite{DalalT05} and SIFT~\cite{Lowe04}, in combination with plain classifiers~\cite{ShottonWRC06,CsurkaP08,ShottonJC08,FulkersonVS09} and hierarchical graphical models~\cite{LadickyRKT09,PlathTN09,KrahenbuhlK11}. This lasted until the resurgence of deep learning, and a pivotal work by Long \etal~\cite{LongSD15}, in which the authors converted a deep image classification network into a fully convolutional one, able to operate on inputs of different sizes. Further, this line of work was extended by using dilated convolutions and probabilistic graphical models~\cite{ChenPKMY14,YuKF17,LinSRH15,LiuLLLT15,ZhengJRVSDHT15}. Other works concentrated around the encoder-decoder paradigm~\cite{LinMSR17,PaszkeCKC16,NohHH15}, where the image is first progressively downsampled, and then progressively upsampled in order to generate the segmentation mask of the same size as the input. Most recently, Lin \etal~\cite{LinMSR17}, proposed RefineNet that extended the encoder-decoder approach by adding residual units inside the skip-connections between encoder and decoder. Zhao \etal~\cite{ZhaoSQWJ17} appended a multi-scale pooling layer to improve the information flow between local and global (contextual) information; Fu \etal~\cite{abs-1708-04943} exploited multiple encoder-decoder blocks arranged in an hour glass manner~\cite{YangLZ17} and achieved competitive results using DenseNet~\cite{HuangLMW17}. The current state-of-the art on the popular benchmark dataset PASCAL VOC~\cite{EveringhamGWWZ10} belongs to DeepLab-v3~\cite{ChenPSA17} ($86.9\%$ mean intersection-over-union on the test set) and DeepLab-v3+~\cite{abs-1802-02611} ($89.0\%$) based on ResNet~\cite{HeZRS16} and Xception~\cite{Chollet17}, correspondingly, where the authors included batch normalisation~\cite{IoffeS15} layers inside the atrous spatial pyramid pooling block (ASPP)~\cite{ChenPK0Y16}, made use of a large multi-label dataset~\cite{SunSSG17}, and added decoder in the latest version.\\
\\
\textbf{Real-Time Segmentation.} Most methods outlined above either suffer from a large number of parameters~\cite{LinMSR17}, a large number of floating point operations~\cite{ChenPKMY14,ChenPK0Y16,ChenPSA17,ZhaoSQWJ17}, or both~\cite{PaszkeCKC16,NohHH15,LongSD15,YangLZ17}. These issues constitute a significant drawback of exploiting such models in applications requiring real-time processing.

To alleviate them, there have been several task-specific approaches, e.g., ICNet~\cite{ZhaoQSSJ17}, where the authors adapt PSPNet~\cite{ZhaoSQWJ17} to deal with multiple image scales progressively. 
They attain the speed of $30$ FPS on $1024\times2048$ images, and $67\%$ mean iou on the validation set of CityScapes~\cite{CordtsORREBFRS16}, but it is not clear whether this approach would still acquire solid performance on other datasets with low-resolution inputs. 
Alternatively, Li \etal~\cite{LiLLLT17} propose a cascading approach where in each progressive level of the cascade only a certain portion of pixels is being processed by explicitly setting a hard threshold on intermediate classifier outputs. They demonstrate performance of $14.3$ FPS on $512\times512$ input resolution with $78.2\%$ mean iou on PASCAL VOC~\cite{EveringhamGWWZ10}. We also note that several other real-time segmentation networks have been following the encoder-decoder paradigm, but have not been able to acquire decent performance levels. Concretely, SegNet~\cite{BadrinarayananH15} achieved $40$ FPS on inputs of size $360\times480$ with only $57.0\%$ mean iou on CityScapes, while ENet~\cite{PaszkeCKC16} were able to perform inference of $20$ FPS on inputs of size $1920\times1080$ with $58.3\%$ mean iou on CityScapes.\\
\\
\textbf{Real-Time Inference in Other Domains.} Multiple task-agnostic solutions have been proposed to speed-up inference in neural networks by compression. Current methods achieve more than $100\times$ compression rates~\cite{IandolaMAHDK16}. Among them, most popular approaches include quantisation~\cite{GongLYB14,ZhouNZWWZ16,ZhouYGXC17}, pruning~\cite{HanPTD15,CunDS89,HassibiS92,HanMD15} and knowledge distillation~\cite{BaC14,HintonVD15,BucilaCN06,RomeroBKCGB14}.
Other notable examples are centered around the idea of low-rank factorisation and decomposition~\cite{DentonZBLF14,JaderbergVZ14}, where a large layer can be decomposed into smaller ones by exploiting (linear) structure. 
	
While all the methods above have proven to be effective in achieving significant compression rates, an initial powerful, but large pre-trained model for the task at hand must be acquired first. We further note that these methods can be applicable as a post-processing step on top of our approach, but we leave this direction for later exploration.

Finally, most recently, a multitude of new, light-weight architectures~\cite{IandolaMAHDK16,HowardZCKWWAA17} has been proposed, and it has empirically been shown that models with a smaller number of parameters are able to attain solid performance. To highlight that our approach is orthogonal to such advances in the design of classification networks, we conduct experiments with two light-weight architectures, namely, NASNet-Mobile~\cite{ZophVSL17}, and MobileNet-v2~\cite{abs-1801-04381}, and show competitive results with very few parameters and floating point operations.
\vspace{-0.12in}

\section{Light-Weight RefineNet}
\label{sec:blind}



\begin{figure*}
	\begin{center}
	\includegraphics[height=11.5cm, trim=70 220 80 10,clip]{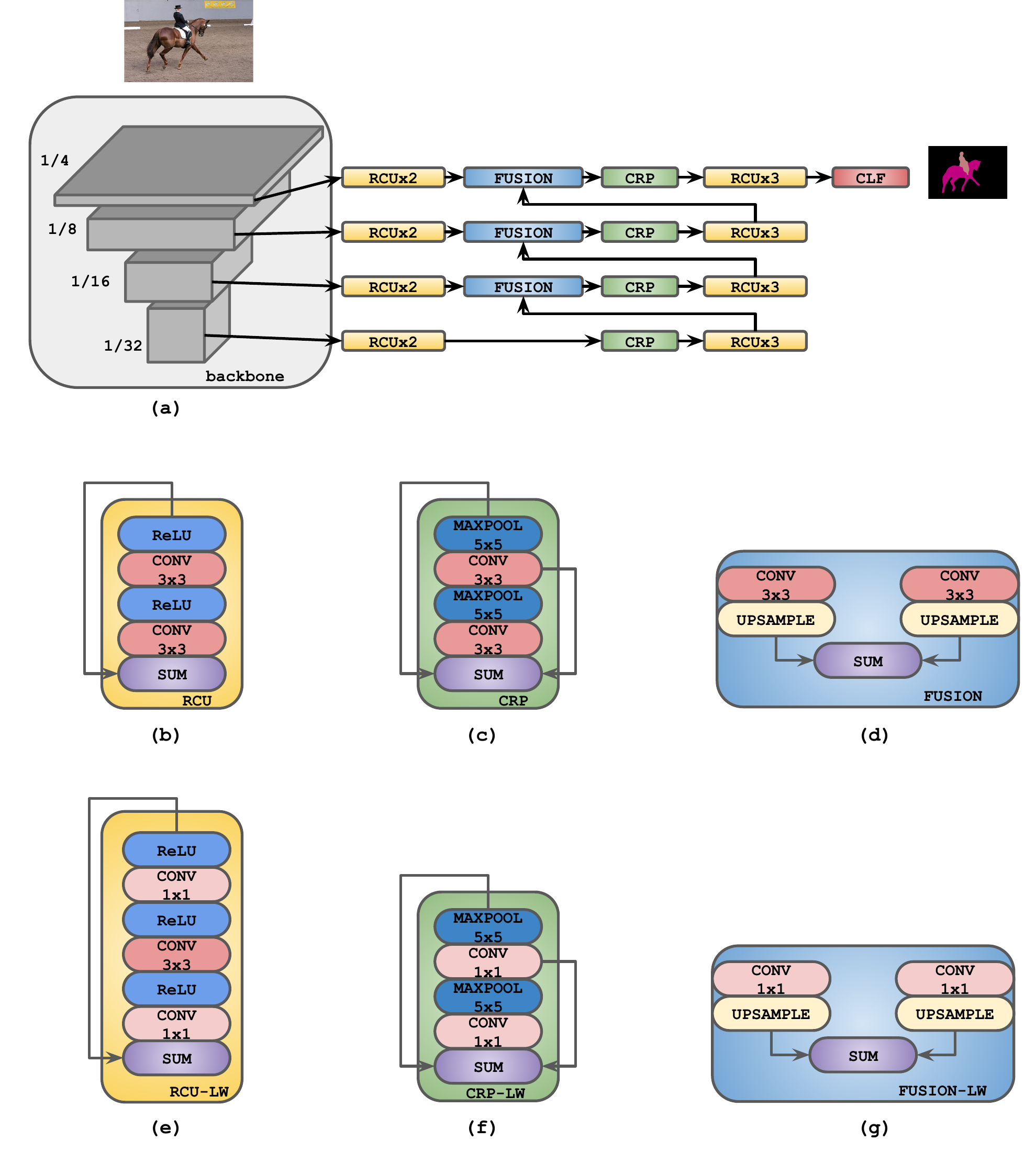}
	\end{center}
	\vskip -0.23in
	\caption{RefineNet structure. (a) General network architecture with RefineNet for semantic segmentation, where \textit{CLF} stands for a single 3x3 convolutional layer with the number of channels being equal to the number of output classes; (b)-(d) general outline of original RCU, CRP and fusion blocks; (e)-(g) light-weight RCU, CRP and fusion blocks. In the interests of brevity, we only visualise 2 convolutional layers for the CRP blocks (instead of 4 used in the original architecture). Note that we do not use any RCU blocks in our final architecture as discussed in Sec.~\ref{rcu_red}.}
	\label{fig:arch}
	\vskip -0.2in
\end{figure*}

Here we will outline our approach aimed to decrease the number of parameters and floating point operations of the original RefineNet architecture, while keeping the performance levels the same.\par
We start by describing the original setup, and then continue with the explanation and discussion of the changes that we propose to be made.

\subsection{RefineNet Primer}

As mentioned above, the RefineNet architecture belongs to the family of the encoder-decoder approaches. As the encoder backbone, it can re-use most popular classification networks; in particular, the original contribution was showcased using residual networks~\cite{HeZRS16}. The approach does not amend an underlying classification network in any way except for omitting last classification layers.

In the decoder part, RefineNet relies on two types of additional abstractions: residual convolutional unit (RCU)~(Fig.~\ref{fig:arch}(b)) and chained residual pooling (CRP)~(Fig.~\ref{fig:arch}(c)). The first one is a simplification of the original residual block~\cite{HeZRS16} without batch normalisation~\cite{IoffeS15}, while the second one is a sequence of multiple convolutional and pooling layers, also arranged in a residual manner. All of these blocks use $3\times3$ convolutions and $5\times5$ pooling with appropriate padding so that the spatial size would stay intact.

The decoding process starts with the propagation of the last output from the classifier (with the lowest resolution) through two RCU blocks followed by four pooling blocks of CRP and another three RCU blocks before being fed into a fusion block~(Fig.~\ref{fig:arch}(d)) along with the second to last feature map. Inside the fusion block, each path is convolved with $3\times3$ convolution and upsampled to the largest resolution among the paths. Two paths are then summed up, and analagoulosly further propagated through several RCU, CRP and fusion blocks until the desired resolution is reached. The final convolutional layer produces the score map.

\subsection{Replacing 3x3 convolutions}
We start by highlighting that the most expensive parts in terms of both the number of parameters and the number of floating point operations of the original RefineNet stem from the ubiquitous usage of $3\times3$ convolutions. Thus, we focus on replacing them with simpler counterparts without performance drops.

Intuitively, convolutions with larger kernel sizes aim to increase the receptive field size (and the global context coverage), while $1\times1$ convolutions are merely transforming per-pixel features from one space to another locally. We argue that in the case of RefineNet we do not require expensive $3\times3$ convolutions at all, and we are able to show empirically that replacing them with $1\times1$ convolutions does not hurt performance. Furthermore, we evaluate the empirical receptive field size~\cite{ZhouKLOT14} for both cases, and do not find any significant difference (Section~\ref{ss:erf}). 
In particular, we replace $3\times3$ convolutions within CRP and fusion blocks with $1\times1$ counterpart~(Fig.~\ref{fig:arch}(f-g)), and we amend RCU into the one with the bottleneck design~\cite{HeZRS16}~(Fig.~\ref{fig:arch}(e)). This way,  we reduce the number of parameters by more than $2\times$ times, and  the number of FLOPs by more than $3\times$ times (Table~\ref{table:r101}).
	
Alternatively, we might have reverted to recently proposed depthwise separable convolutions~\cite{Chollet17} that are proven to save lots of parameters without sacrificing performance, but our approach is even simpler as it does not use any $3\times3$ convolutions at all, except for the last classification layer.

\begin{table}
	\begin{center}
		\begin{tabular}{l|c|c}
			\hline
			Model & Parameters,M & FLOPs,B\\
			\hline
			RefineNet-101~\cite{LinMSR17} & $118$ & $263$\\
			\textbf{RefineNet-101-LW-WITH-RCU}  & $54$ & $76$\\
			\textbf{RefineNet-101-LW}  & $\textbf{46}$ & $\textbf{52}$\\
			\hline
		\end{tabular}
	\end{center}
	\caption{Comparison of the number of parameters and floating point operations on $512\times512$ inputs between original RefineNet, Light-Weight RefineNet with RCU (\textbf{LW-WITH-RCU}), and Light-Weight RefineNet without RCU (\textbf{LW}). All the networks exploit ResNet-101 as backbone.\label{table:r101}}
	\vskip -0.2in
\end{table}

\subsection{Omitting RCU blocks}
\label{rcu_red}
We proceeded with initial experiments and trained the light weight architecture outlined in the previous section using PASCAL VOC~\cite{EveringhamGWWZ10}. We were able to achieve close to the original network performance, and, furthermore, observed that removing RCU blocks did not lead to any accuracy deterioration, and, in fact, the weights in RCU blocks almost completely saturated.

To confirm that this only occurs in the light weight case, we explored dropping RCU blocks in the original RefineNet architecture, and we experienced more than $5\%$ performance drop on the same dataset. We argue that this happens due to the RCU blocks being redundant in the $1\times1$ convolution regime, as the only important goal of increasing the contextual coverage is essentially performed by pooling layers inside CRP. To back up this claim, we conduct a series of ablation experiments which are covered later in Section~\ref{ss:abl}.

Our final architecture does not contain any RCU blocks and only relies on CRP blocks with $1\times1$ convolutions and $5\times5$ max-pooling inside, which makes our method extremely fast and light-weight.
    
\subsection{Adaptation to different backbones}
\label{ss:ada}
A significant practical benefit of the RefineNet architecture is that it can be mixed with any backbone network as long as the backbone has several subsampling operations inside (which is the case for most SOTA classifiers). Thus, there are no specific changes needed to be made in order to apply the RefineNet architecture using any other model. In particular, in the experiments section, we showcase its adaptation using efficient NASNet-Mobile~\cite{ZophVSL17} and MobileNet-v2~\cite{abs-1801-04381} networks; we still achieve solid performance with the limited number of parameters and floating point operations.
	
	
	
\section{Experiments}
In the experimental part, our aims are to prove empirically that we are able i) to achieve similar performance levels with the original RefineNet while ii) significantly reducing the number of parameters and iii) drastically increasing the speed of a forward pass; and iv) to highlight the possibility of applying our method using other architectures.
	
To this end, we consider three segmentation datasets, namely, NYUDv2~\cite{SilbermanHKF12}, PASCAL VOC~\cite{EveringhamGWWZ10} and PASCAL Person-Part dataset~\cite{ChenMLFUY14,ChenYWXY16}, and five classification networks, i.e., ResNet-50, ResNet-101, ResNet-152~\cite{HeZRS16}, NASNet-Mobile~\cite{ZophVSL17} and MobileNet-v2~\cite{abs-1801-04381} (only for Pascal VOC), all of which have been pre-trained on ImageNet~\cite{DengDSLL009}. As a general practice, we report mean intersection over union~\cite{EveringhamGWWZ10} on each benchmark. Additional results on PASCAL Context~\cite{MottaghiCLCLFUY14} and CityScapes~\cite{CordtsORREBFRS16} are given in the supplementary material.
	
We perform all our experiments in PyTorch~\cite{paszke2017automatic}, and train using stochastic gradient descent with momentum. For all residual networks we start with the initial learning rate of $5$e-$4$, for NASNet-mobile and MobileNet-v2 we start with the learning rate of $1$e-$3$. We keep batch norm statistics frozen during the training.
	
For benchmarking, we use a workstation with $8$GB RAM, Intel i5-7600 processor, and one GT1080Ti GPU card under CUDA 9.0 and CuDNN 7.0. For fair comparison of runtime, we re-implement original RefineNet in PyTorch. We compute $100$ forward passes with random inputs, and average the results; when reporting, we provide both the mean and standard deviation values.\\
\\
\textbf{NYUDv2.} We first conduct a series of initial experiments on NYUDv2 dataset~\cite{SilbermanHKF12,GuptaAM13}. This dataset comprises $1449$ RGB-D images with $40$ segmented class labels, of which $795$ are used for training and $654$ for testing, respectively. We do not make use of depth information in any way. We reduce the learning rate by half after $100$ and $200$ epochs, and keep training until $300$ epochs, or until earlier convergence.

\begin{table}
\begin{center}
\resizebox{\textwidth}{!}{\begin{tabular}{l|c|c|c|c}
\hline
Model & NYUD mIoU,\% & Person mIoU,\% & Params,M & Runtime,ms\\
\hline
FCN16-s RGB-HHA~\cite{LongSD15} & 34.0 & - & - & -\\
Context~\cite{LinSRH15} & 40.6 & - & - & -\\
DeepLab-v2-CRF~\cite{ChenPK0Y16} & - & $64.9$~(\emph{msc}) & \textbf{44} & - \\
RefineNet-50~\cite{LinMSR17}  & $42.5$ & $65.7$ & 99 & $54.18 \pm 0.46$\\
RefineNet-101~\cite{LinMSR17}  & $43.6$ & $67.6$ & 118 & $60.25 \pm 0.53$\\
RefineNet-152~\cite{LinMSR17} & \textbf{46.5}~(\emph{msc}) & \textbf{68.8}~(\emph{msc}) & 134 & $69.37 \pm 0.78$\\
\hline
\textbf{RefineNet-LW-50} (ours) & $41.7$ & $64.9$ & \textbf{27} & $\textbf{19.56} \pm 0.29$\\
\textbf{RefineNet-LW-101} (ours) & $43.6$ & $66.7$ & \textbf{46} & $\textbf{27.16} \pm 0.19$\\
\textbf{RefineNet-LW-152} (ours)  & $44.4$ & $67.6$ & \textbf{62} & $\textbf{35.82} \pm 0.23$\\
\hline
\end{tabular}}
\end{center}
\caption{Quantitative results on the test sets of NYUDv2 and PASCAL Person-Part. Mean iou, the number of parameters and the runtime (mean$\pm$std) of one forward pass on $625\times468$ inputs are reported, where possible. Multi-scale evaluation is defined as \emph{msc}.\label{table:nyu-person}}
\vskip -0.15in
\end{table}

Our quantitative results are provided in Table~\ref{table:nyu-person} along with the results from the original RefineNet, and other competitive methods on this dataset. We note that for all ResNet networks we are able to closely match the performance of the original RefineNet, while having only a slight portion of the original parameters. This further leads to a significant twofold speedup.\\
\\
\textbf{Person-Part.} PASCAL Person-Part dataset~\cite{ChenMLFUY14,ChenYWXY16} consists of $1716$ training and $1817$ validation images with $6$ semantic classes, including head, torso, and upper/lower legs/arms, plus background. We follow the same training strategy as for NYUDv2, and provide our results in Table~\ref{table:nyu-person}. Again, we achieve analagous to the original models results.\\
\\
\textbf{PASCAL VOC.} We continue our experiments with a standard benchmark dataset for semantic segmentation, PASCAL VOC~\cite{EveringhamGWWZ10}. This dataset consists of $4369$ images with $20$ semantic classes plus background, of which $1464$ constitute the training set, $1449$ - validation, and $1456$ - test, respectively. As commonly done, we augment it with additionally annotated VOC images from BSD~\cite{HariharanABMM11}, as well as images from MS COCO~\cite{LinMBHPRDZ14}.
	
We train all the networks using the same strategy, but with different learning rates as outlined above: in particular, we reduce the learning rate by half after $20$ epochs of COCO training and after $50$ epochs of the BSD training, and keep training until $200$ total epochs, or until the accuracy on the validation set stops improving.

\begin{figure}[t]
\centering
\resizebox{\textwidth}{!}{\begin{tabular}{ccc|ccc|cc}
        \subfloat{\includegraphics[width = 0.13\linewidth]{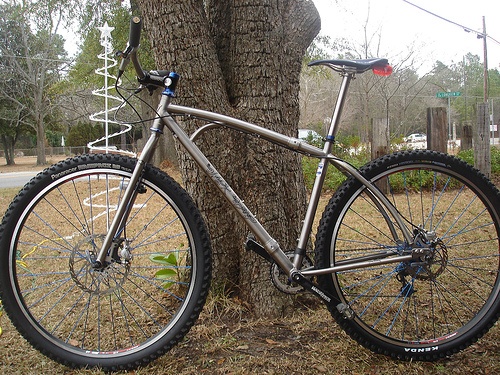}} &
        \subfloat{\includegraphics[width = 0.13\linewidth]{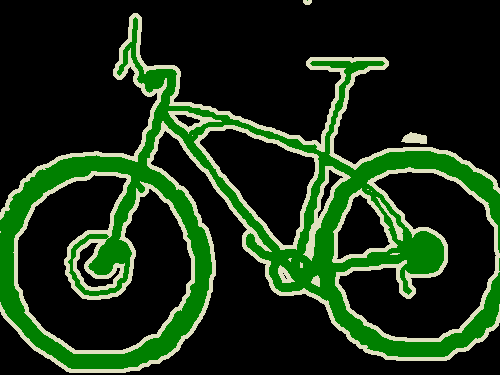}} &
        \subfloat{\includegraphics[width = 0.13\linewidth]{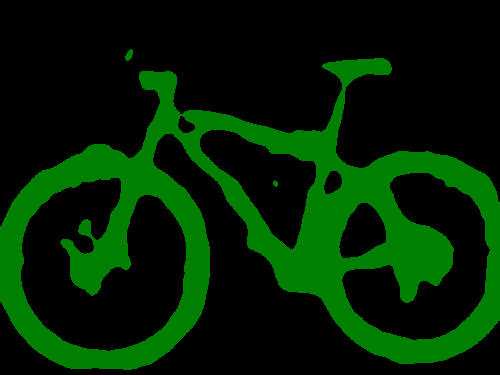}} &
        \subfloat{\includegraphics[width = 0.13\linewidth]{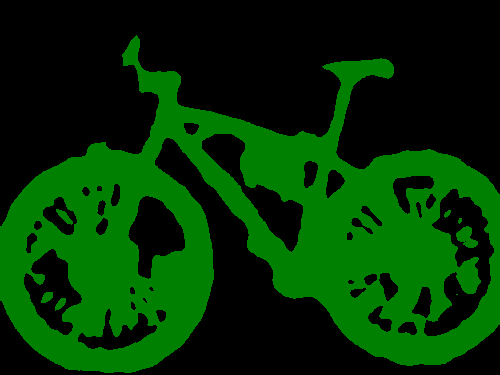}} &
        \subfloat{\includegraphics[width = 0.13\linewidth]{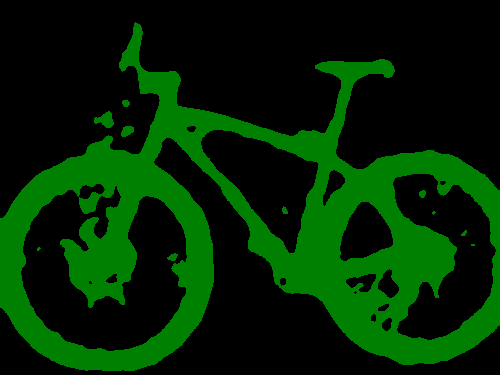}} &
        \subfloat{\includegraphics[width = 0.13\linewidth]{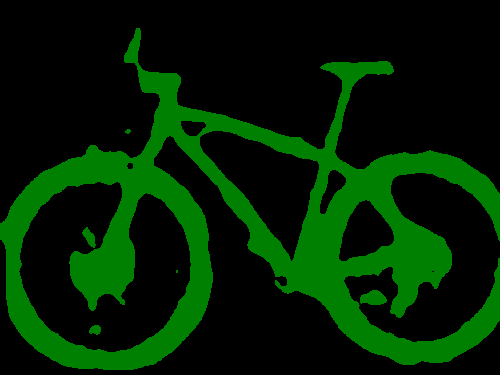}} &
        \subfloat{\includegraphics[width = 0.13\linewidth]{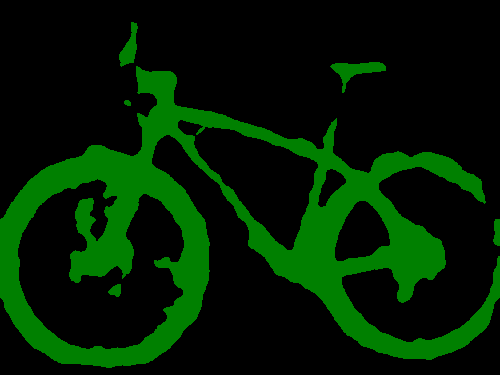}} &
        \subfloat{\includegraphics[width = 0.13\linewidth]{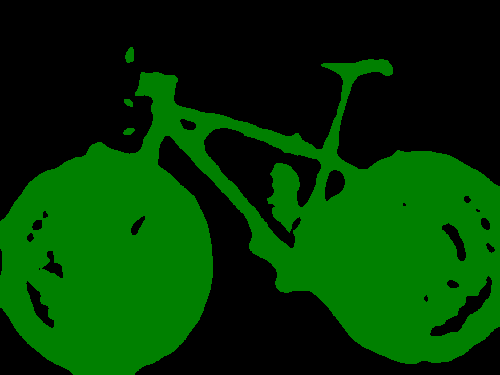}}\\[-0.15in]
        \subfloat{\includegraphics[width = 0.13\linewidth]{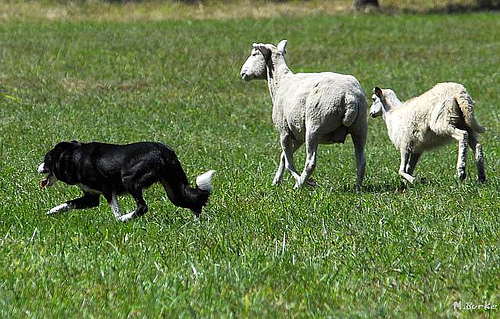}} &
        \subfloat{\includegraphics[width = 0.13\linewidth]{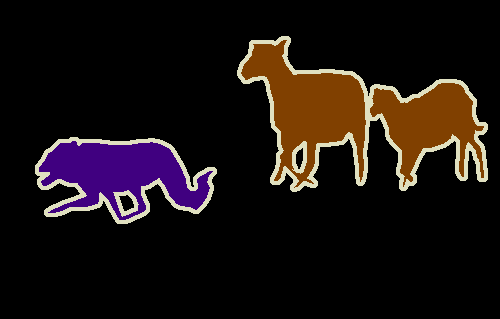}} &
        \subfloat{\includegraphics[width = 0.13\linewidth]{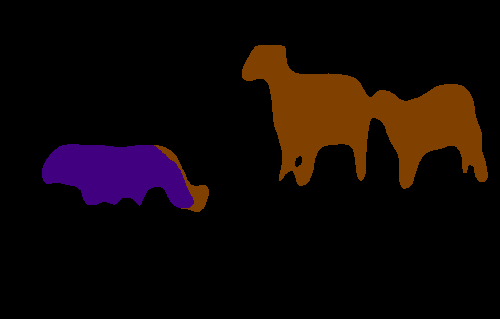}} &
        \subfloat{\includegraphics[width = 0.13\linewidth]{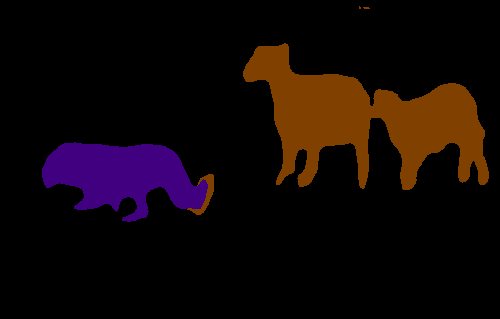}} &
        \subfloat{\includegraphics[width = 0.13\linewidth]{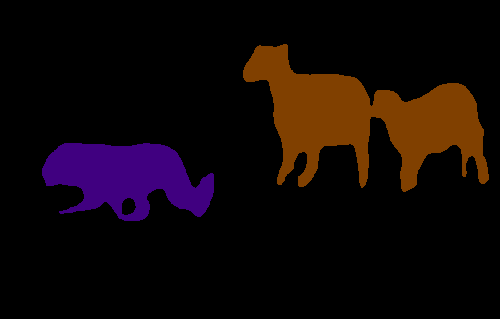}} &
        \subfloat{\includegraphics[width = 0.13\linewidth]{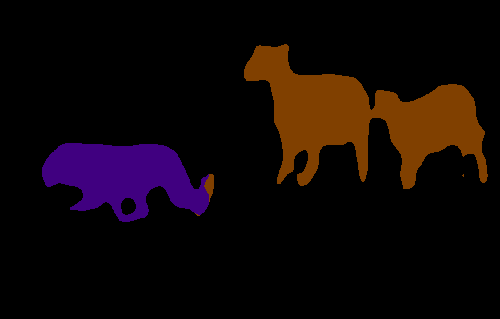}} &
        \subfloat{\includegraphics[width = 0.13\linewidth]{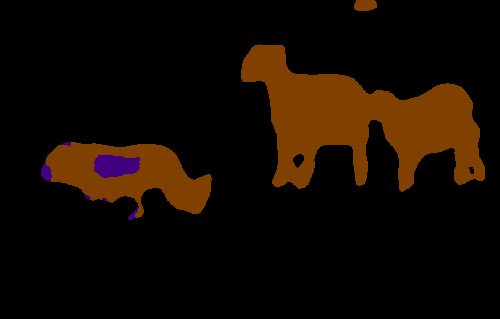}} &
        \subfloat{\includegraphics[width = 0.13\linewidth]{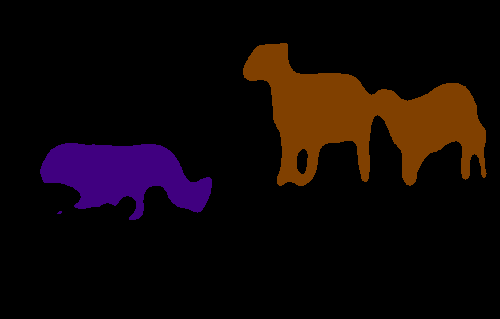}}\\[-0.15in]
        \subfloat{\includegraphics[width = 0.13\linewidth]{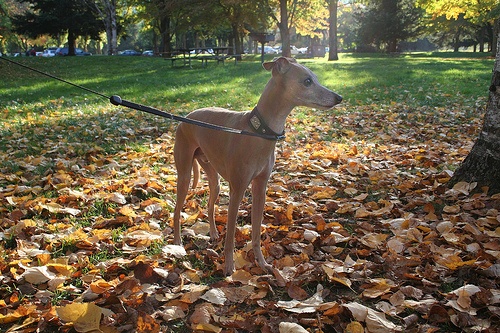}} &
        \subfloat{\includegraphics[width = 0.13\linewidth]{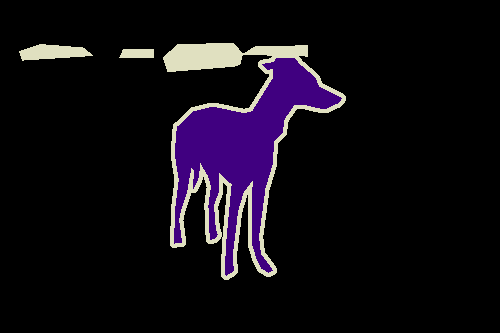}} &
        \subfloat{\includegraphics[width = 0.13\linewidth]{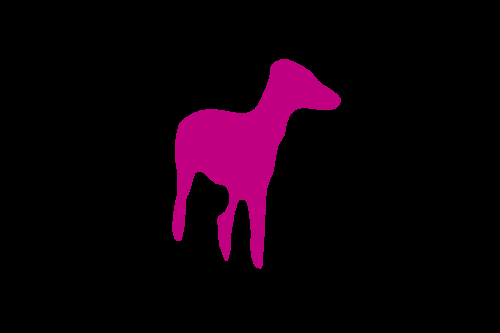}} &
        \subfloat{\includegraphics[width = 0.13\linewidth]{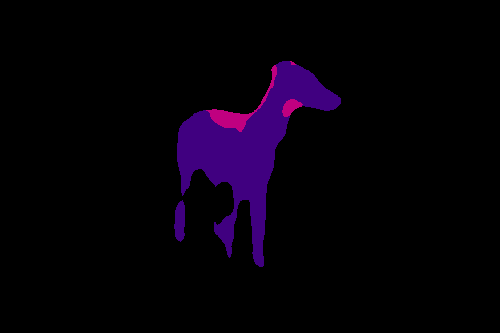}} &
        \subfloat{\includegraphics[width = 0.13\linewidth]{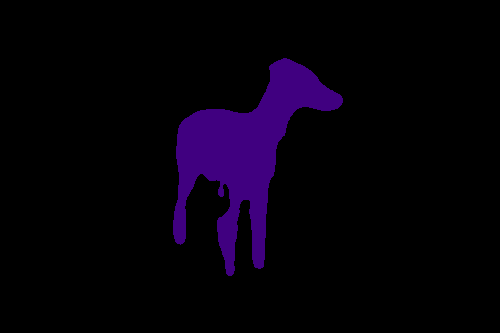}} &
        \subfloat{\includegraphics[width = 0.13\linewidth]{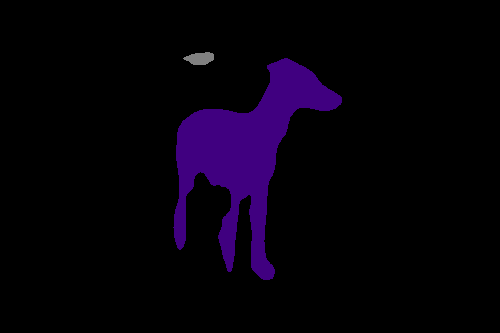}} &
        \subfloat{\includegraphics[width = 0.13\linewidth]{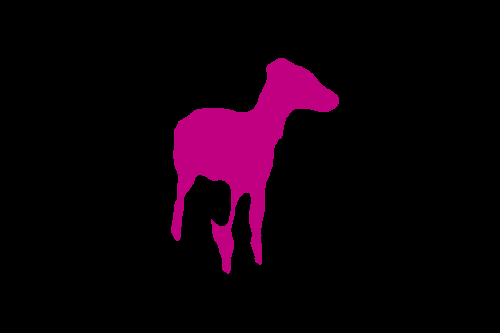}} &
        \subfloat{\includegraphics[width = 0.13\linewidth]{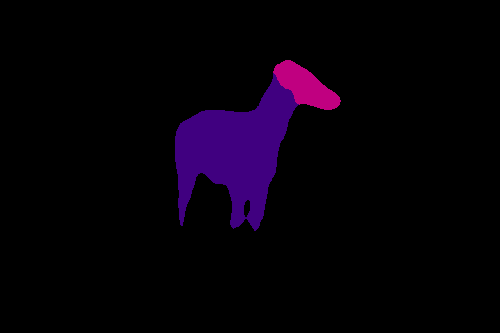}}\\[-0.15in]
        \subfloat{\includegraphics[width = 0.13\linewidth]{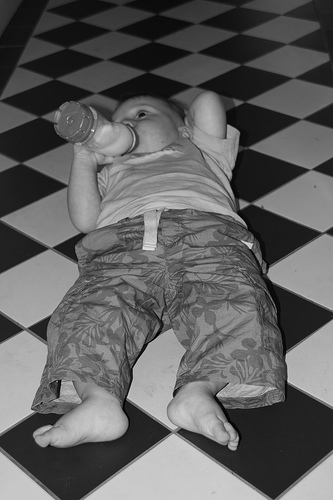}} &
        \subfloat{\includegraphics[width = 0.13\linewidth]{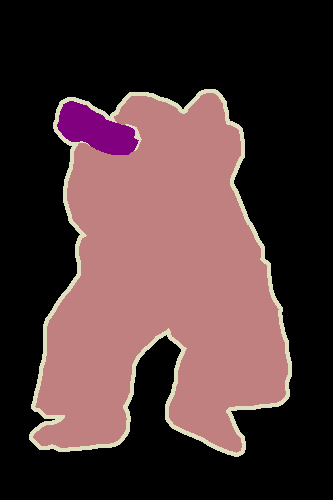}} &
        \subfloat{\includegraphics[width = 0.13\linewidth]{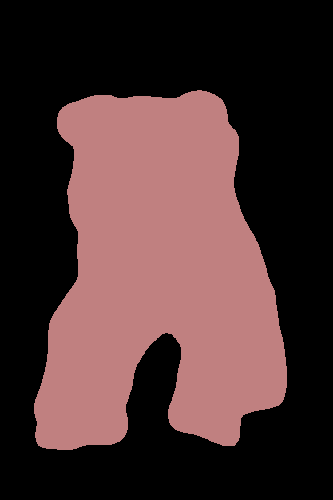}} &
        \subfloat{\includegraphics[width = 0.13\linewidth]{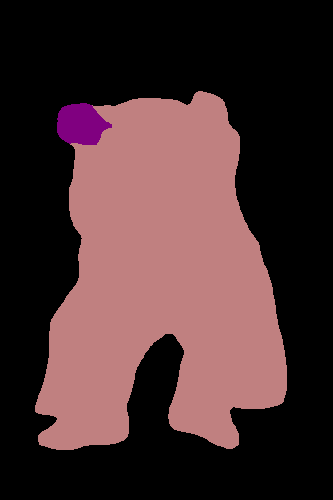}} &
        \subfloat{\includegraphics[width = 0.13\linewidth]{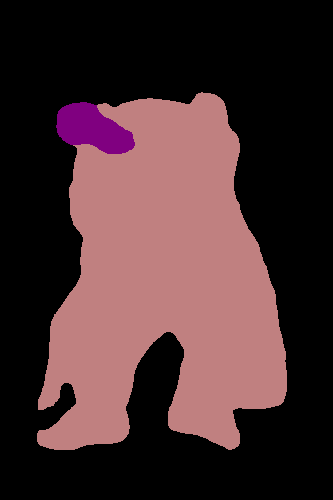}} &
        \subfloat{\includegraphics[width = 0.13\linewidth]{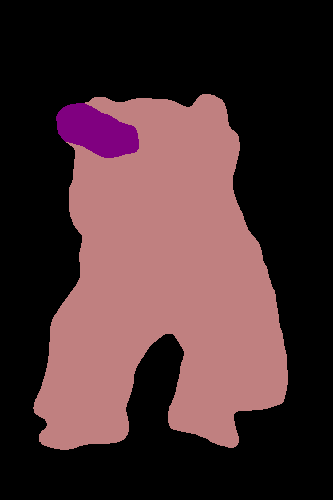}} &
        \subfloat{\includegraphics[width = 0.13\linewidth]{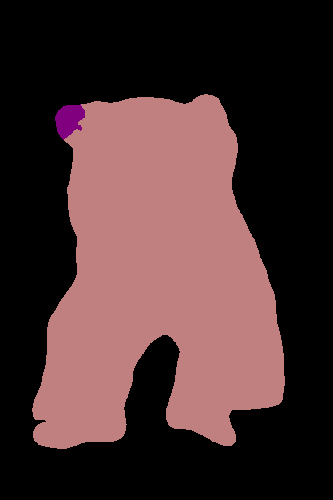}} &
        \subfloat{\includegraphics[width = 0.13\linewidth]{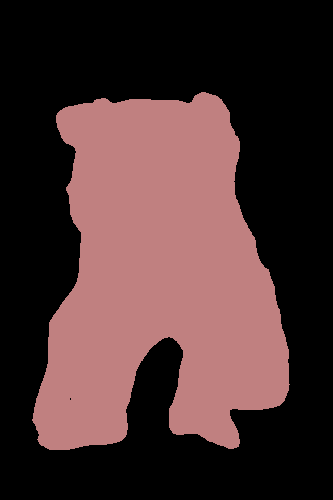}}\\
        Image&GT&RF-101&RF-50-LW&RF-101-LW&RF-152-LW&MOB-LW&NAS-LW
\end{tabular}}
\vskip 0.1in
\caption{Visual results on validation set of PASCAL VOC with residual models (RF), MobileNet-v2 (MOB) and NASNet-Mobile (NAS). The original RefineNet-101 (RF-101) is our re-implementation.}
\vskip -0.2in
\label{fig:voc-res}
\end{figure}

Our quantitative results on validation and test sets are given in Table~\ref{table:voc} along with the results from the original RefineNet. We again notice that for all ResNet networks, we achieve performance on-par with the original RefineNet; for NASNet-Mobile and MobileNet-v2, we outperform both MobileNet-v1+DeepLab-v3 and MobileNet-v2+DeepLab-v3 approaches~\cite{abs-1801-04381}, which are closely related to our method. Qualitative results are provided on Figure~\ref{fig:voc-res}.

\begin{table}
	\begin{center}
		\begin{tabular}{l|c|c|c}
			\hline
			Model & val mIoU,\% & test mIoU,\% & FLOPS,B\\
			\hline
			DeepLab-v2-ResNet-101-CRF~\cite{ChenPK0Y16} & $77.7$ & $79.7$ & -\\
			RefineNet-101~\cite{LinMSR17} & - & $82.4$ & $263$\\
			RefineNet-152~\cite{LinMSR17} & -  & \textbf{83.4} & $283$\\
			\textbf{RefineNet-LW-50} (ours) & $78.5$ & $81.1$\footnote{http://host.robots.ox.ac.uk:8080/anonymous/ZAC8JH.html} & \textbf{33}\\
			\textbf{RefineNet-LW-101} (ours) & $80.3$ & $82.0$\footnote{http://host.robots.ox.ac.uk:8080/anonymous/RBAJXC.html} & $52$\\
			\textbf{RefineNet-LW-152} (ours) & $82.1$ & $82.7$\footnote{http://host.robots.ox.ac.uk:8080/anonymous/EFA17T.html} & $71$\\
			\hline
			MobileNet-v1-DeepLab-v3~\cite{abs-1801-04381}  & $75.3$ & - & $14.2$\\
			MobileNet-v2-DeepLab-v3~\cite{abs-1801-04381}  & $75.7$ & - & \textbf{5.8}\\
			\textbf{RefineNet-LW-MobileNet-v2} (ours) & \textbf{76.2} & \textbf{79.2}\footnote{http://host.robots.ox.ac.uk:8080/anonymous/90TQVN.html} & \textbf{9.3}\\
			\textbf{RefineNet-LW-NASNet-Mobile} (ours) & \textbf{77.4} & \textbf{79.3}\footnote{http://host.robots.ox.ac.uk:8080/anonymous/WRYPBD.html} & 11.4\\
			\hline
		\end{tabular}
	\end{center}
	\caption{Quantitative results on PASCAL VOC. Mean iou and the number of FLOPs on $512\times512$ inputs are reported, where possible.\label{table:voc}}
	\vskip -0.15in
\end{table}

\section{Discussion}
As noted above, we are able to achieve similar results by omitting more than half of the original parameters. This raises the question on how and why this does happen. We conduct a series of experiments aimed to provide some insights into this matter. More details are provided in the supplementary material.

\subsection{Receptive field size}
\label{ss:erf}
First, we consider the issue of the receptive field size. Intuitively, dropping $3\times3$ convolutions should significantly harm the receptive field size of the original architecture. Nevertheless, we note that we do not experience this due to i) the skip-design structure of RefineNet, where low-level features are being summed up with the high-level ones, and ii) keeping CRP blocks that are responsible for gathering contextual information.  

\begin{figure}
\centering
\includegraphics[height=6.5cm]{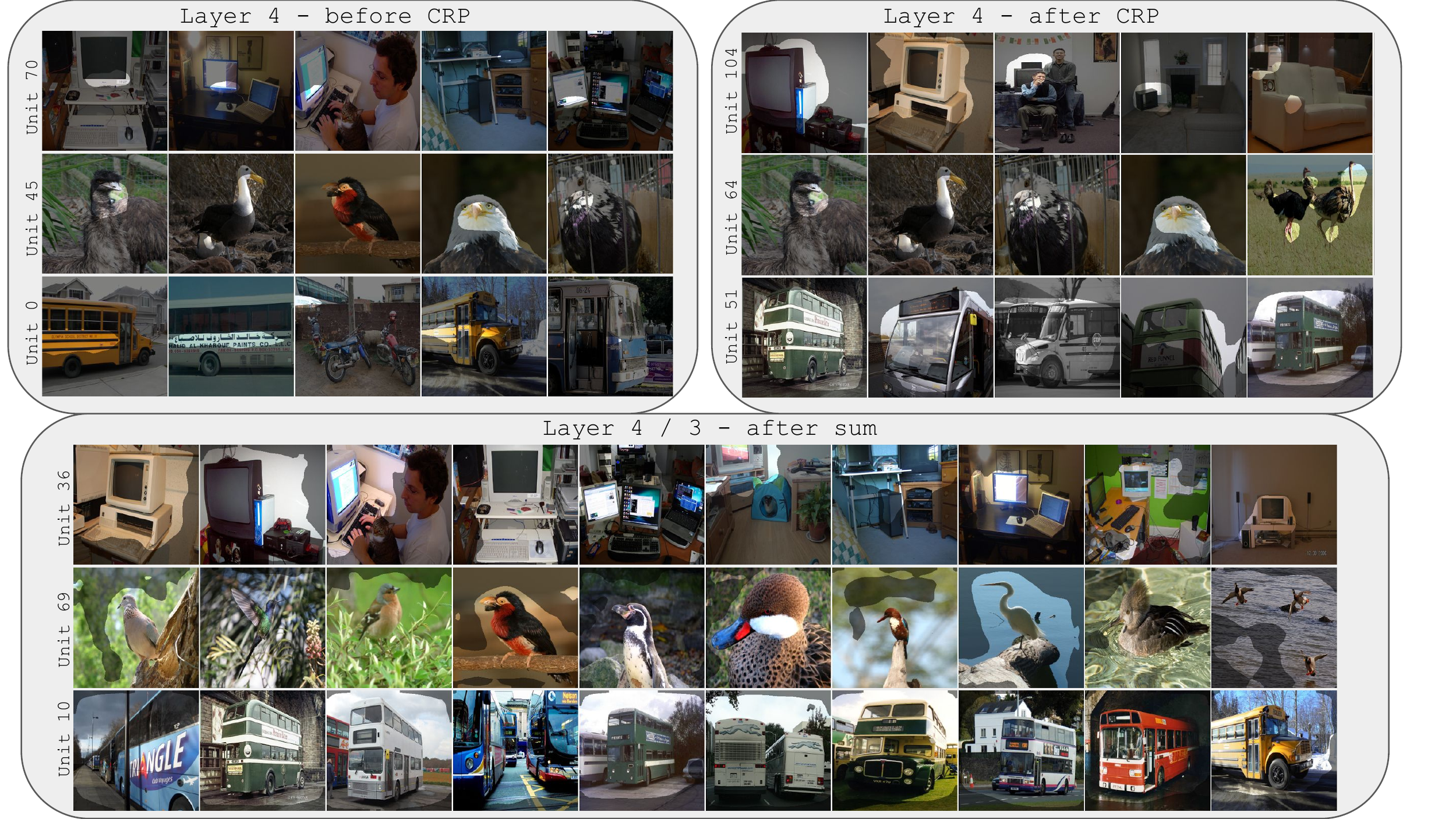}
\caption{Comparison of empirical receptive field before CRP (top left), after CRP (top right), and after the first summation of levels $4$ and $3$ (bottom) in RefineNet-LW-101 pre-trained on PASCAL VOC. Top activated images along with top activated regions are shown.}
\label{fig:erf1}
\end{figure}

In particular, we explore the empirical receptive field (ERF) size~\cite{ZhouKLOT14} in Light-Weight RefineNet-101 pre-trained on PASCAL VOC. Before CRP ERF is concentrated around small object parts, and barely covers the objects~(Fig.~\ref{fig:erf1}). The CRP block tends to enlarge ERF, while the summation with the lower layer features further produces significantly larger activation contours.
		
\subsection{Representational power}
\label{ss:abl}
Even though the RCU block seems not to influence the receptive field size of the network, it might still be responsible for producing important features for both segmentation and classification.

To evaluate whether it is the case indeed, we conduct ablation experiments by adding multi-label classification and segmentation heads in the lowest resolution block of RefineNet-101 pre-trained on PASCAL VOC 1) before RCU, 2) after RCU, and 3) after CRP. We fix the rest of the trained network and fine-tune only these heads separately using $1464$ training images of the VOC dataset. We evaluate all the results on the validation subset of VOC.

As seen from Table~\ref{table:crprcu}, CRP is the main driving force behind accurate segmentation and classification, while the RCU block improves the results just marginally. 

\begin{table}
	\begin{center}
		\begin{tabular}{l|c|c}
			\hline
			Model & mIoU,\% & acc,\%\\
			\hline
			Before RCU  & $54.63$ & $95.39$\\
			After RCU  & $55.16$ & $95.42$\\
			After CRP  & $\textbf{57.83}$ & $\textbf{97.71}$\\
			\hline
		\end{tabular}
	\end{center}
	\caption{Ablation experiments comparing CRP and RCU. Multi-label accuracy (without background) and mean IoU are reported as measured on the validation set of PASCAL VOC.\label{table:crprcu}}
\vskip -0.15in
\end{table}

\section{Conclusions}

In this work, we tackled the problem of rethinking an existing semantic segmentation architecture into the one suitable for real-time performance, while keeping the performance levels mostly intact. We achieved that by proposing simple modifications to the existing network and highlighting which building blocks were redundant for the final result. Our method can be applied along with any classification network for any dataset and can further benefit from using light-weight backbone networks, and other compression approaches. Quantitatively, we were able to closely match the performance of the original network while significantly surpassing its runtime and even acquiring $55$ FPS on $512\times512$ inputs (from initial $20$ FPS). Besides that, we demonstrate that having convolutions with large kernel sizes can be unnecessary in the decoder part of segmentation networks, and we will devote future work to further cover this topic.\\

\noindent
\textbf{Acknowledgements}
The authors would like to thank the anonymous reviewers for their helpful and constructive comments. This research was supported by the Australian Research Council through the Australian Centre for Robotic Vision (CE140100016), the ARC Laureate Fellowship FL130100102 to IR, and the HPC cluster Phoenix at the University of Adelaide.

\bibliography{egbib}
\newpage

\part*{Supplementary Material}

\vspace{-0.2in}
\section{Experiments}
\label{sec:exps}

In addition to the experiments outlined in the main section, we evaluate our approach on PASCAL Context~\cite{MottaghiCLCLFUY14} and CityScapes~\cite{CordtsORREBFRS16}.\\
\\
\textbf{PASCAL-Context.} This dataset comprises $60$ semantic labels (including background), and consists of $4998$ training images, and $5105$ validation images. During training, we divide the learning rate by half twice after $50$ epochs and after $100$ epochs, respectively, and keep training until $200$ epochs, or earlier convergence. We do not pre-train on PASCAL VOC or COCO.

\begin{figure}[htb]
	\centering
	\resizebox{\textwidth}{!}{\begin{tabular}{cc|cc}
			\subfloat{\includegraphics[width = 0.23\linewidth]{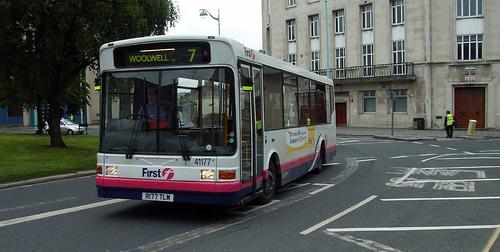}} &
			\subfloat{\includegraphics[width = 0.23\linewidth]{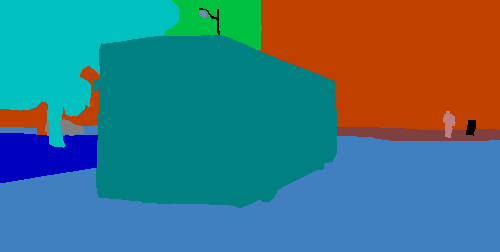}} &
			\subfloat{\includegraphics[width = 0.23\linewidth]{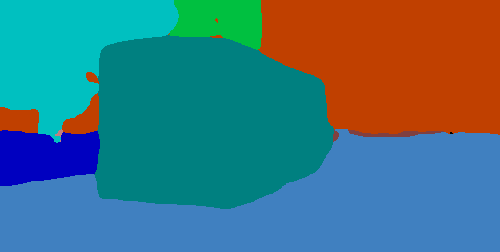}} &
			\subfloat{\includegraphics[width = 0.23\linewidth]{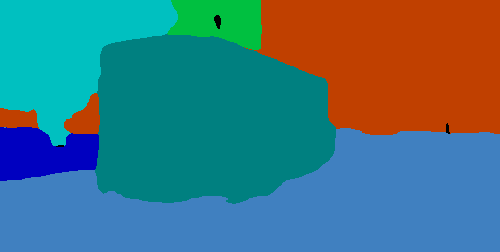}}\\[-0.15in]
			\subfloat{\includegraphics[width = 0.23\linewidth]{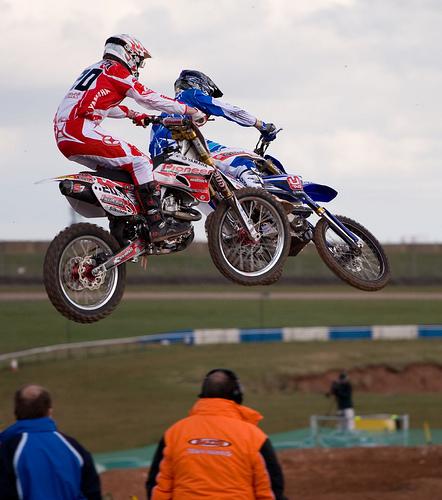}} &
			\subfloat{\includegraphics[width = 0.23\linewidth]{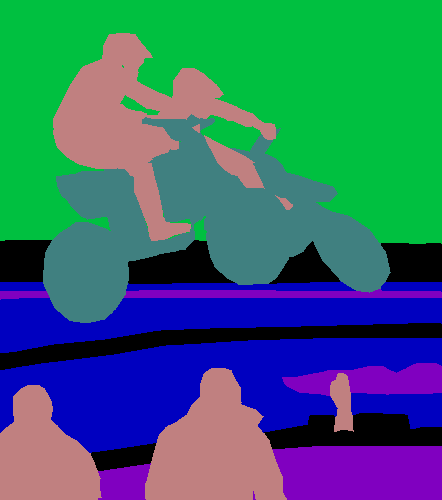}} &
			\subfloat{\includegraphics[width = 0.23\linewidth]{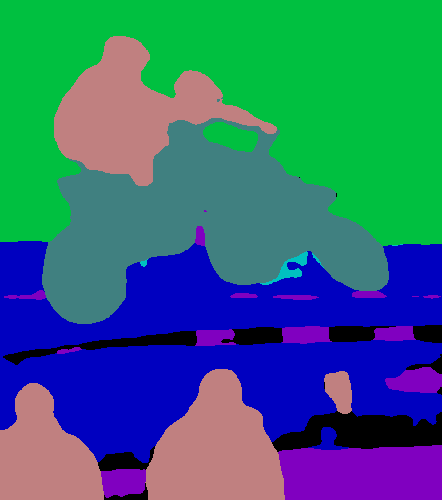}} &
			\subfloat{\includegraphics[width = 0.23\linewidth]{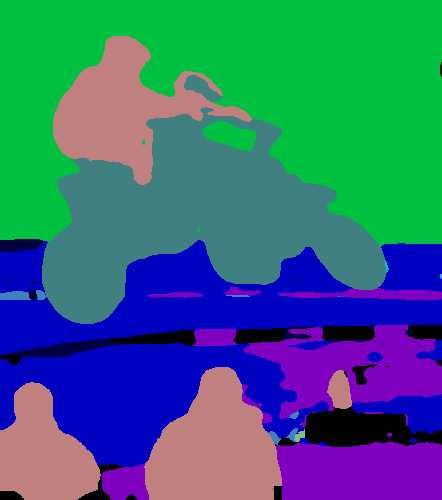}}\\[-0.15in]
			\subfloat{\includegraphics[width = 0.23\linewidth]{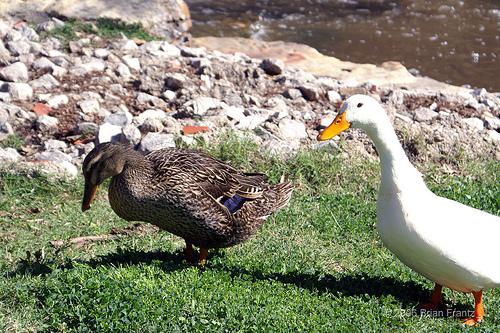}} &
			\subfloat{\includegraphics[width = 0.23\linewidth]{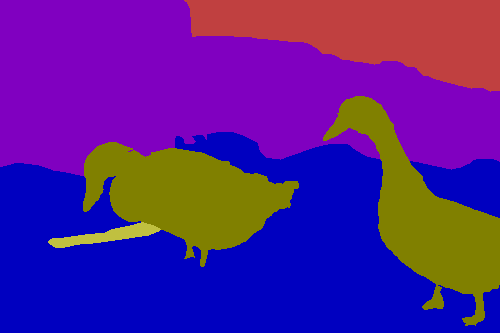}} &
			\subfloat{\includegraphics[width = 0.23\linewidth]{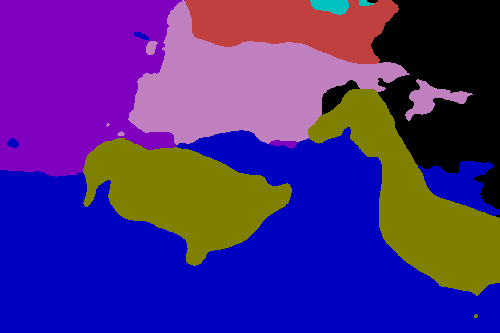}} &
			\subfloat{\includegraphics[width = 0.23\linewidth]{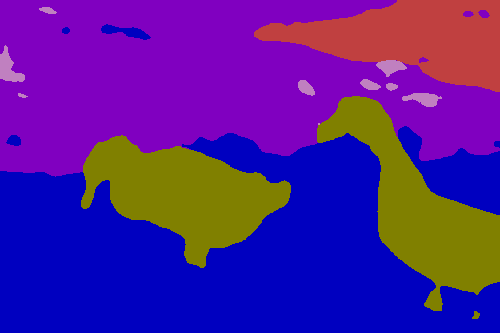}}\\[-0.15in]
			\subfloat{\includegraphics[width = 0.23\linewidth]{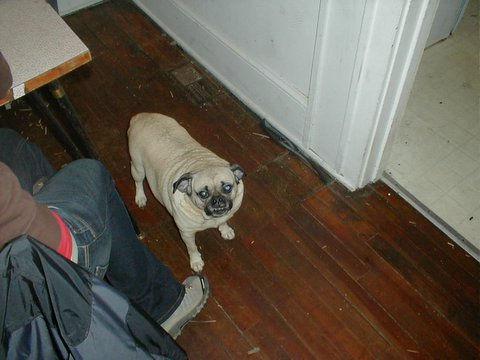}} &
			\subfloat{\includegraphics[width = 0.23\linewidth]{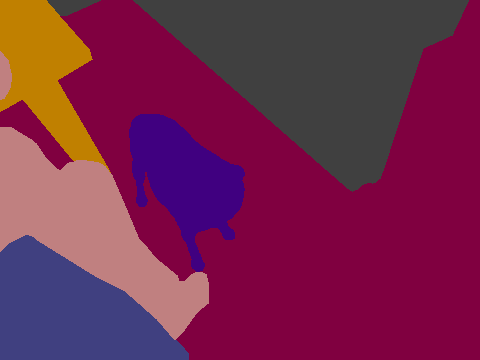}} &
			\subfloat{\includegraphics[width = 0.23\linewidth]{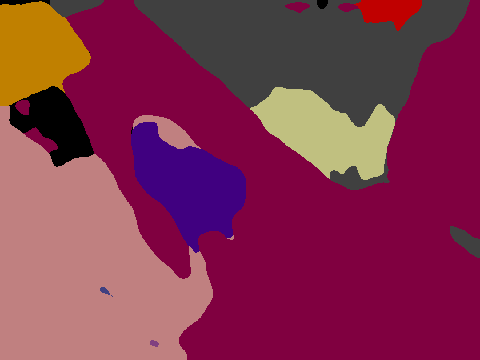}} &
			\subfloat{\includegraphics[width = 0.23\linewidth]{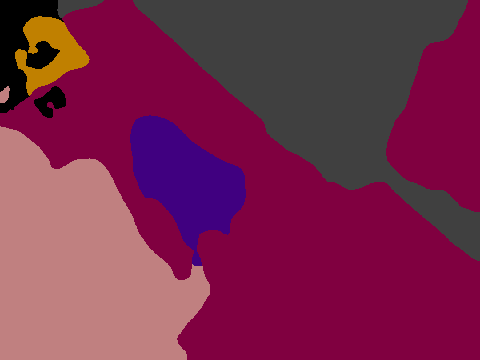}}\\
			Image&GT&RF-101-LW&RF-152-LW
		\end{tabular}}
		\vskip 0.1in
		\caption{Visual results on validation set of PASCAL-Context with residual models.}
		\label{fig:context}
	\end{figure}

Our quantitative results are provided in Table~\ref{table:context} and qualitative results are on Figure~\ref{fig:context}.\\
\\
\textbf{CityScapes.}
Finally, we turn our attention to the CityScapes dataset~\cite{CordtsORREBFRS16}, that contains $5000$ high-resolution ($1024\times2048$) images with $19$ semantic classes, of which $2975$ images are used for training, $500$ for validation, and $1525$ for testing, respectively. We use the same learning strategy as for Context. Our single-scale model with ResNet-101 as backbone, is able to achieve $72.1\%$ mean iou on the test set, which is close to the original RefineNet result of $73.6\%$ with multi-scale evaluation~\cite{LinMSR17}.\par
\begin{figure}[htb]
		\centering
		\resizebox{\textwidth}{!}{\begin{tabular}{cc|ccc}
				\subfloat{\includegraphics[width = 0.19\linewidth]{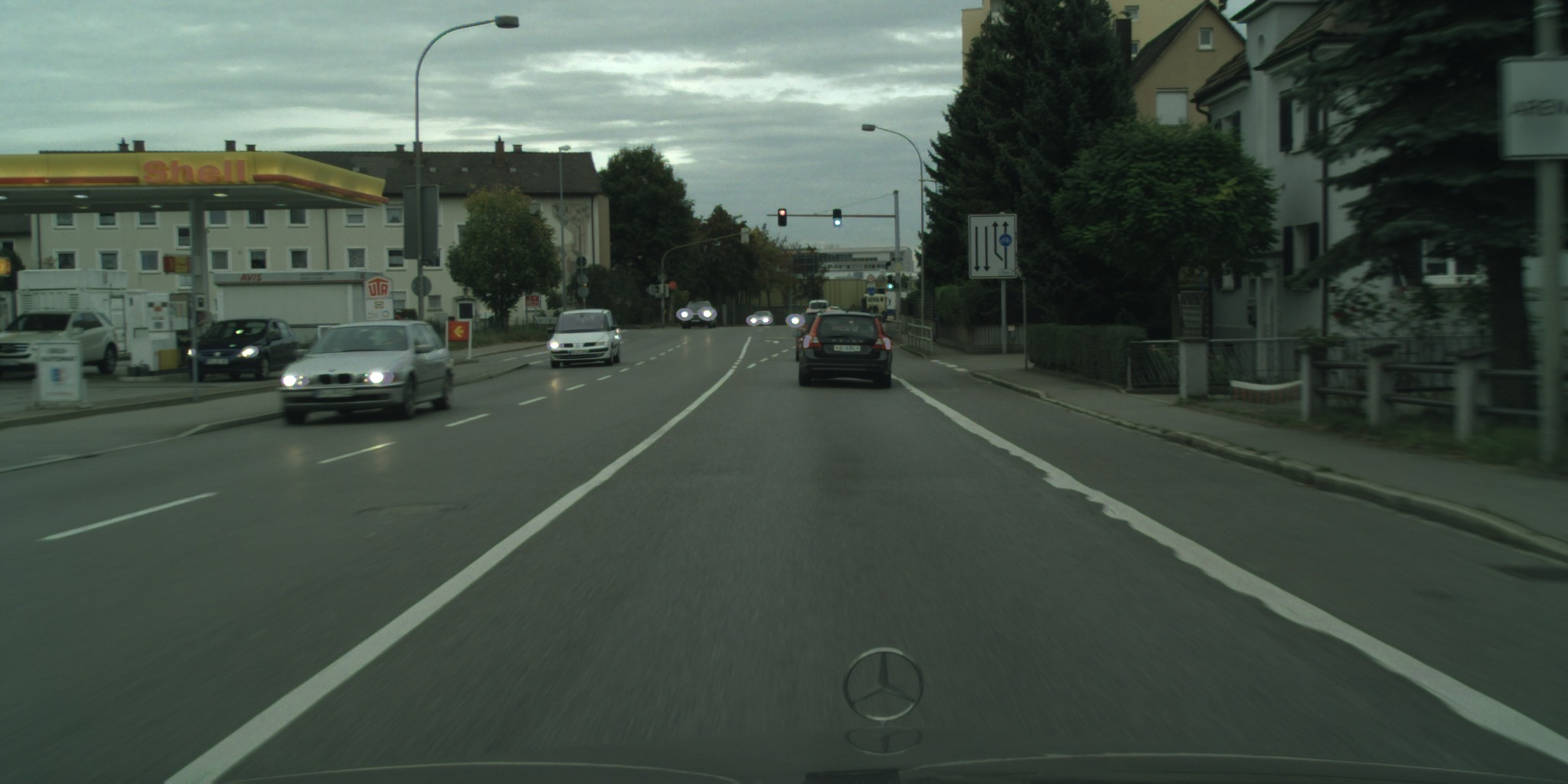}} &
				\subfloat{\includegraphics[width = 0.19\linewidth]{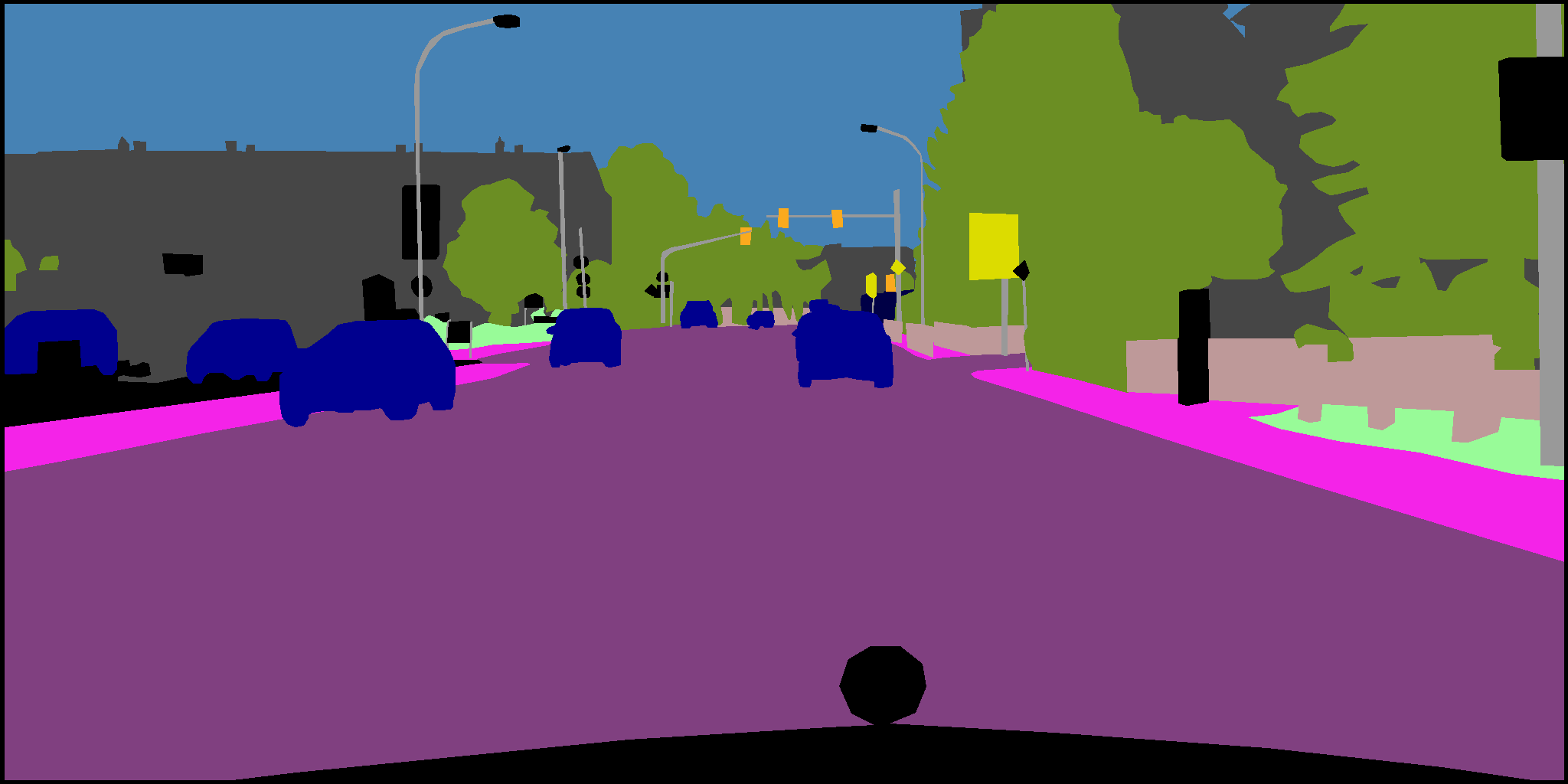}} &
				\subfloat{\includegraphics[width = 0.19\linewidth]{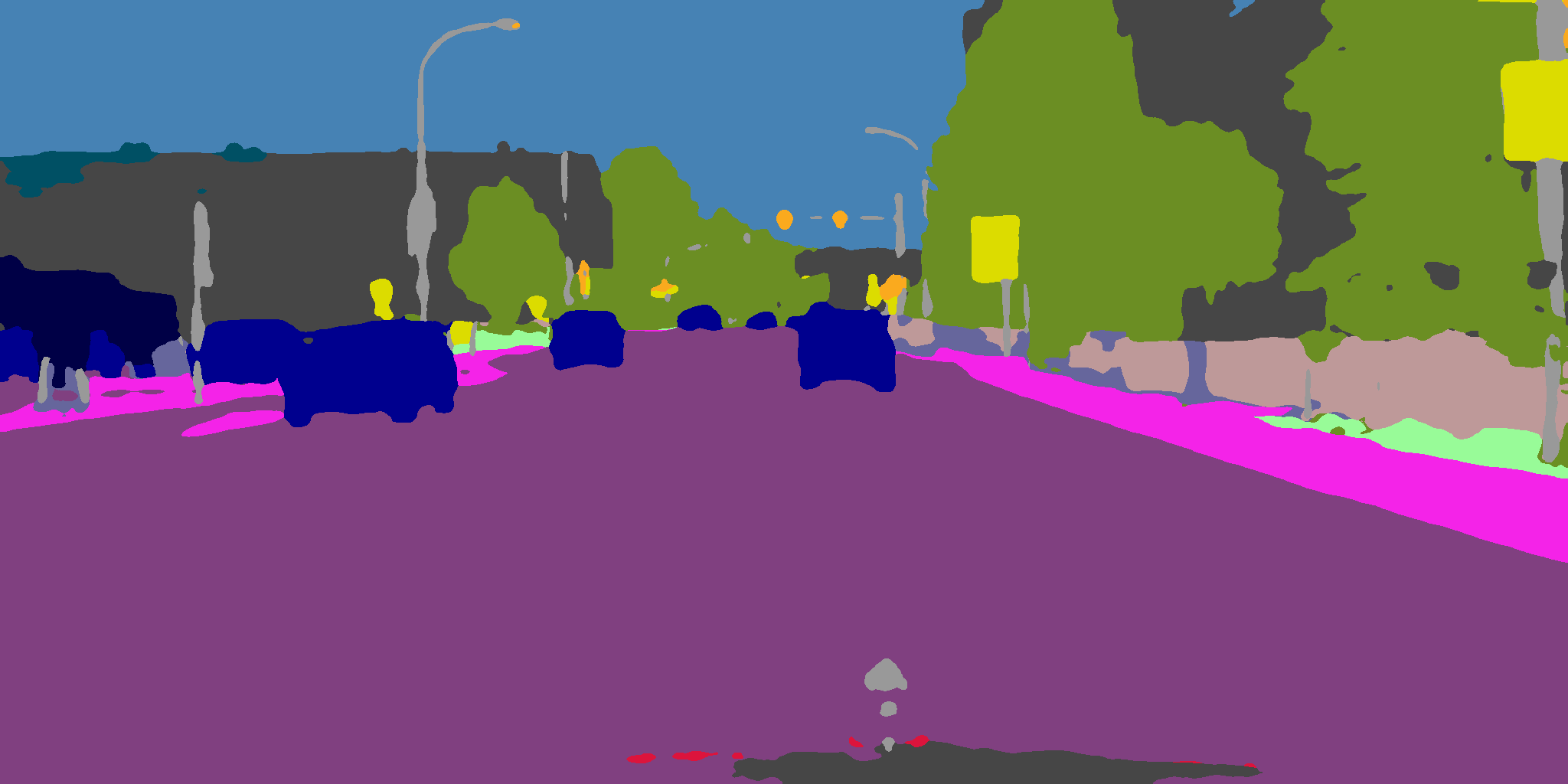}} &
				\subfloat{\includegraphics[width = 0.19\linewidth]{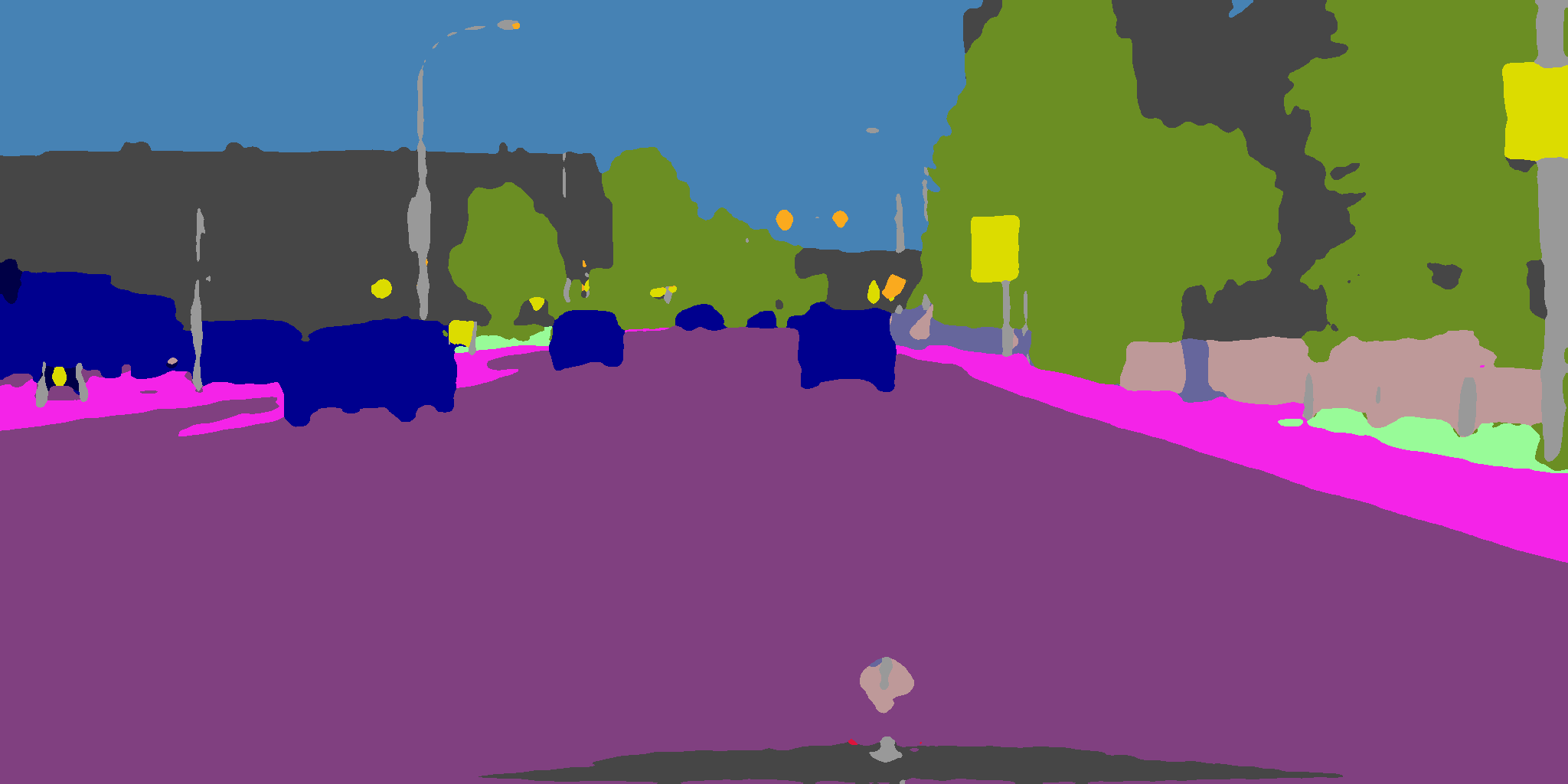}} &
				\subfloat{\includegraphics[width = 0.19\linewidth]{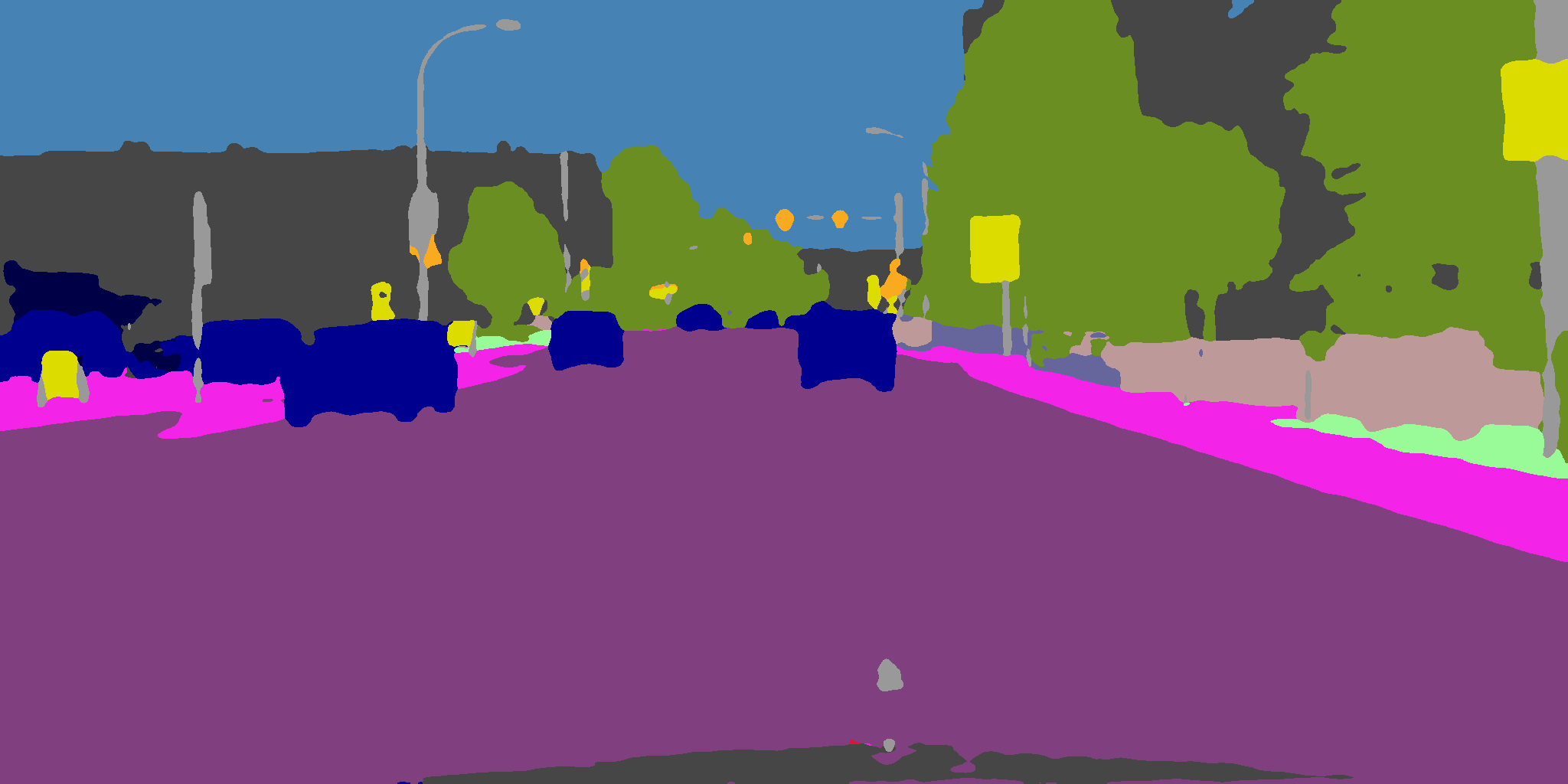}}\\[-0.15in]
				\subfloat{\includegraphics[width = 0.19\linewidth]{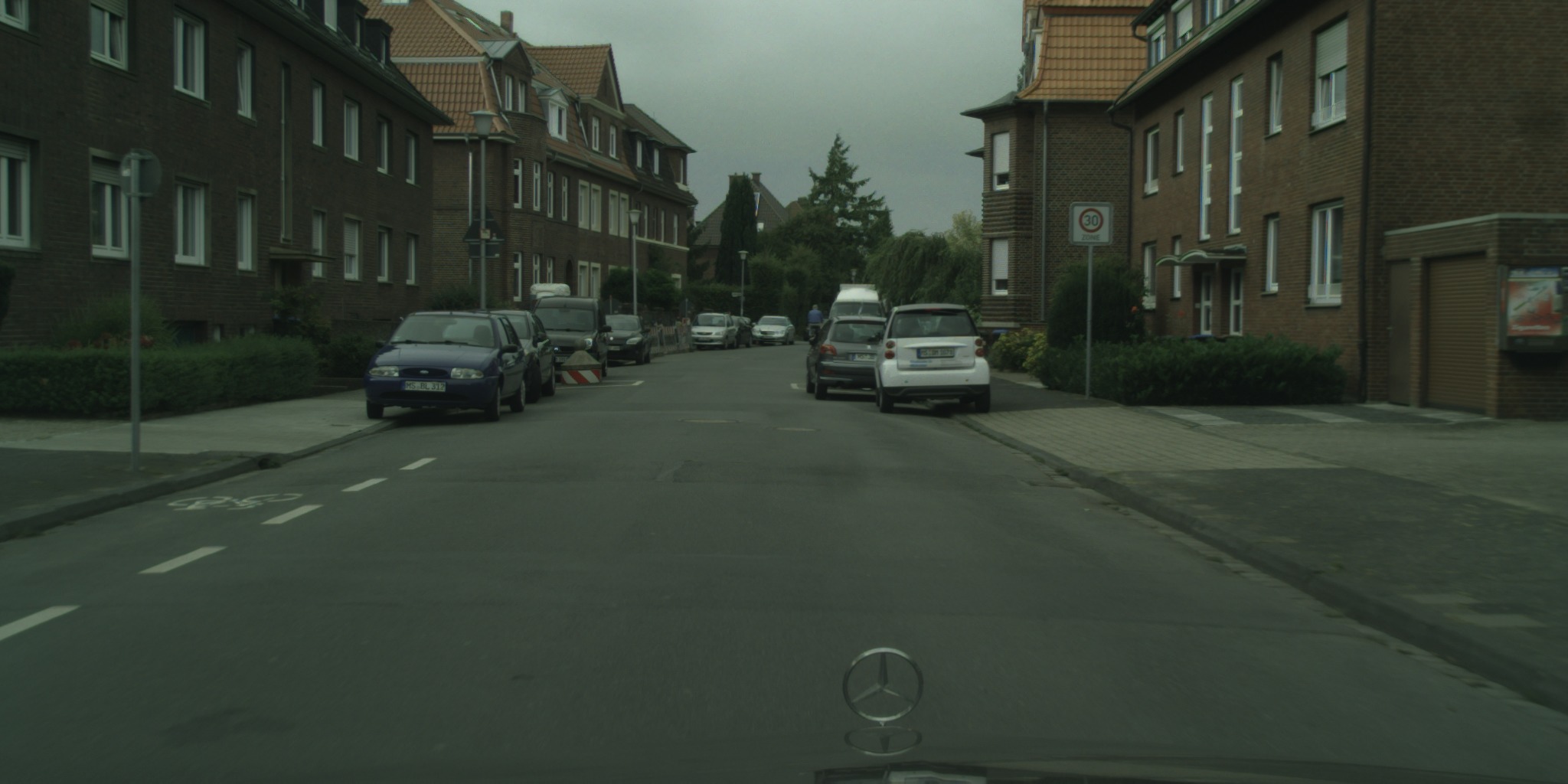}} &
				\subfloat{\includegraphics[width = 0.19\linewidth]{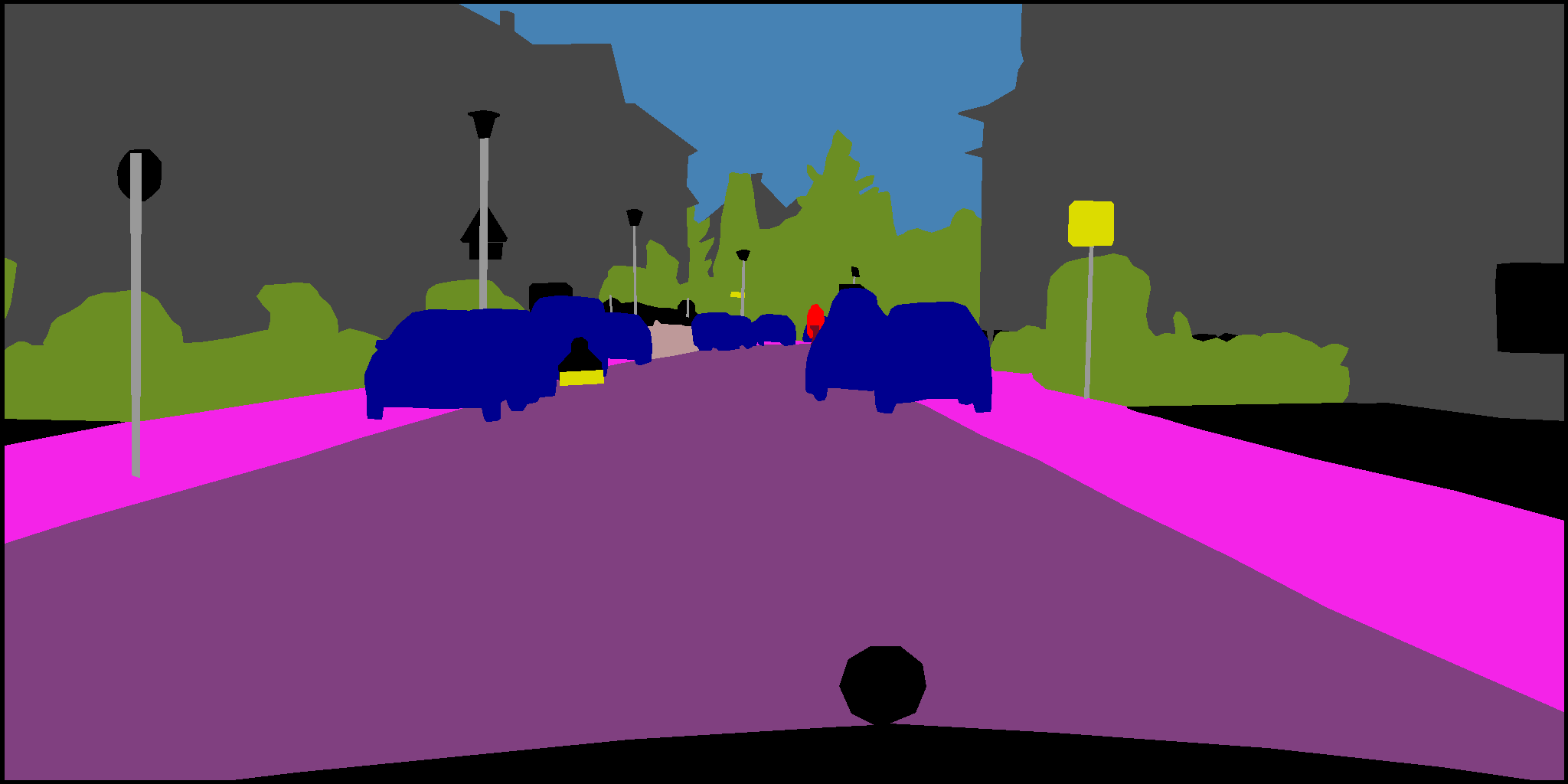}} &
				\subfloat{\includegraphics[width = 0.19\linewidth]{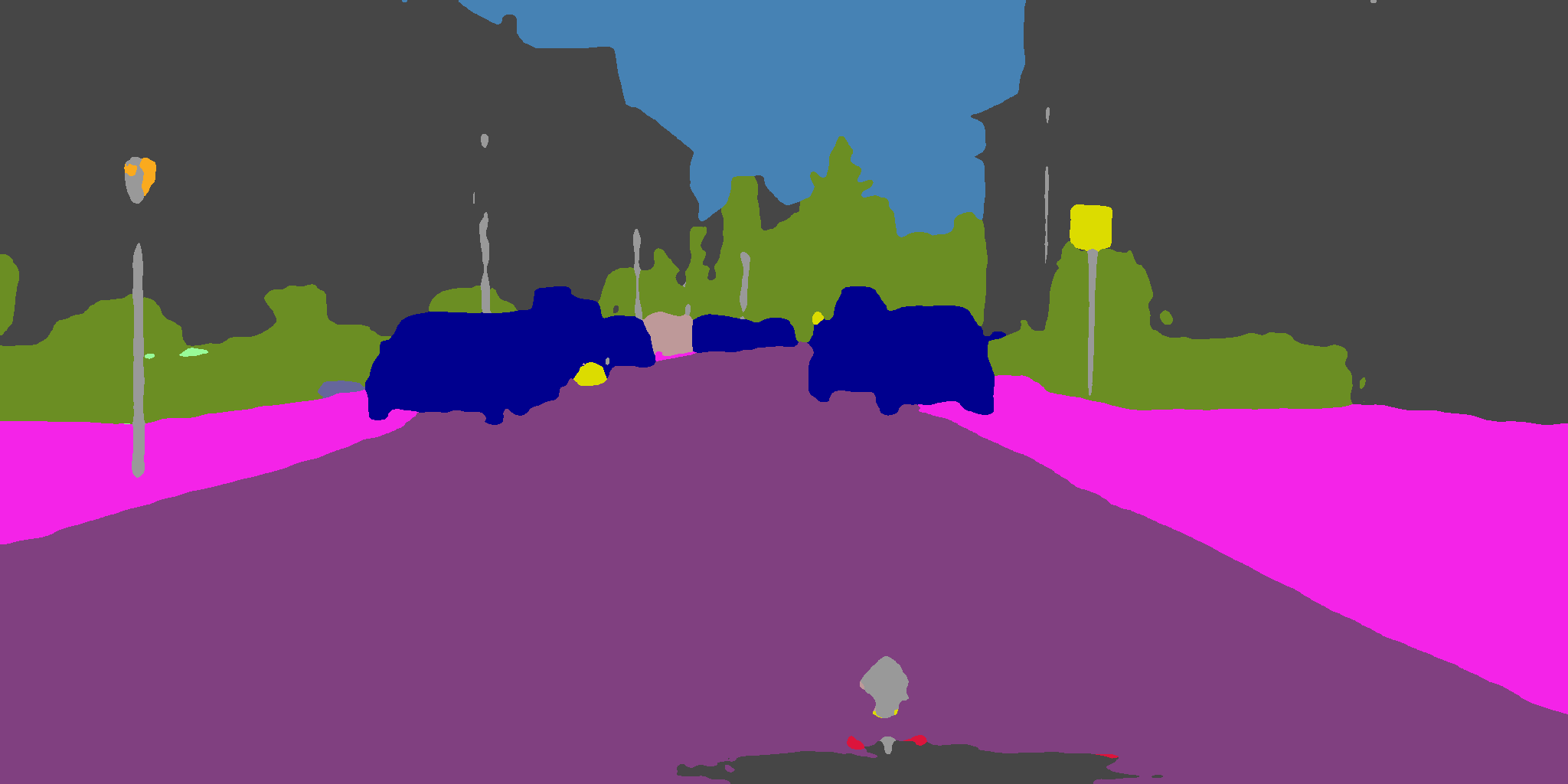}} &
				\subfloat{\includegraphics[width = 0.19\linewidth]{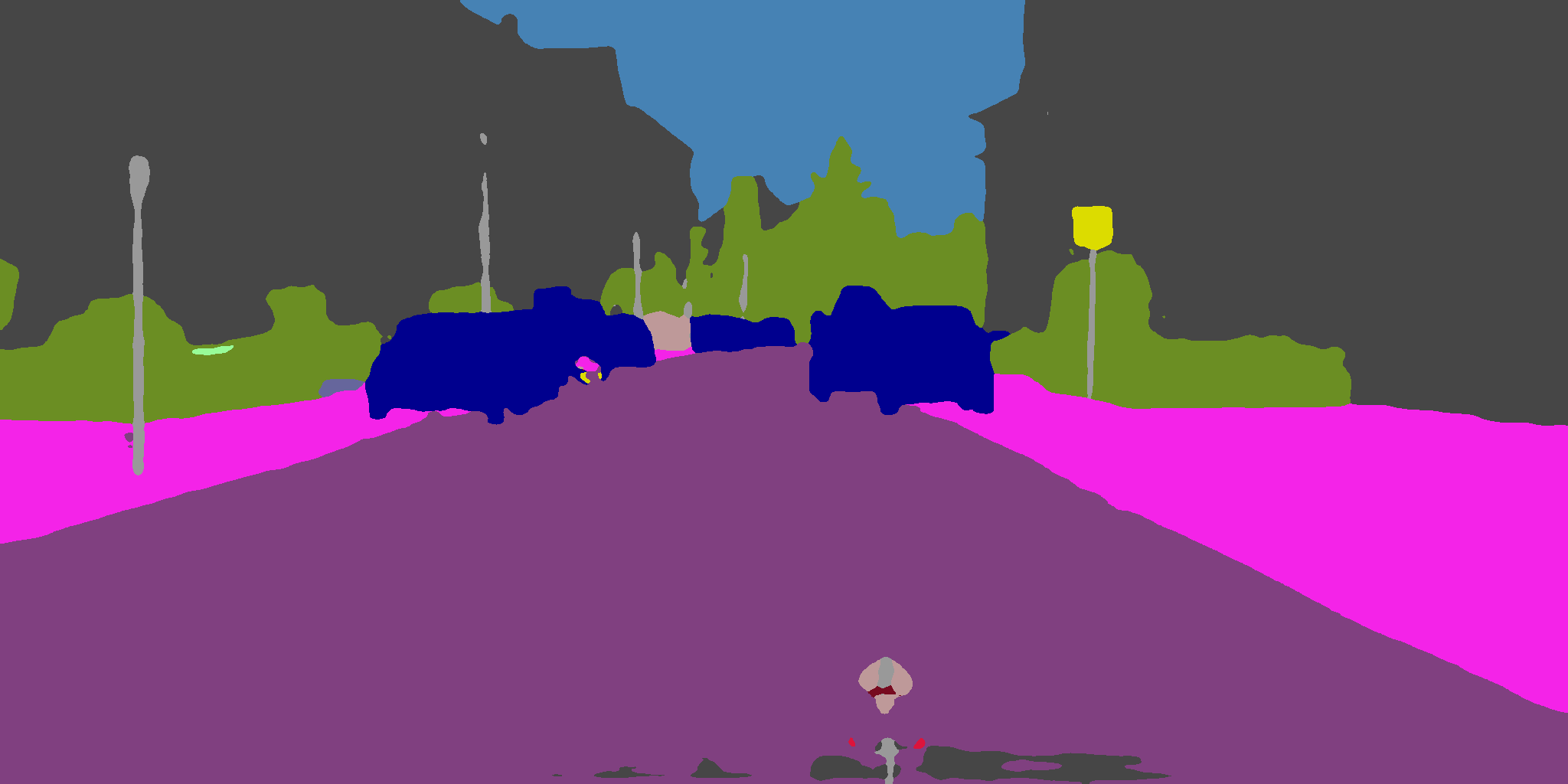}} &
				\subfloat{\includegraphics[width = 0.19\linewidth]{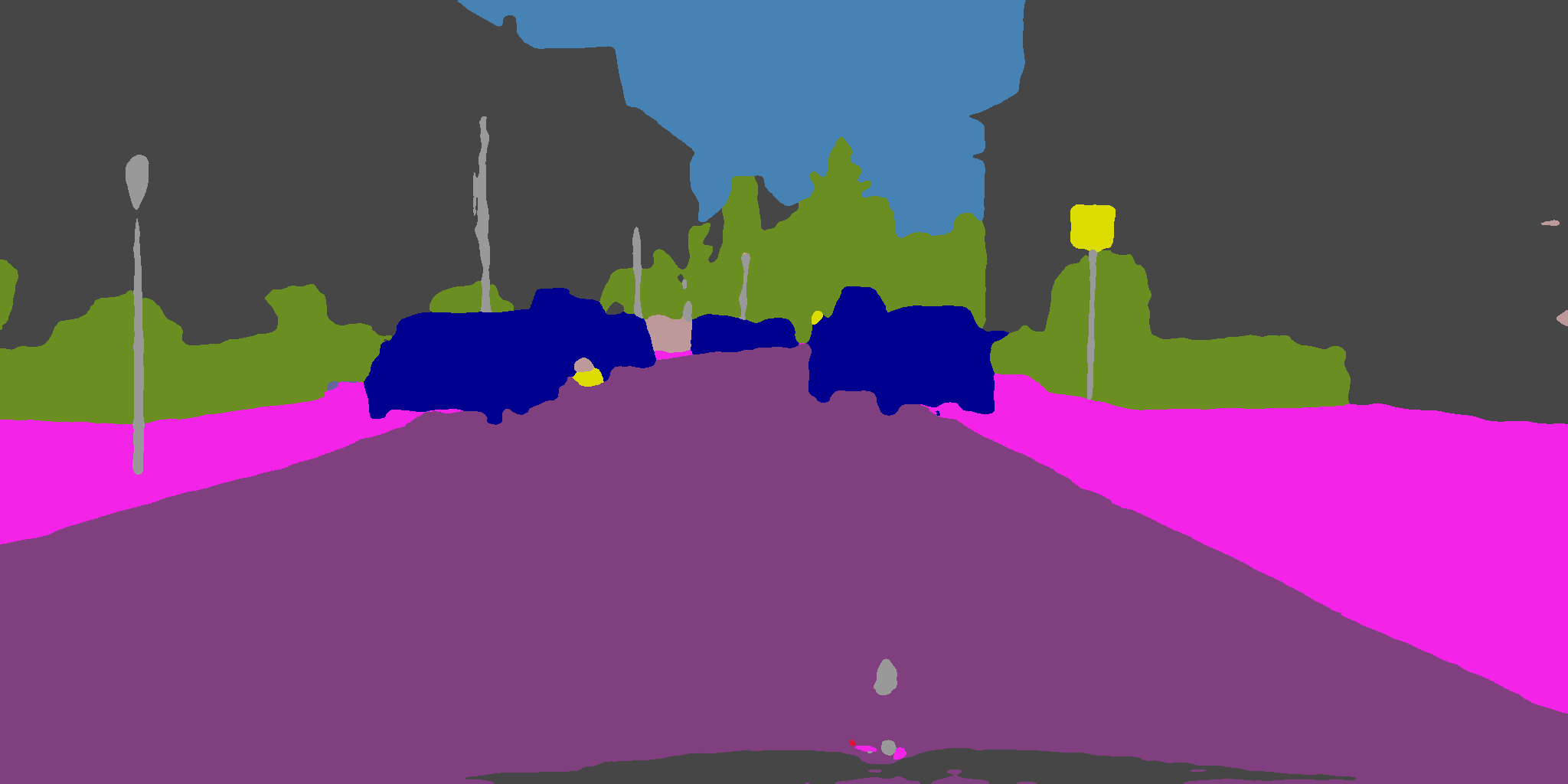}}\\[-0.15in]
				\subfloat{\includegraphics[width = 0.19\linewidth]{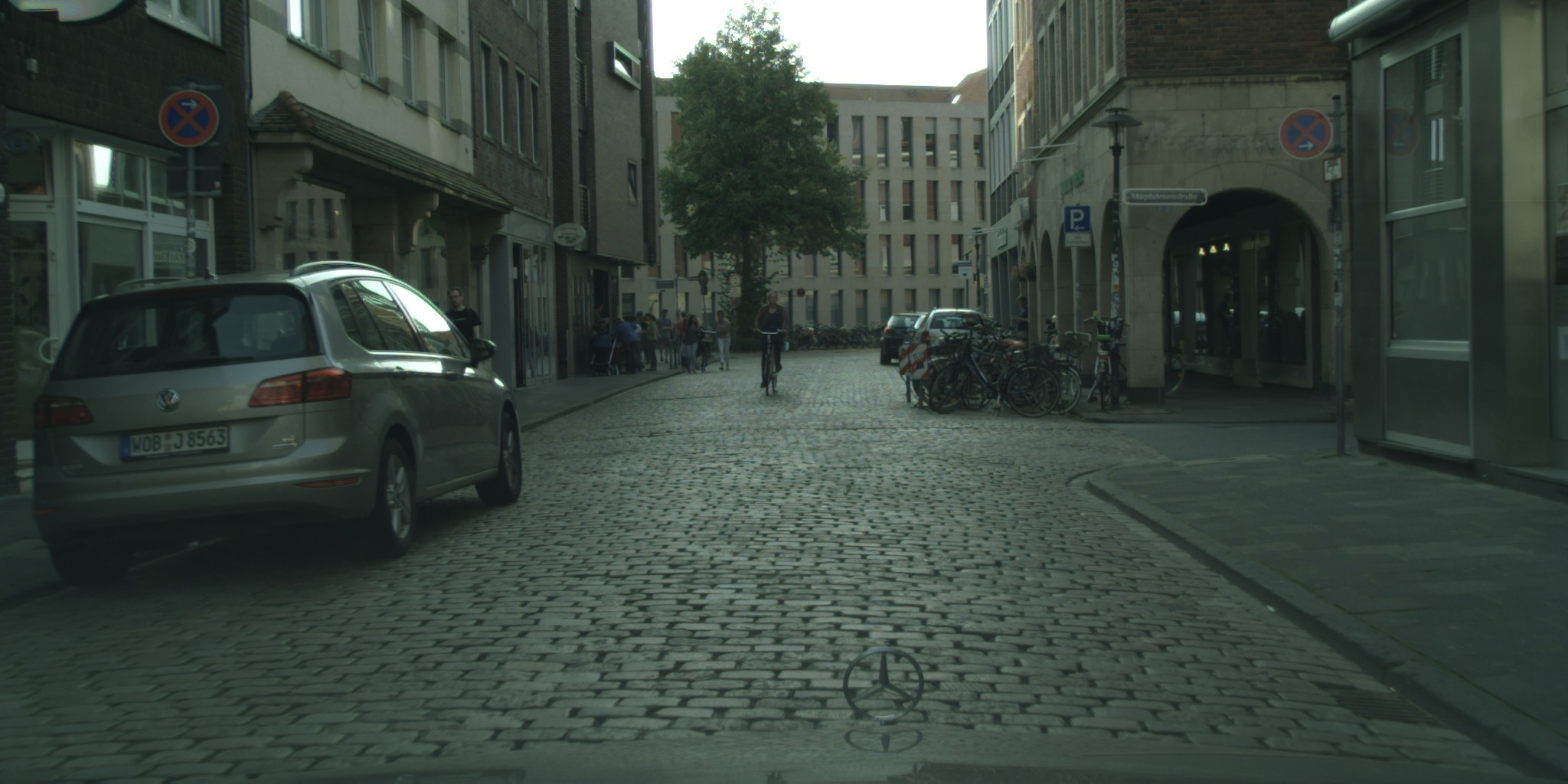}} &
				\subfloat{\includegraphics[width = 0.19\linewidth]{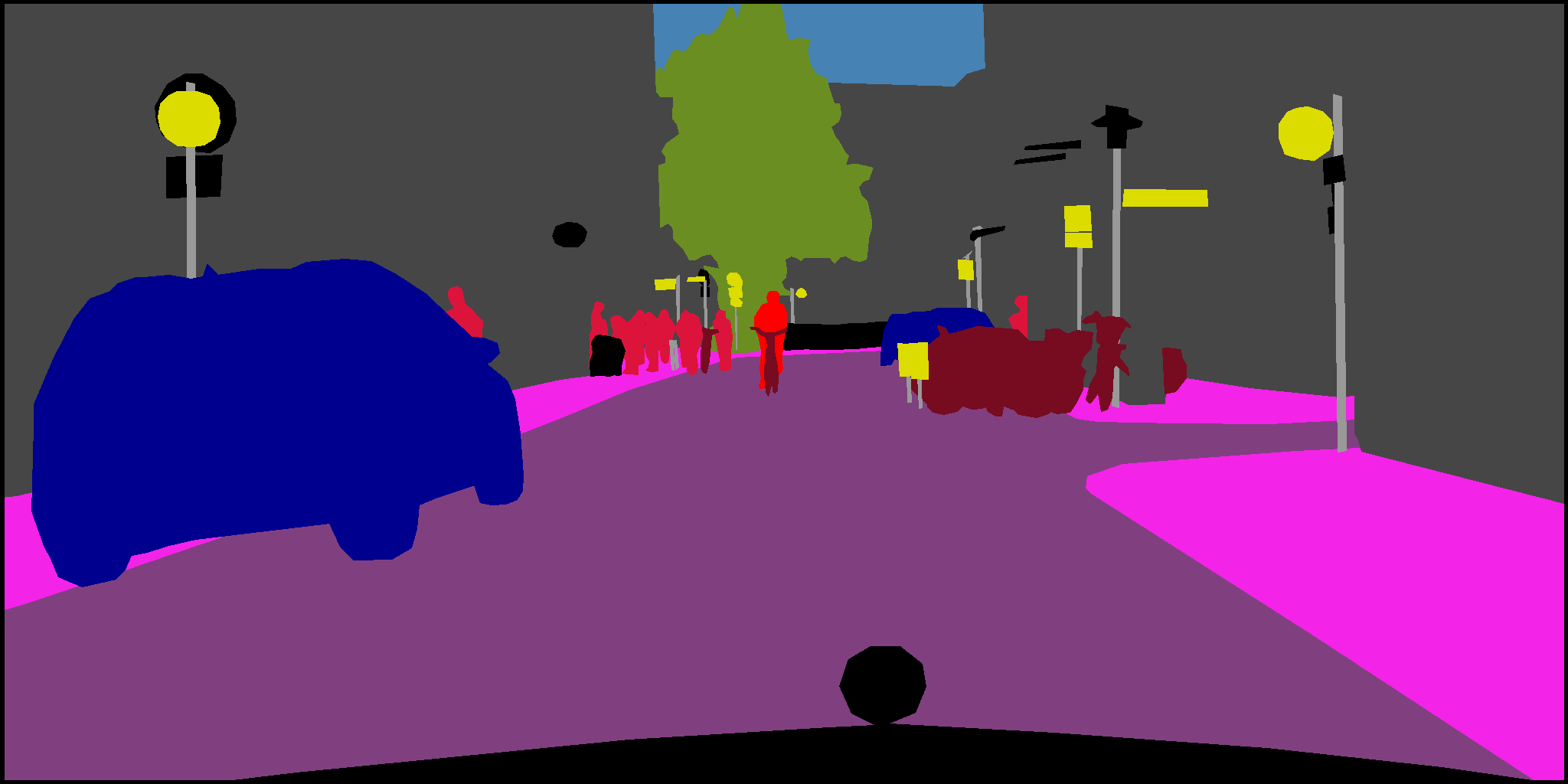}} &
				\subfloat{\includegraphics[width = 0.19\linewidth]{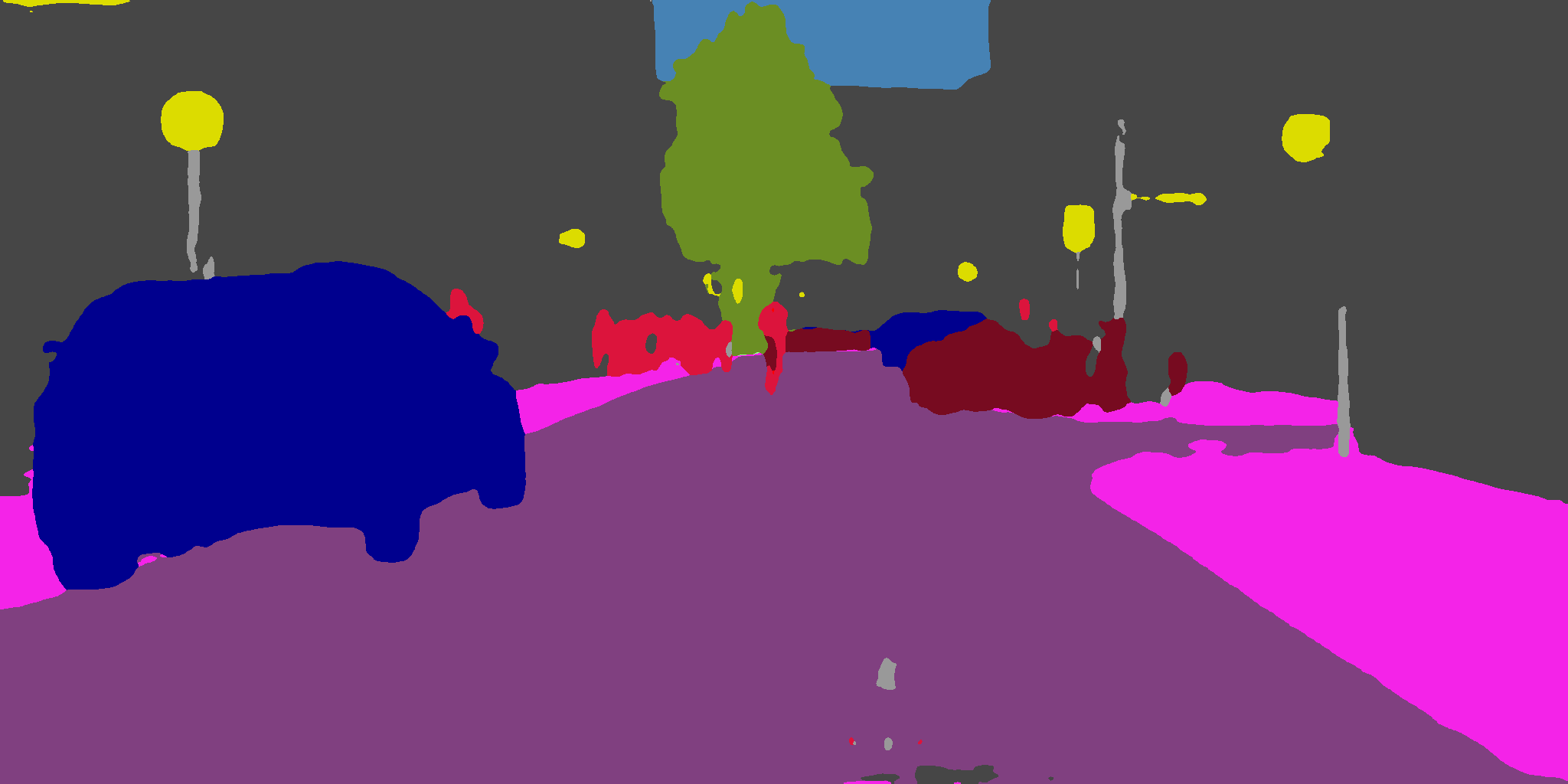}} &
				\subfloat{\includegraphics[width = 0.19\linewidth]{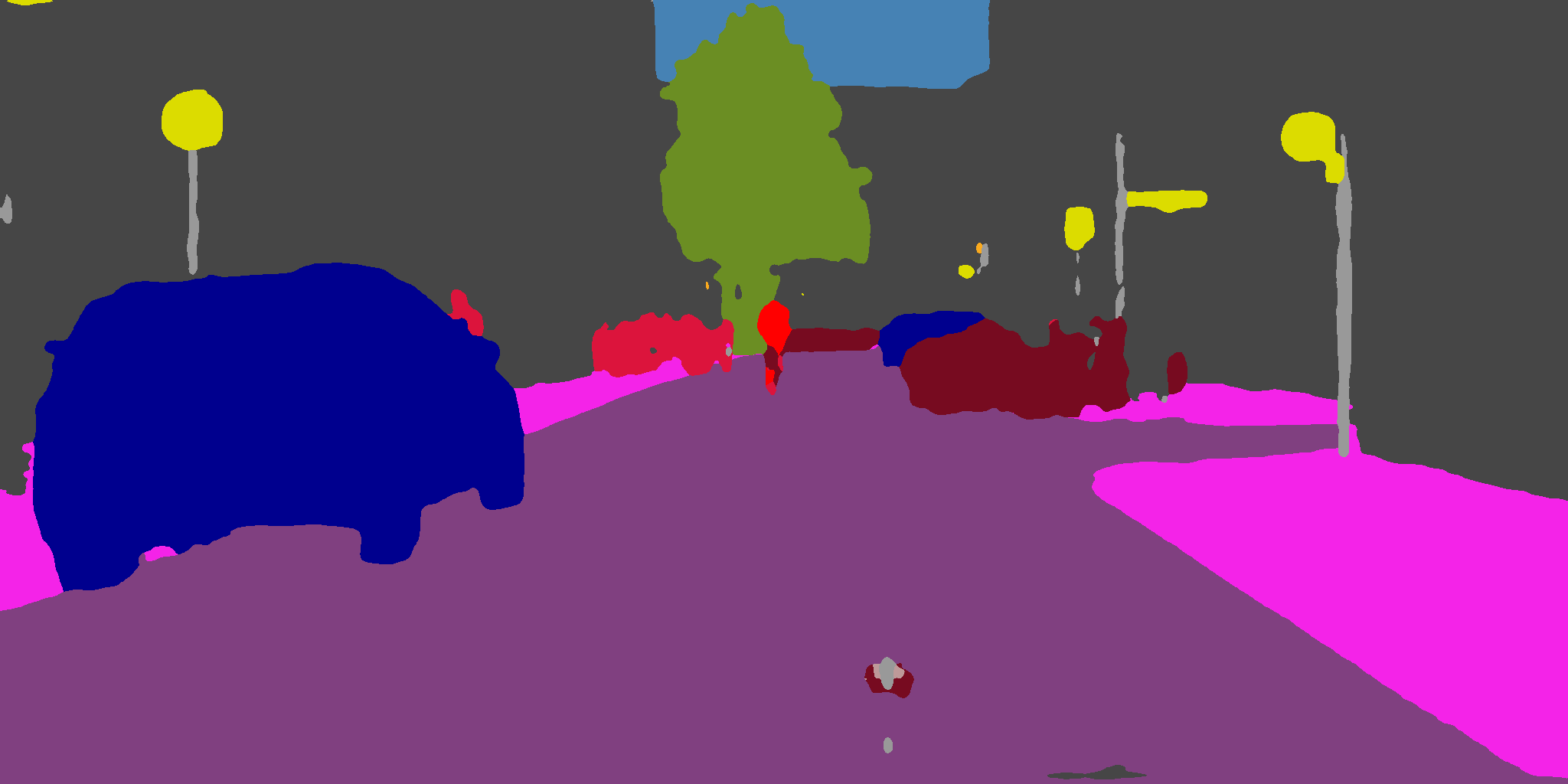}} &
				\subfloat{\includegraphics[width = 0.19\linewidth]{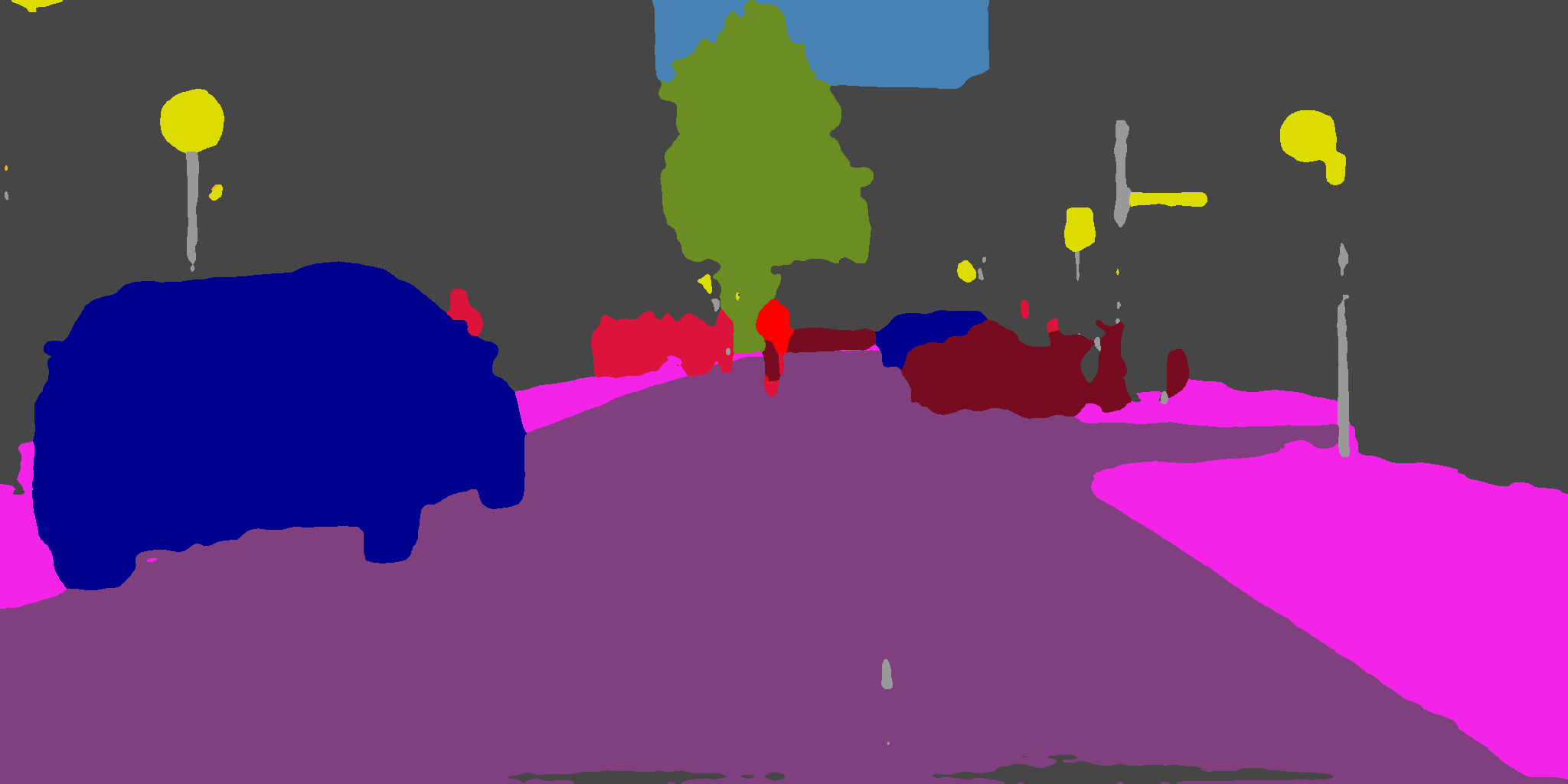}}\\[-0.15in]
				\subfloat{\includegraphics[width = 0.19\linewidth]{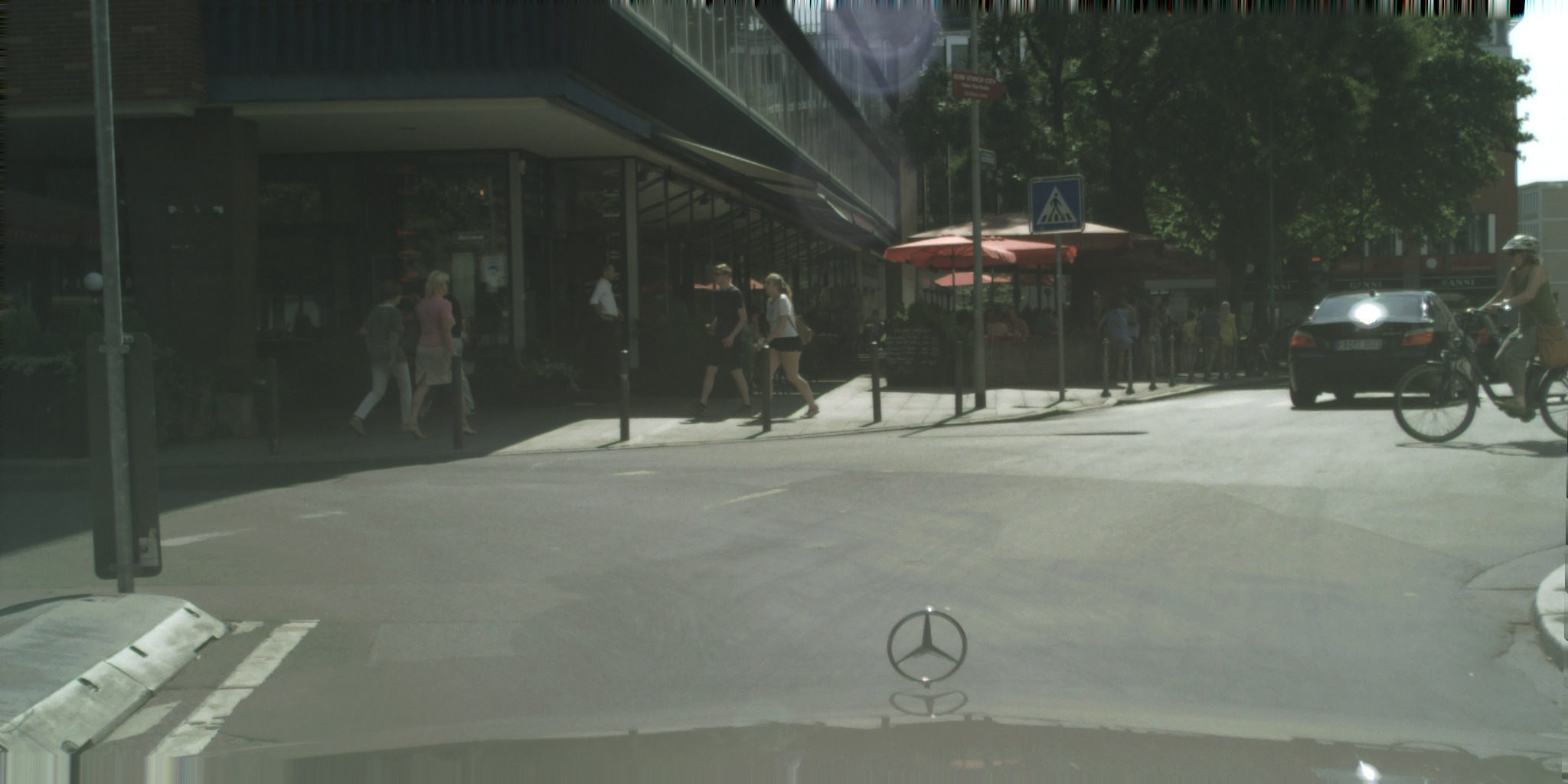}} &
				\subfloat{\includegraphics[width = 0.19\linewidth]{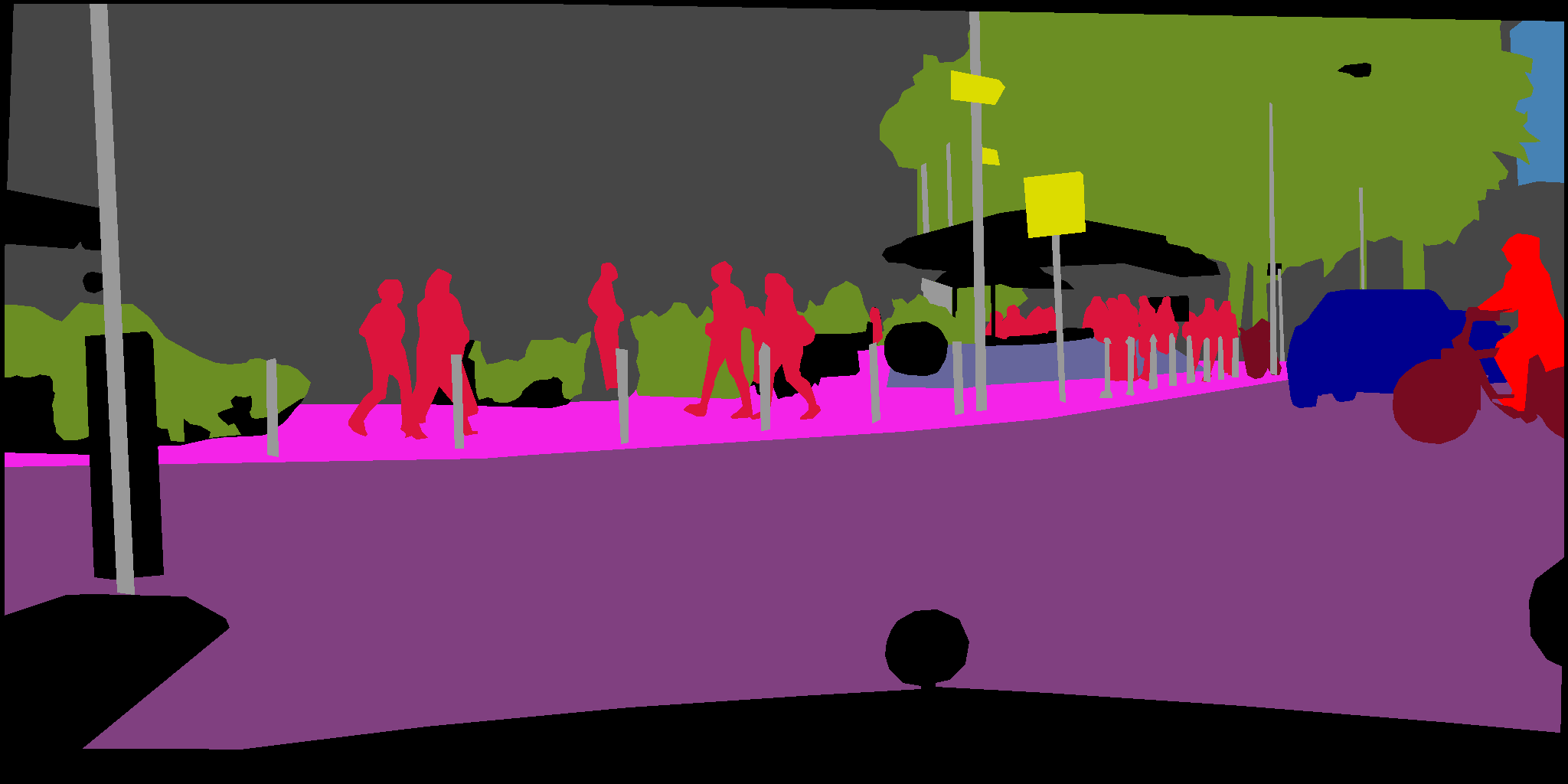}} &
				\subfloat{\includegraphics[width = 0.19\linewidth]{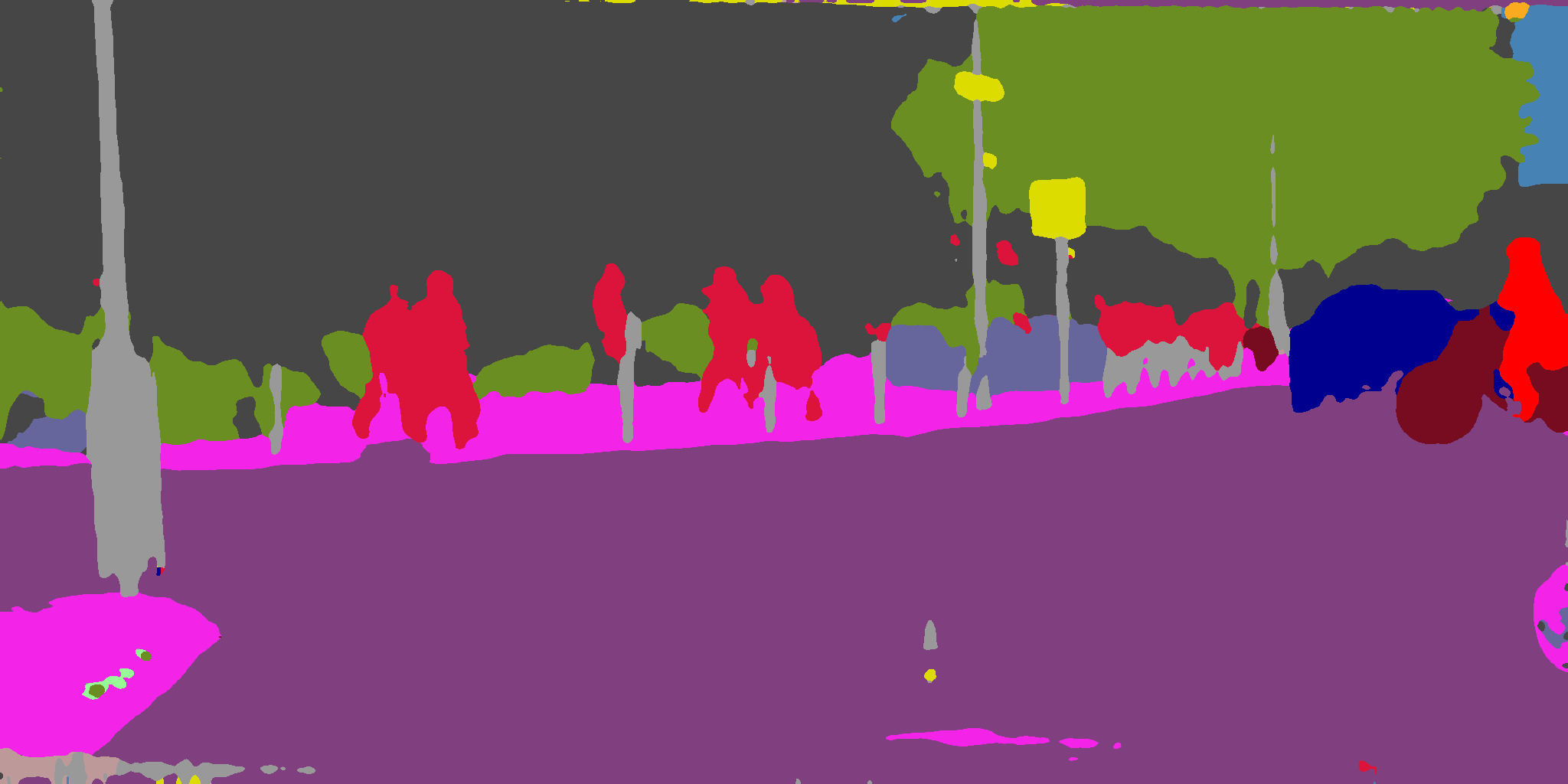}} &
				\subfloat{\includegraphics[width = 0.19\linewidth]{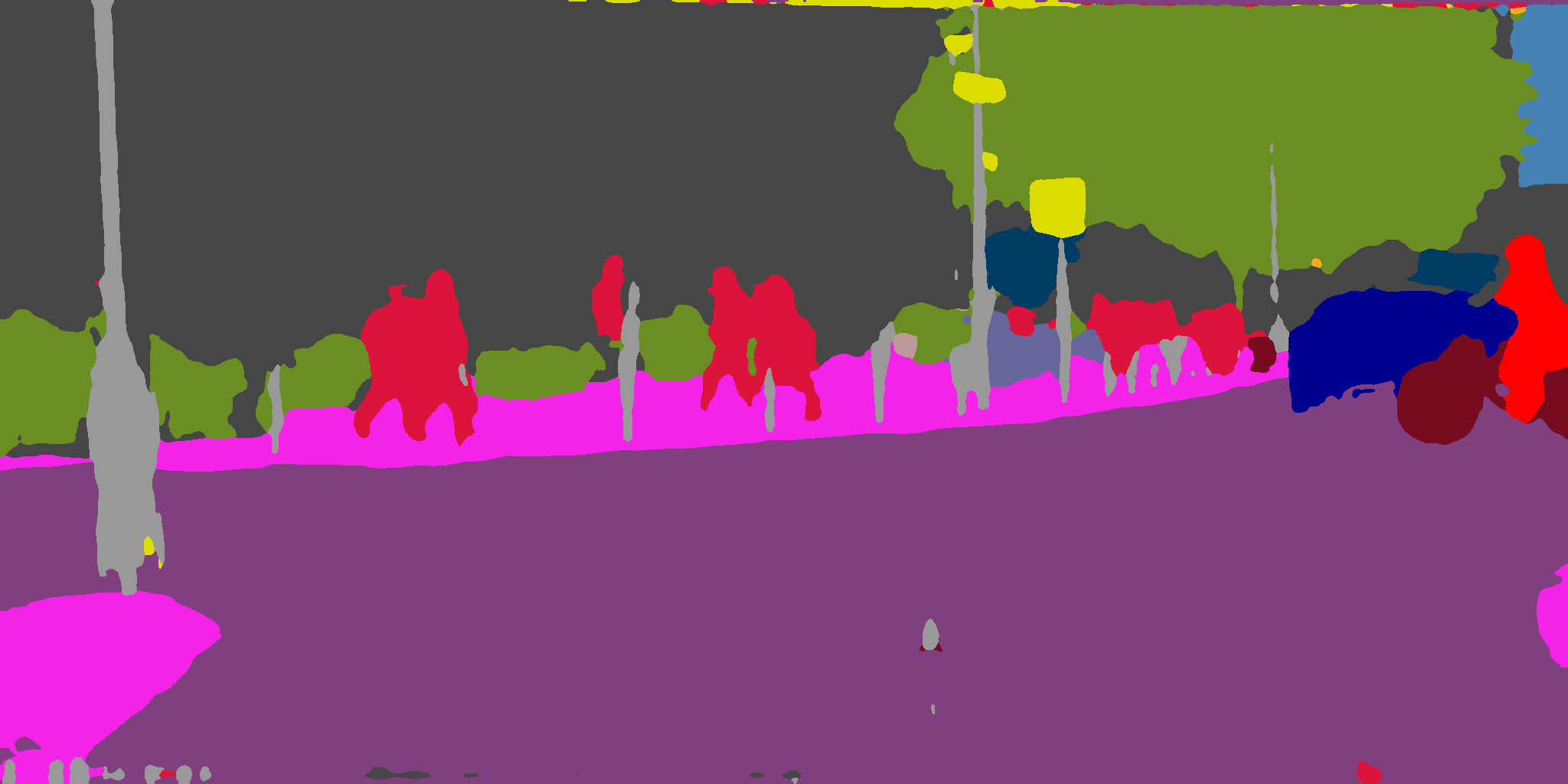}} &
				\subfloat{\includegraphics[width = 0.19\linewidth]{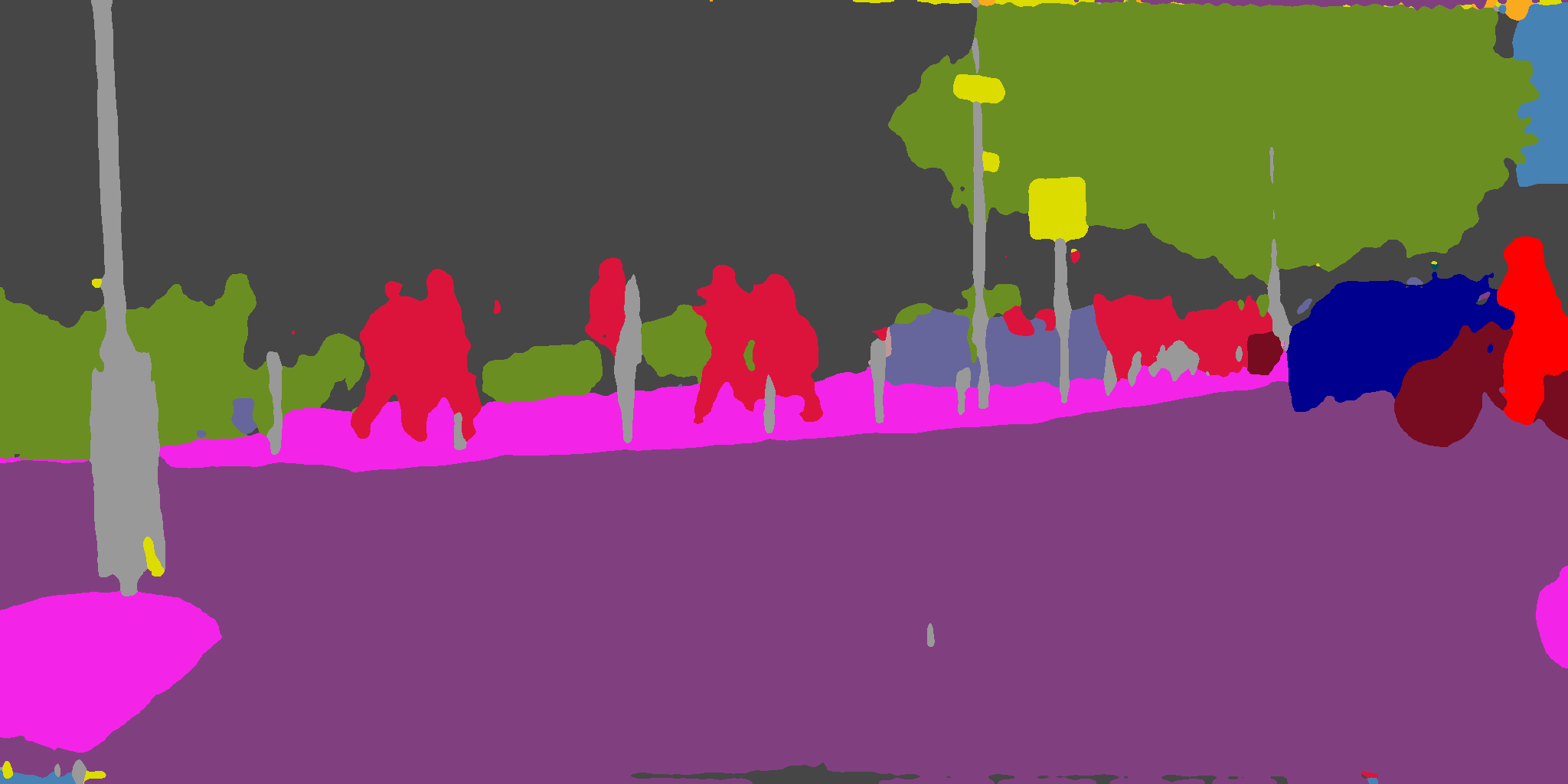}}\\[-0.15in]
				\subfloat{\includegraphics[width = 0.19\linewidth]{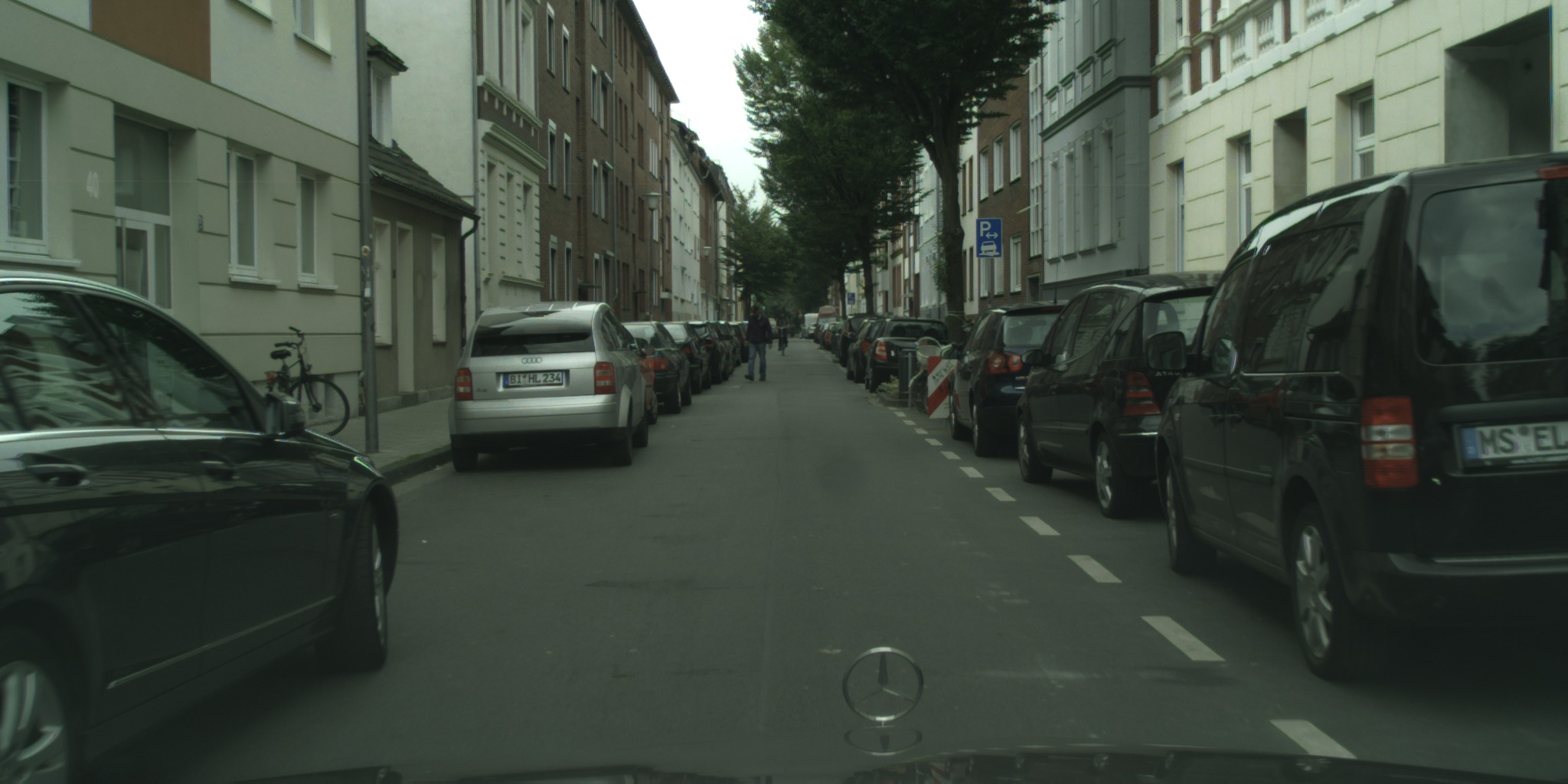}} &
				\subfloat{\includegraphics[width = 0.19\linewidth]{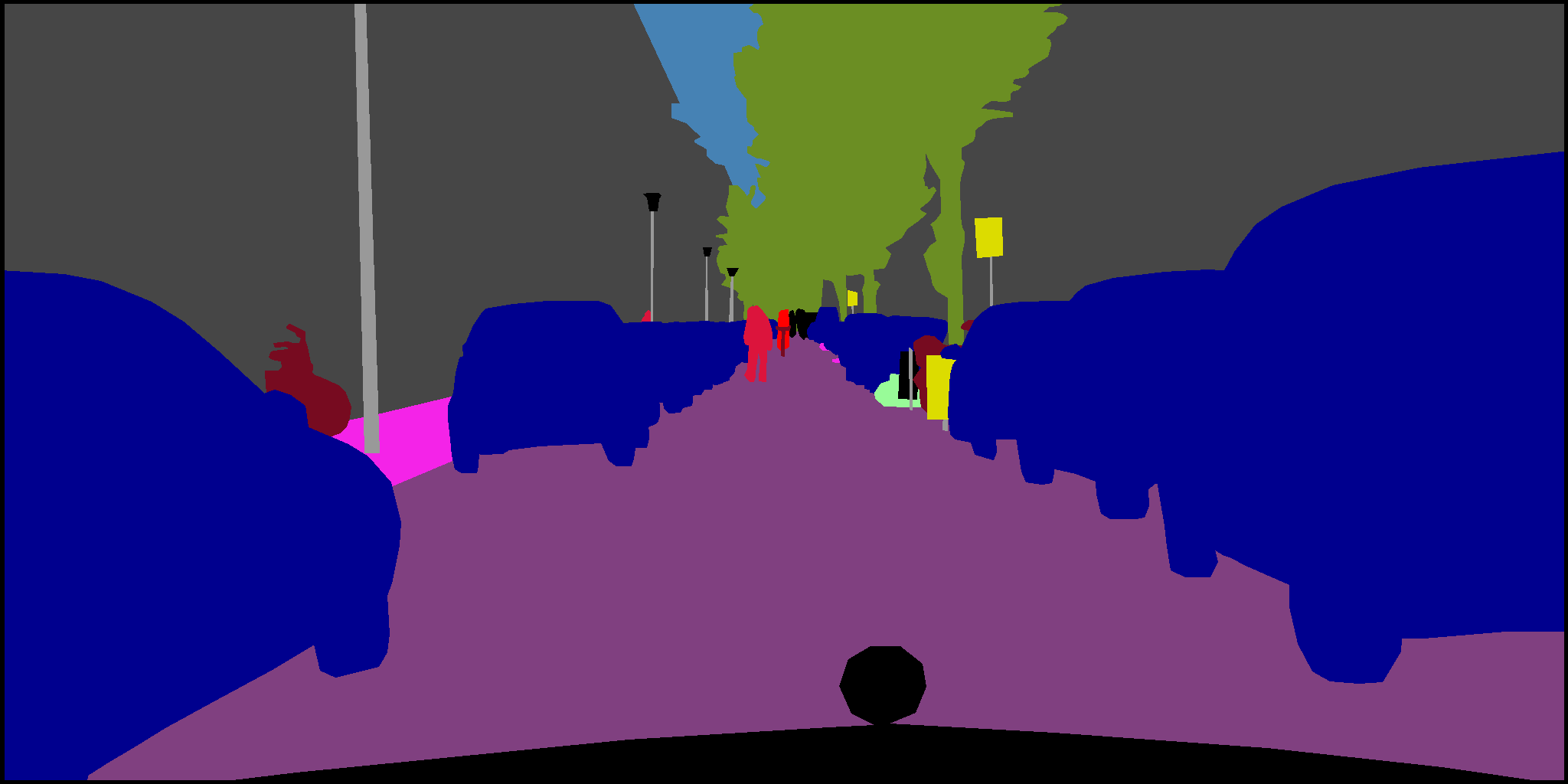}} &
				\subfloat{\includegraphics[width = 0.19\linewidth]{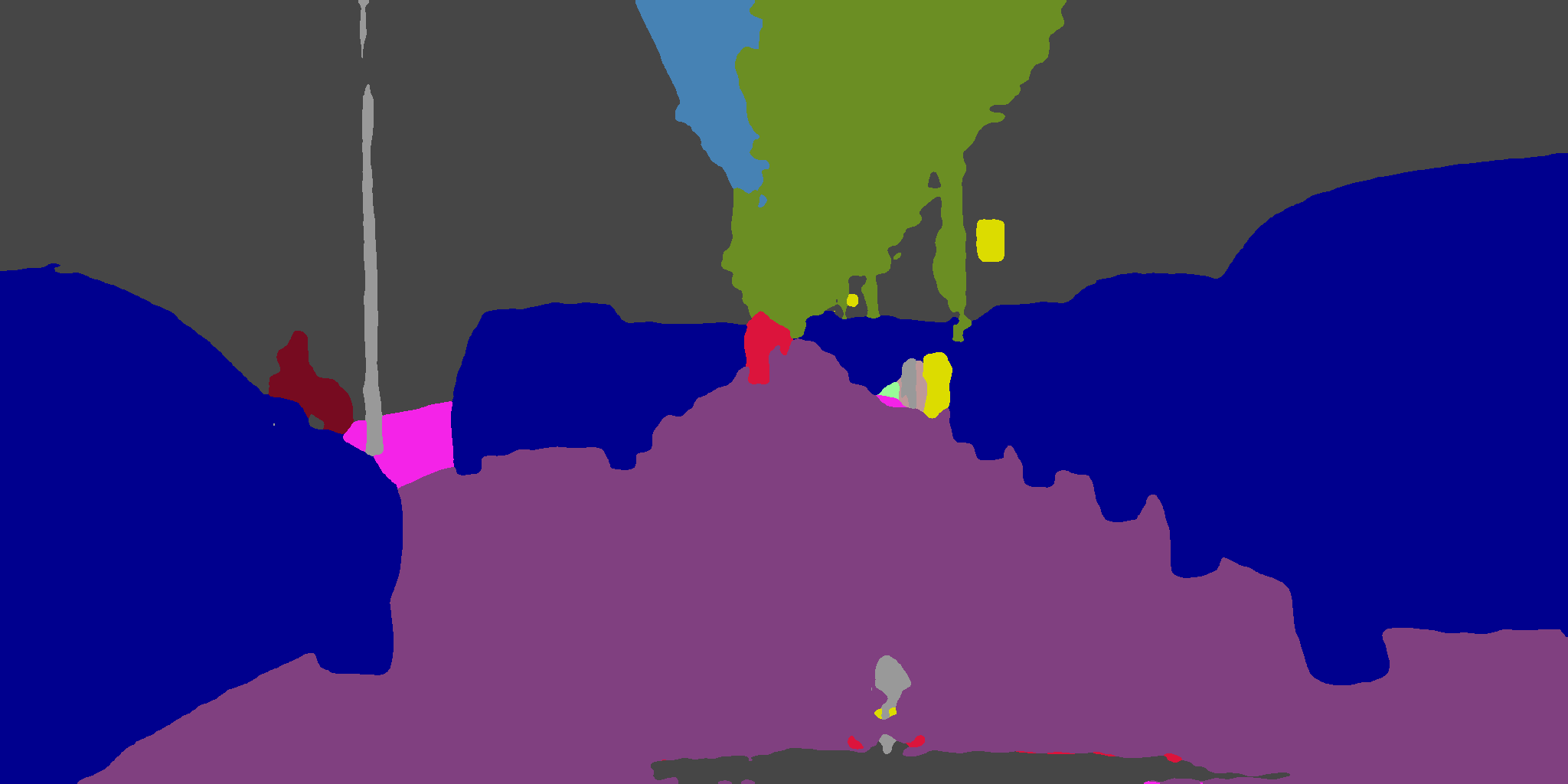}} &
				\subfloat{\includegraphics[width = 0.19\linewidth]{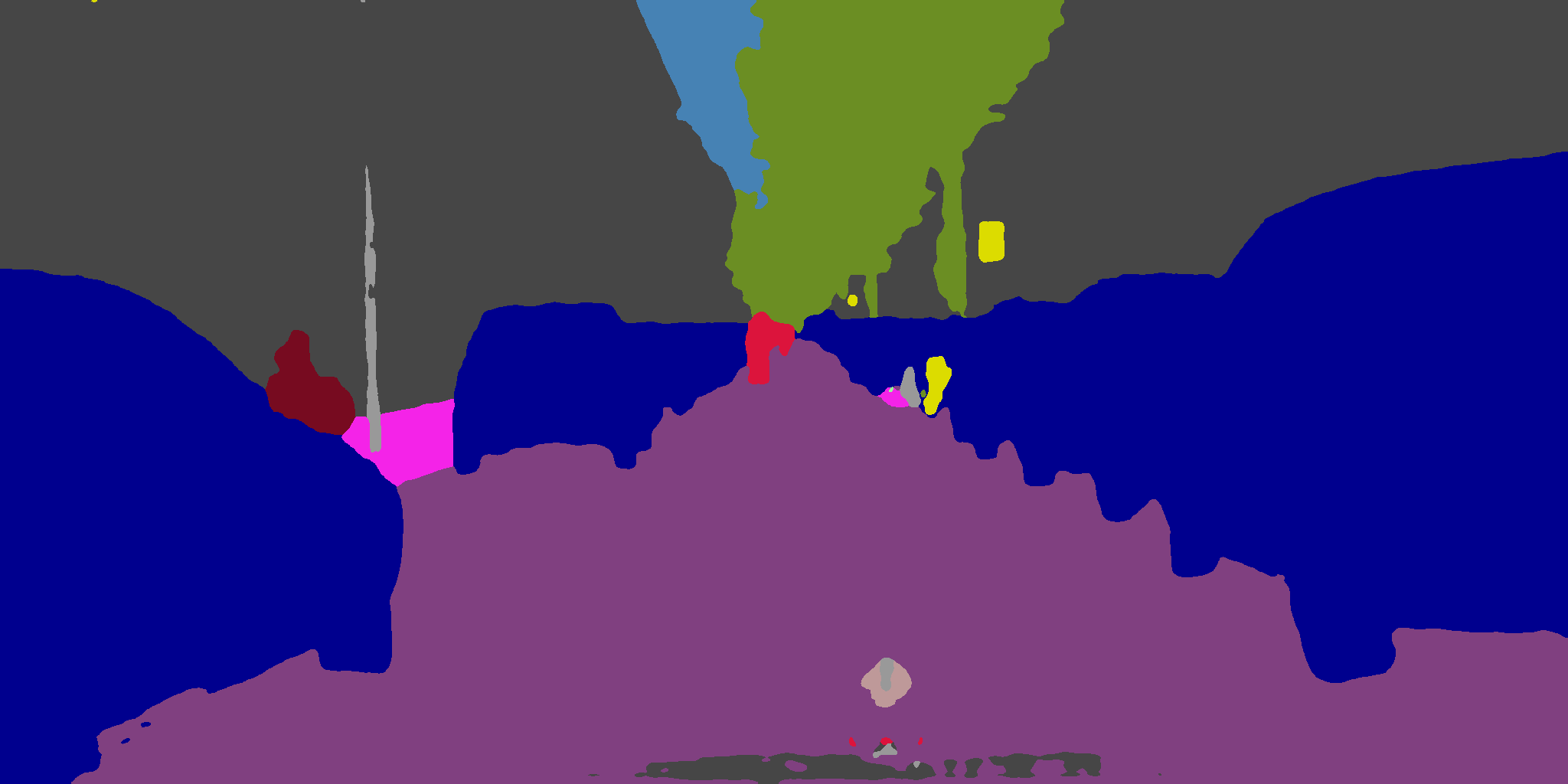}} &
				\subfloat{\includegraphics[width = 0.19\linewidth]{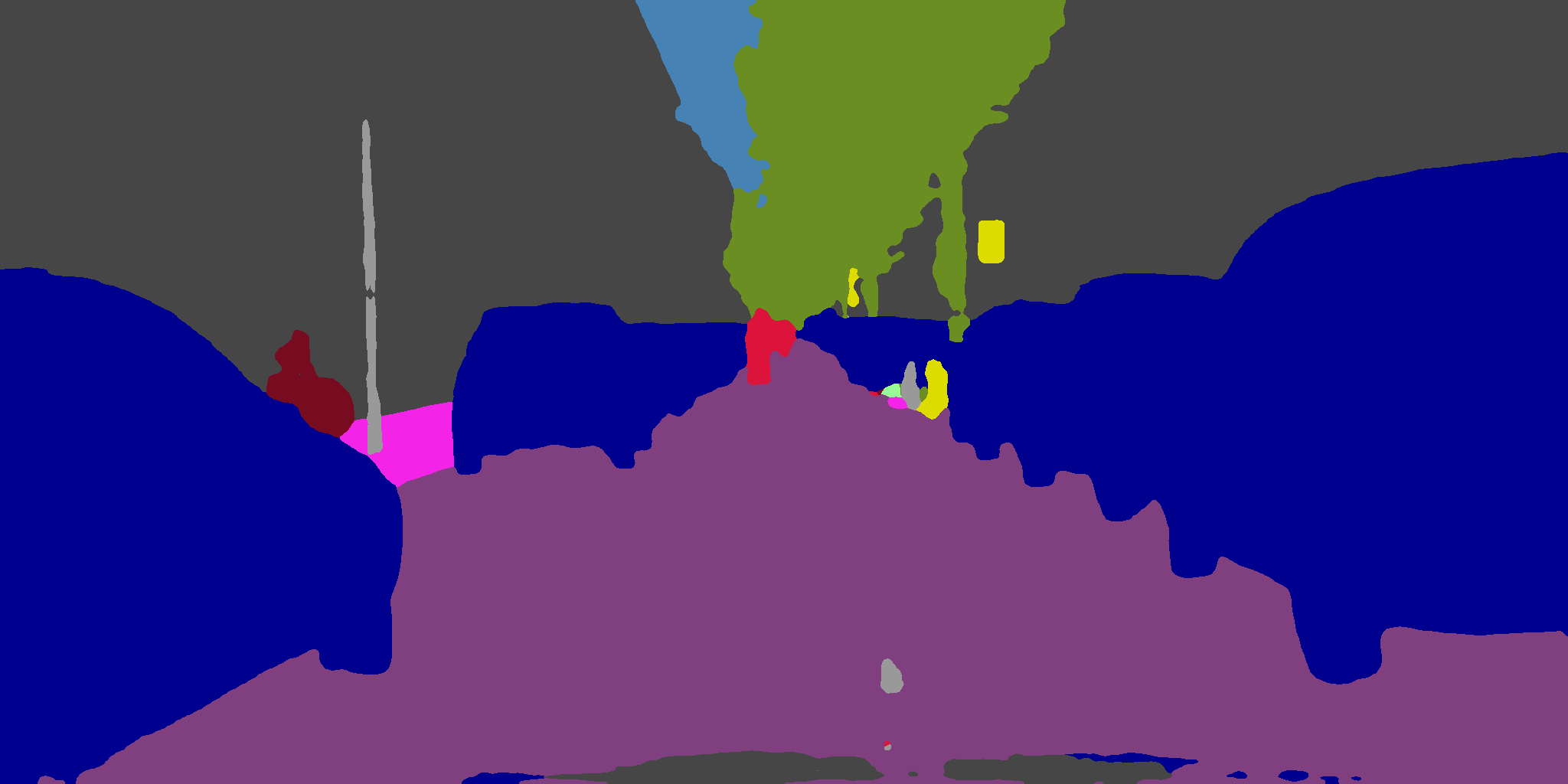}}\\[-0.15in]
				\subfloat{\includegraphics[width = 0.19\linewidth]{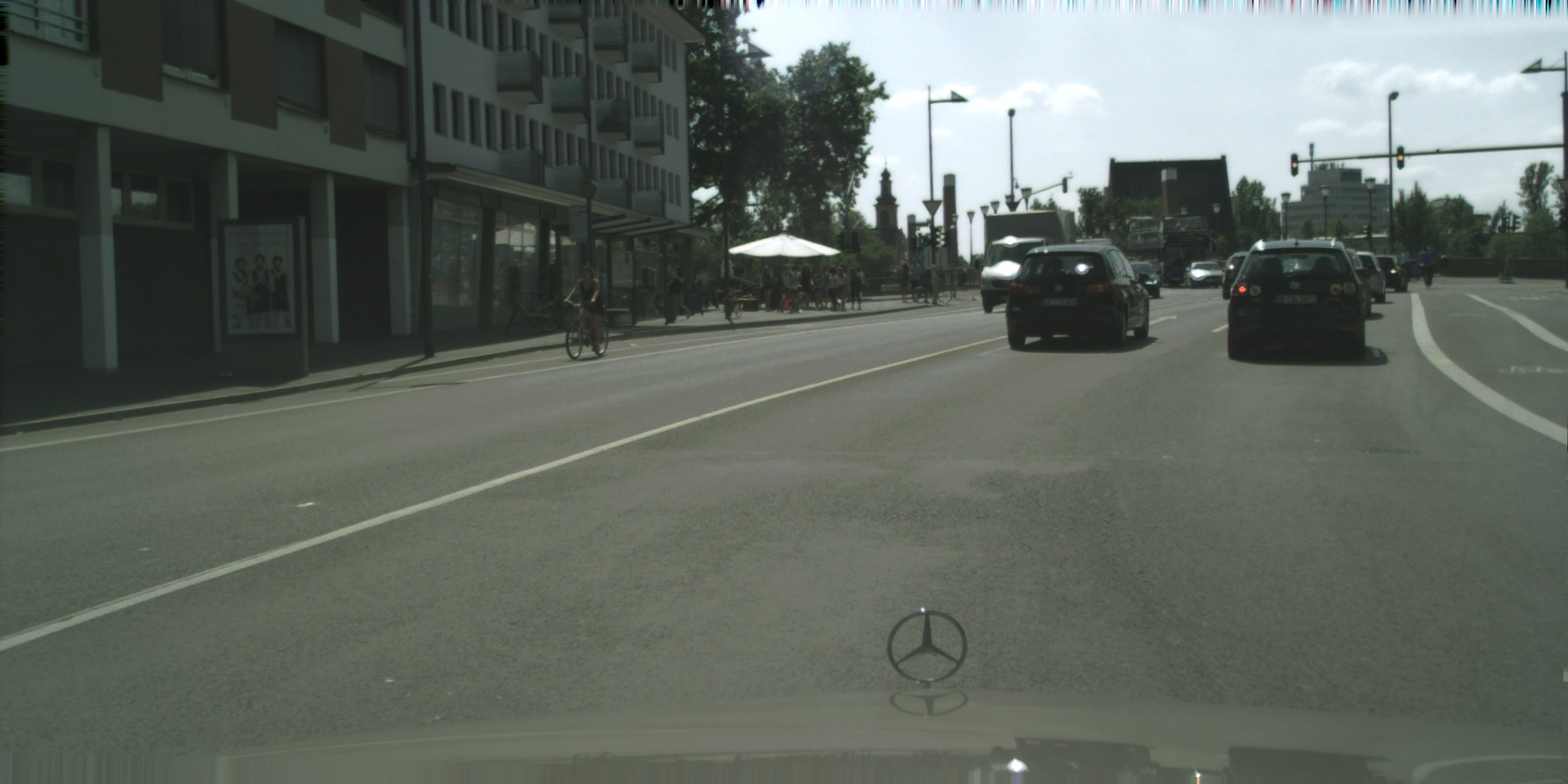}} &
				\subfloat{\includegraphics[width = 0.19\linewidth]{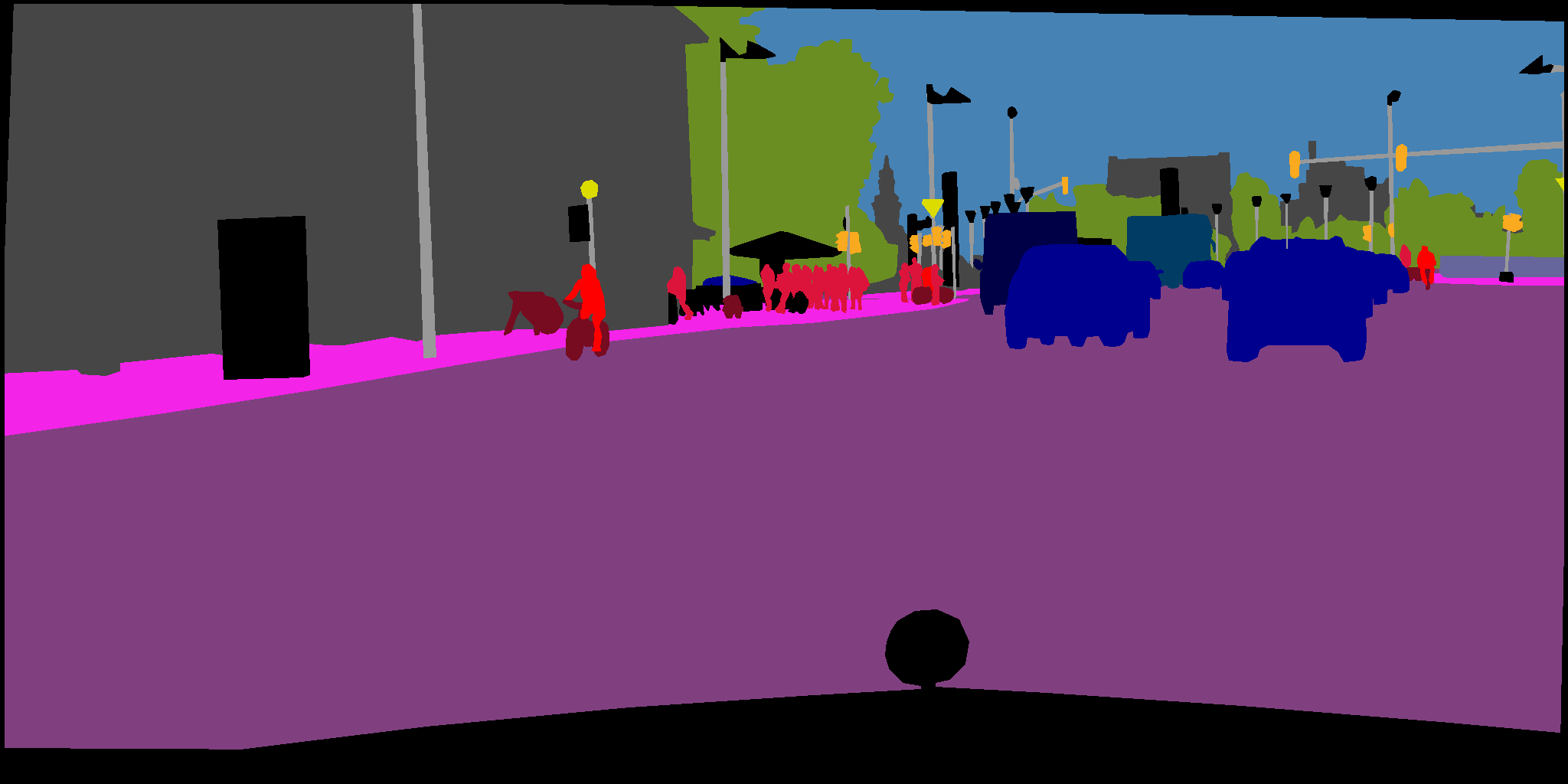}} &
				\subfloat{\includegraphics[width = 0.19\linewidth]{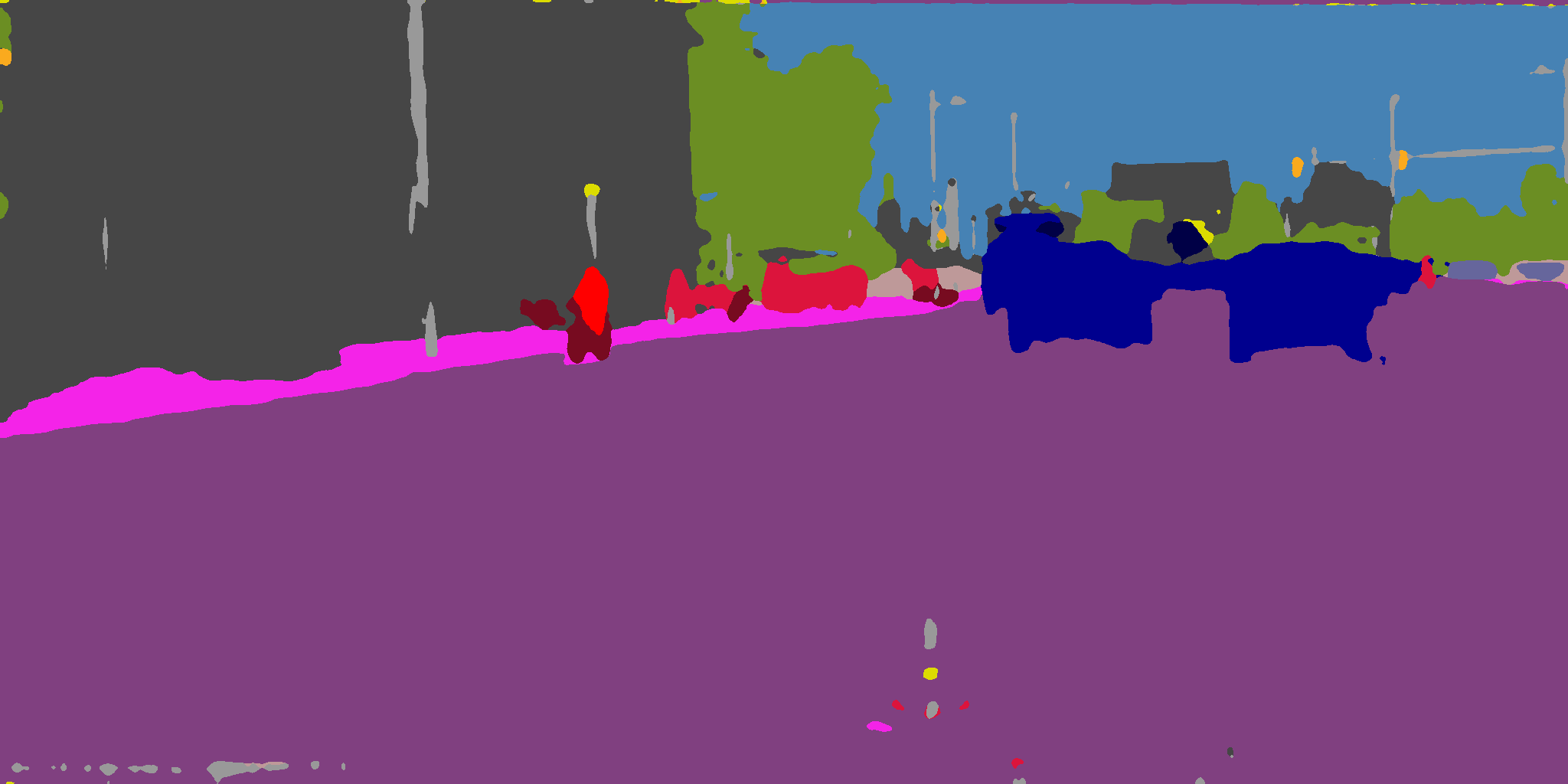}} &
				\subfloat{\includegraphics[width = 0.19\linewidth]{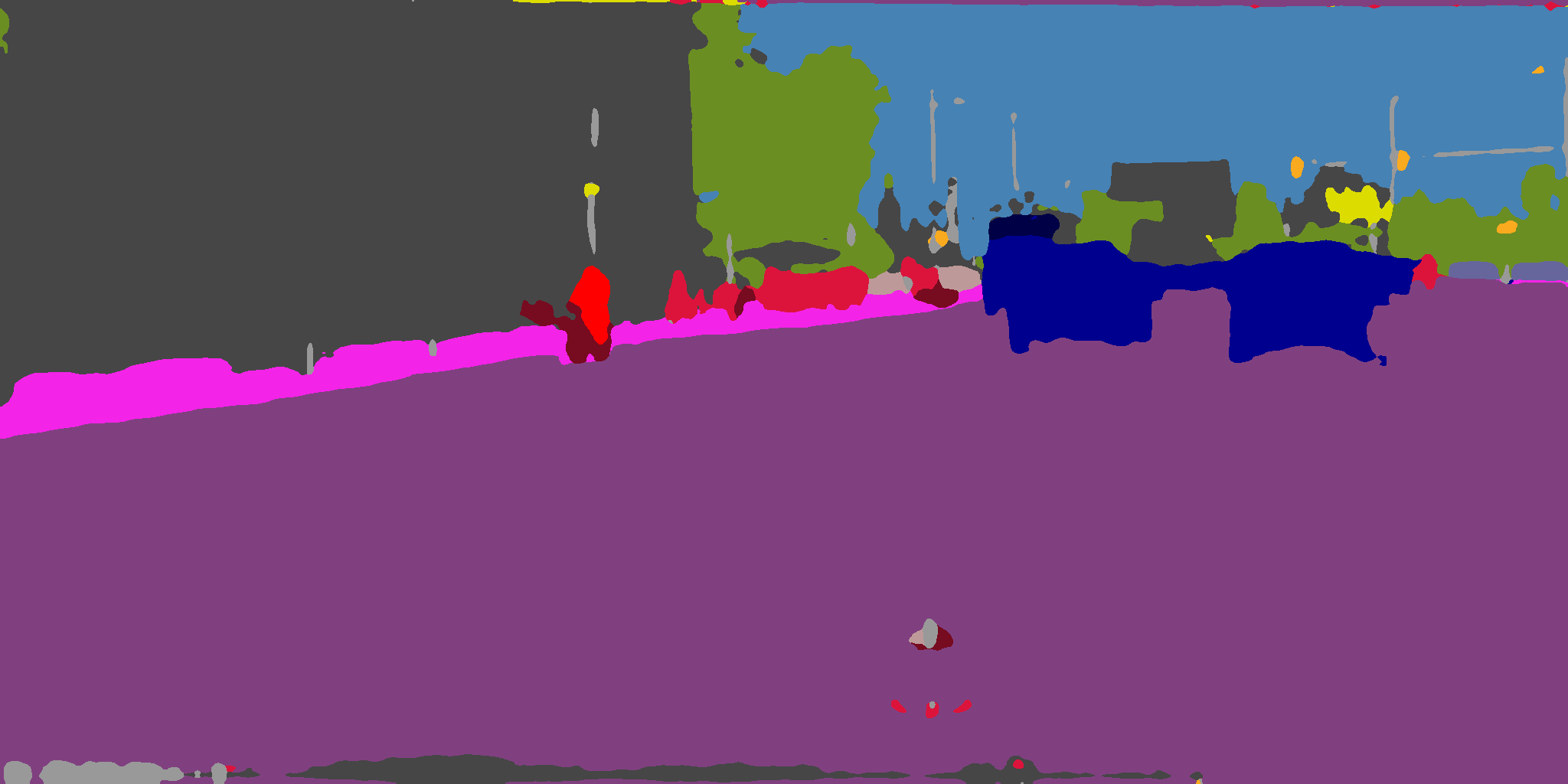}} &
				\subfloat{\includegraphics[width = 0.19\linewidth]{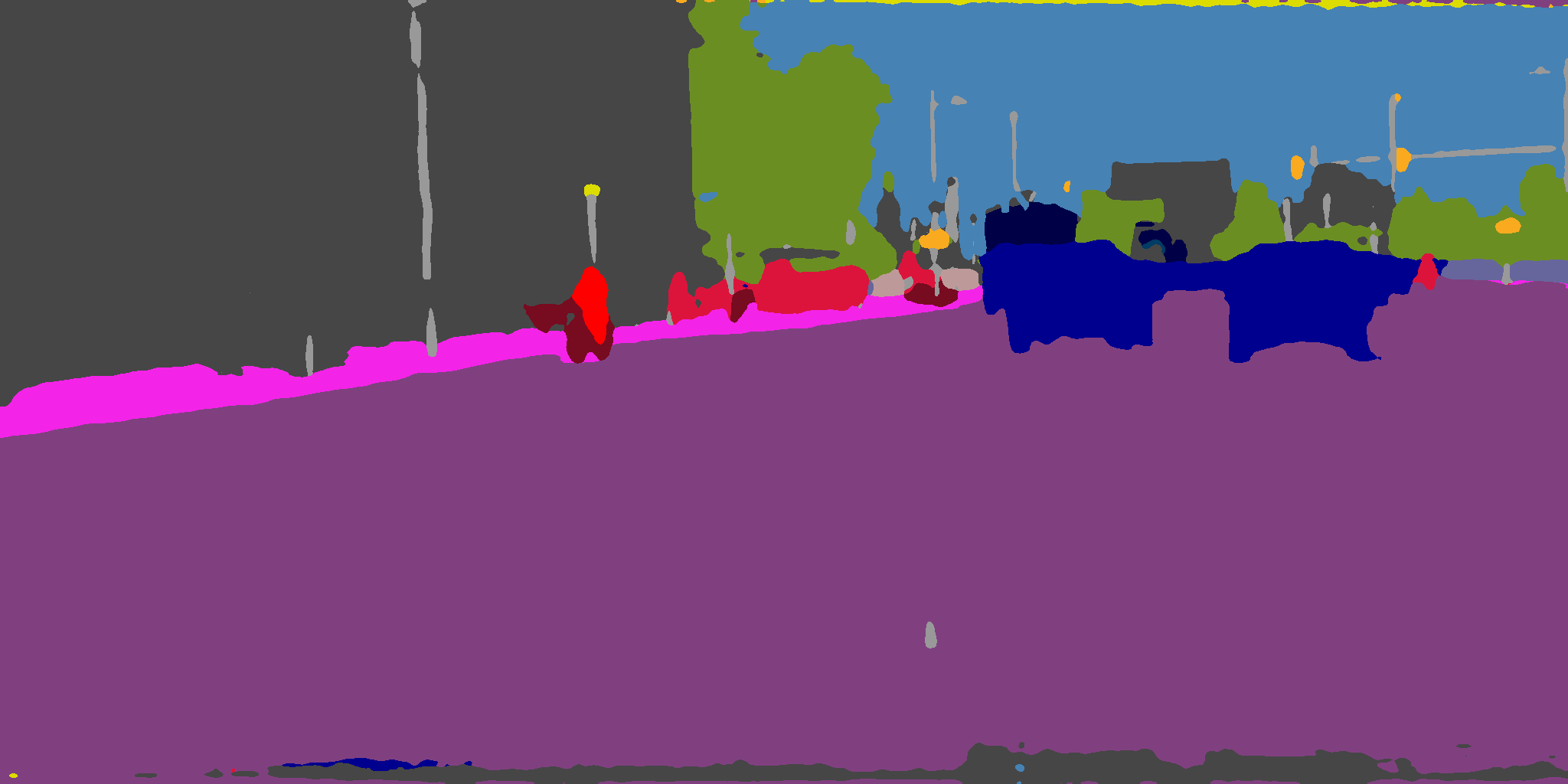}}\\
				Image&GT&RF-50-LW&RF-101-LW&RF-152-LW
			\end{tabular}}
			\vskip 0.1in
			\caption{Visual results on validation set of CityScapes with residual models.}
			\label{fig:cs}
		\end{figure}

Visual results are presented on Figure~\ref{fig:cs}.
\begin{table}[hbt]
	\vskip -0.1in
	\begin{center}
		\begin{tabular}{l|c}
			\hline
			Model & mIoU,\% \\
			\hline
			DeepLab-v2-CRF~\cite{ChenPK0Y16} & $45.7$~(\emph{msc})\\
			RefineNet-101~\cite{LinMSR17}  & $47.1$~(\emph{msc})\\
			RefineNet-152~\cite{LinMSR17} & $47.3$~(\emph{msc})\\
			\hline
			\textbf{RefineNet-LW-101} (ours) & $45.1$\\
			\textbf{RefineNet-LW-152} (ours)  & $45.8$\\
			\hline
		\end{tabular}
	\end{center}
	\caption{Quantitative results on the test set of PASCAL Context. Multi-scale evaluation is defined as \emph{msc}.\label{table:context}}
	\vskip -0.15in
\end{table} 

\begin{figure}[hbt]
			\centering
			\resizebox{\textwidth}{!}{\begin{tabular}{cc|ccc}
					\subfloat{\includegraphics[width = 0.19\linewidth]{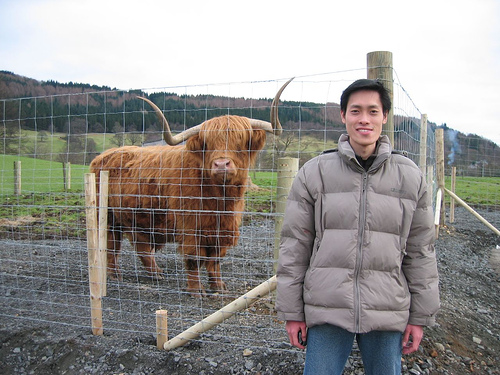}} &
					\subfloat{\includegraphics[width = 0.19\linewidth]{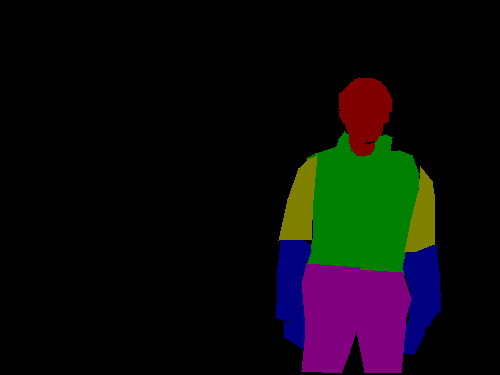}} &
					\subfloat{\includegraphics[width = 0.19\linewidth]{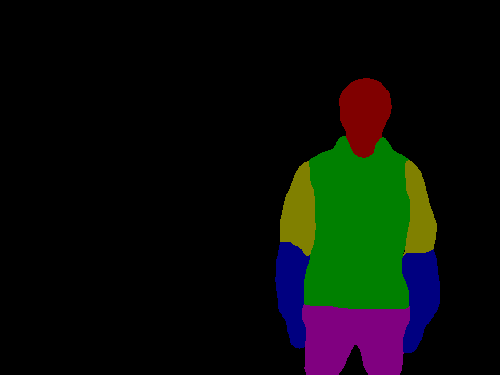}} &
					\subfloat{\includegraphics[width = 0.19\linewidth]{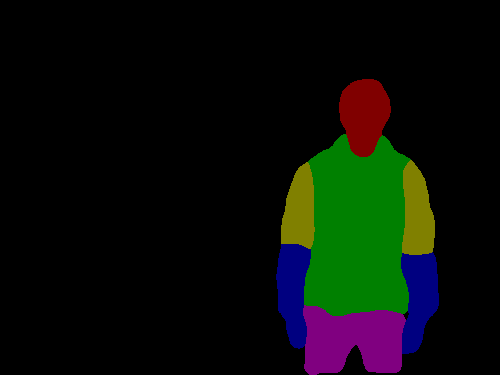}} &
					\subfloat{\includegraphics[width = 0.19\linewidth]{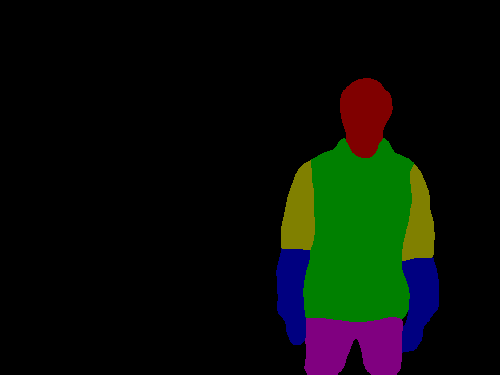}}\\[-0.15in]
					\subfloat{\includegraphics[width = 0.19\linewidth]{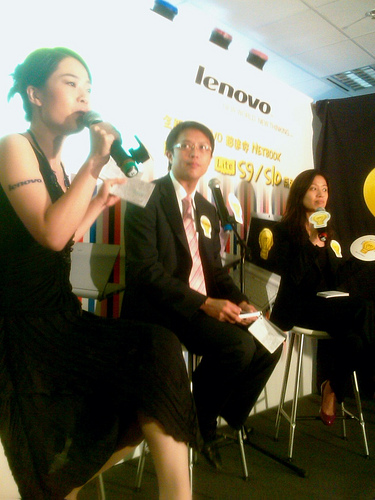}} &
					\subfloat{\includegraphics[width = 0.19\linewidth]{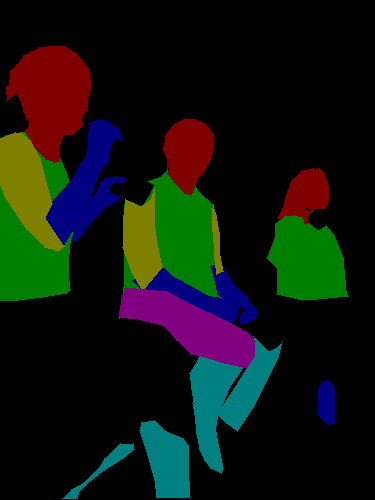}} &
					\subfloat{\includegraphics[width = 0.19\linewidth]{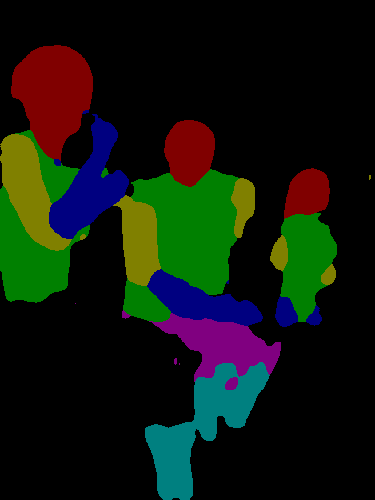}} &
					\subfloat{\includegraphics[width = 0.19\linewidth]{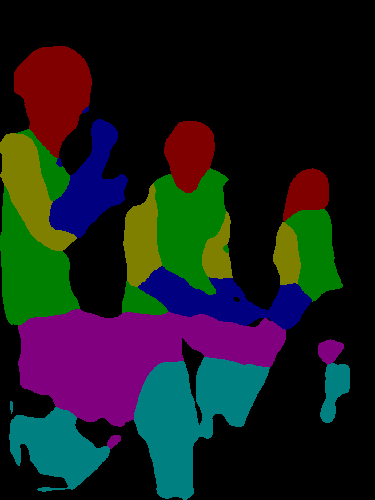}} &
					\subfloat{\includegraphics[width = 0.19\linewidth]{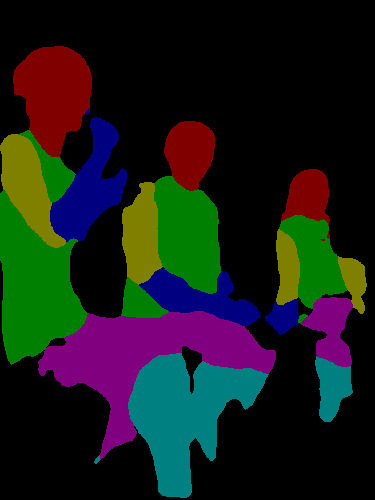}}\\[-0.15in]
					\subfloat{\includegraphics[width = 0.19\linewidth]{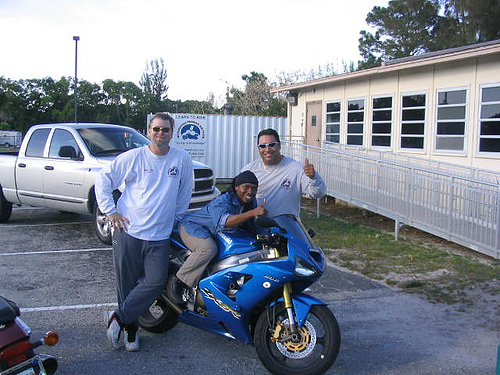}} &
					\subfloat{\includegraphics[width = 0.19\linewidth]{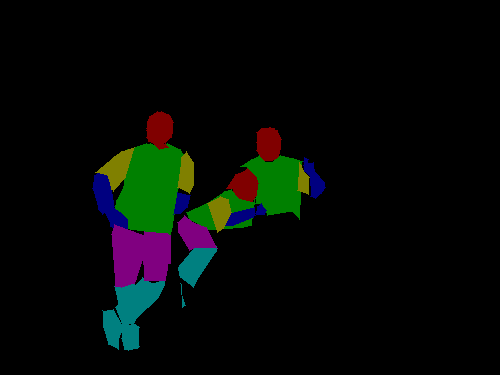}} &
					\subfloat{\includegraphics[width = 0.19\linewidth]{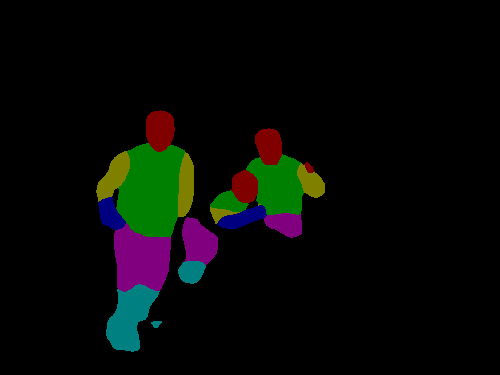}} &
					\subfloat{\includegraphics[width = 0.19\linewidth]{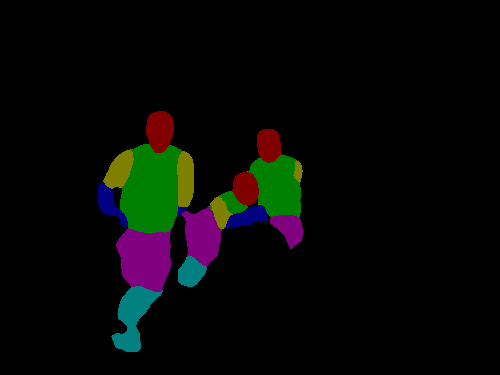}} &
					\subfloat{\includegraphics[width = 0.19\linewidth]{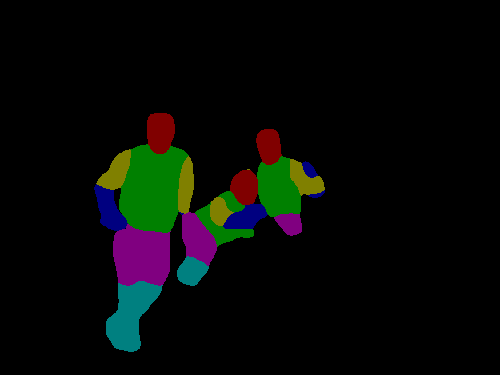}}\\[-0.15in]
					\subfloat{\includegraphics[width = 0.19\linewidth]{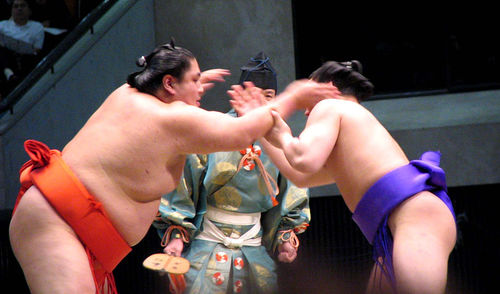}} &
					\subfloat{\includegraphics[width = 0.19\linewidth]{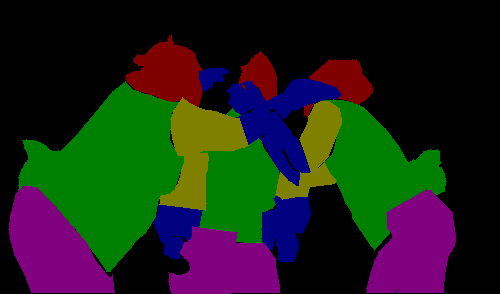}} &
					\subfloat{\includegraphics[width = 0.19\linewidth]{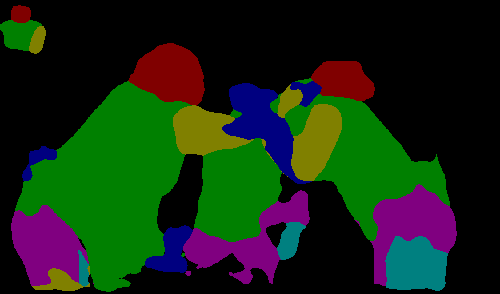}} &
					\subfloat{\includegraphics[width = 0.19\linewidth]{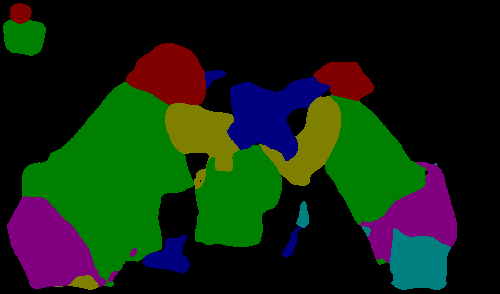}} &
					\subfloat{\includegraphics[width = 0.19\linewidth]{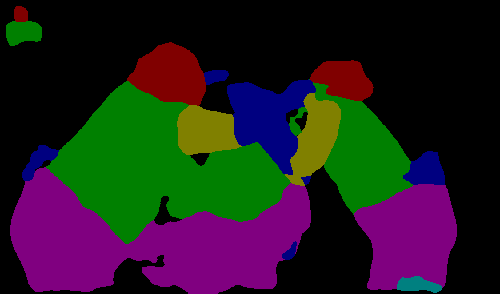}}\\
					Image&GT&RF-50-LW&RF-101-LW&RF-152-LW
				\end{tabular}}
				\vskip 0.1in
				\caption{Visual results on validation set of PASCAL Person-Part with residual models.}
				\label{fig:person}
			\end{figure}

\begin{figure}[hbt]
				\centering
				\resizebox{\textwidth}{!}{\begin{tabular}{cc|ccc}
						\subfloat{\includegraphics[width = 0.19\linewidth]{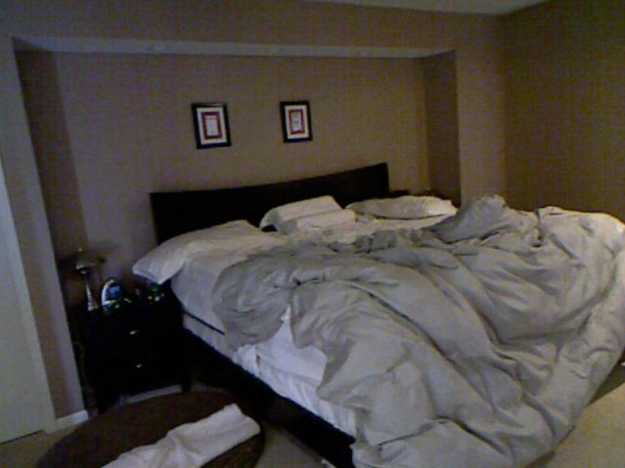}} &
						\subfloat{\includegraphics[width = 0.19\linewidth]{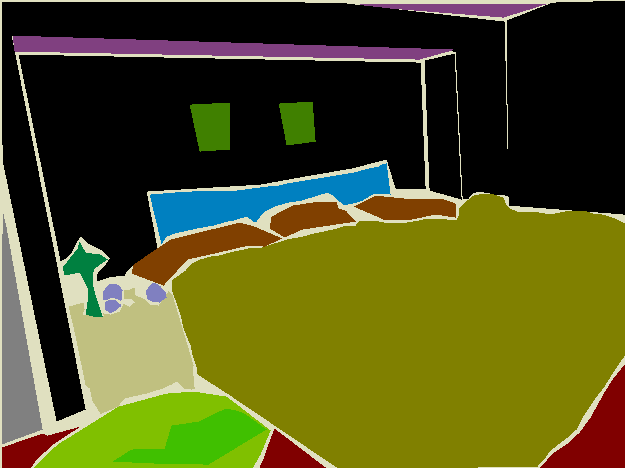}} &
						\subfloat{\includegraphics[width = 0.19\linewidth]{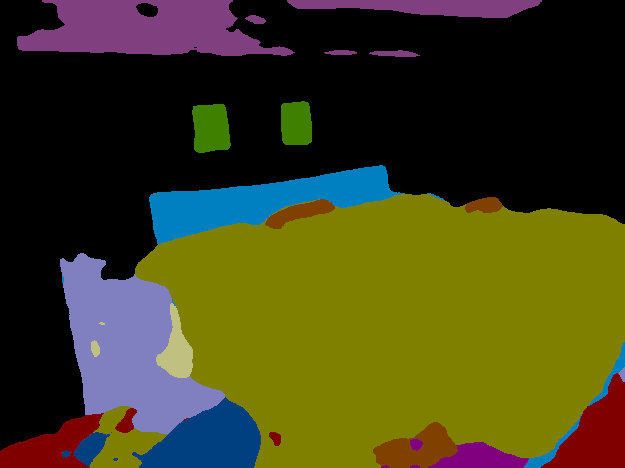}} &
						\subfloat{\includegraphics[width = 0.19\linewidth]{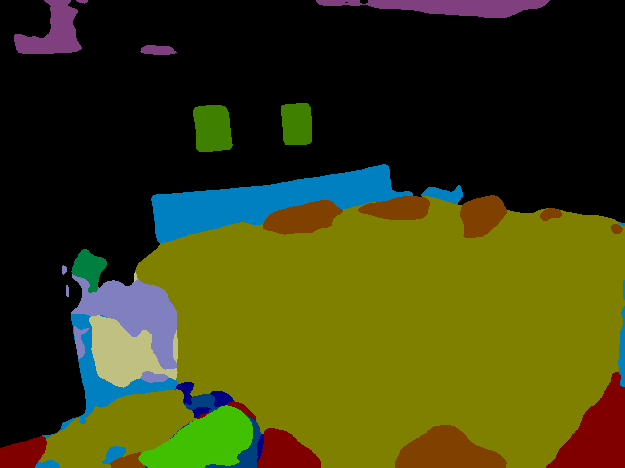}} &
						\subfloat{\includegraphics[width = 0.19\linewidth]{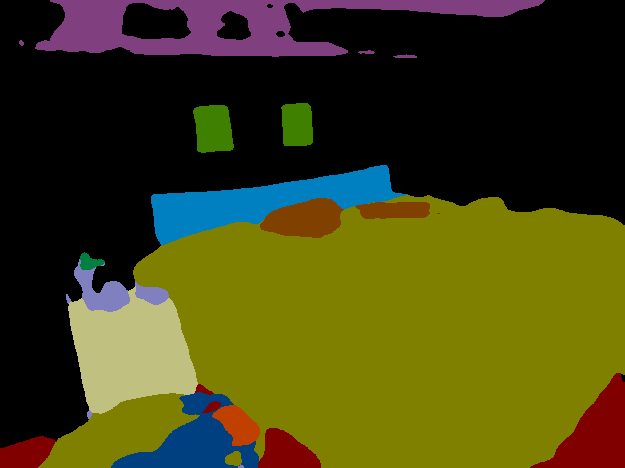}}\\[-0.15in]
						\subfloat{\includegraphics[width = 0.19\linewidth]{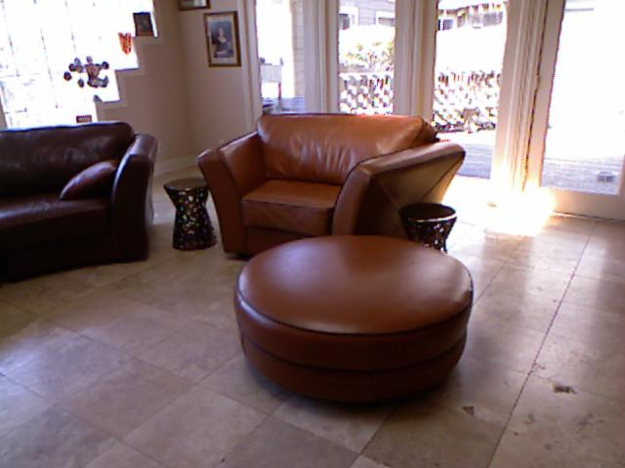}} &
						\subfloat{\includegraphics[width = 0.19\linewidth]{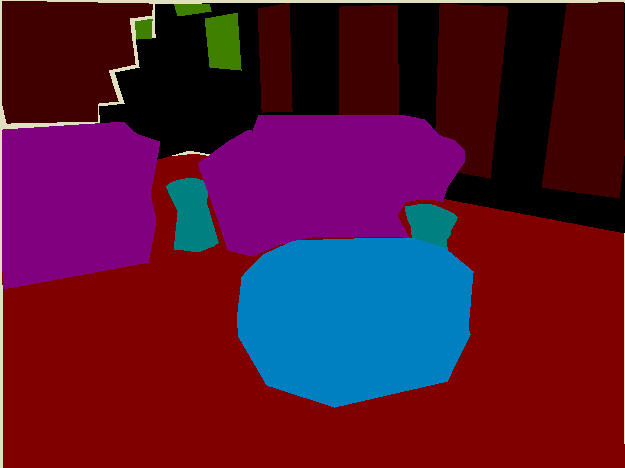}} &
						\subfloat{\includegraphics[width = 0.19\linewidth]{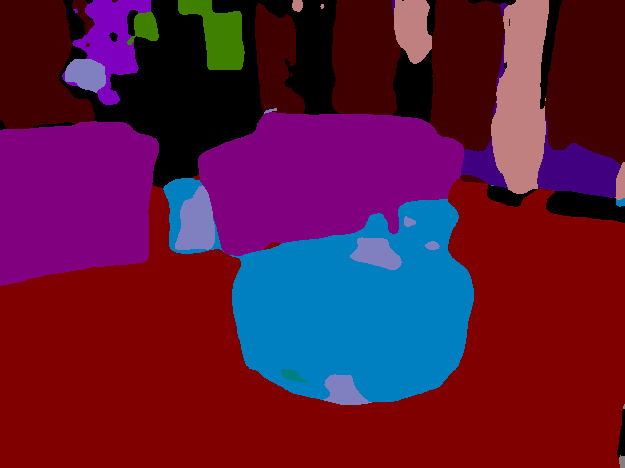}} &
						\subfloat{\includegraphics[width = 0.19\linewidth]{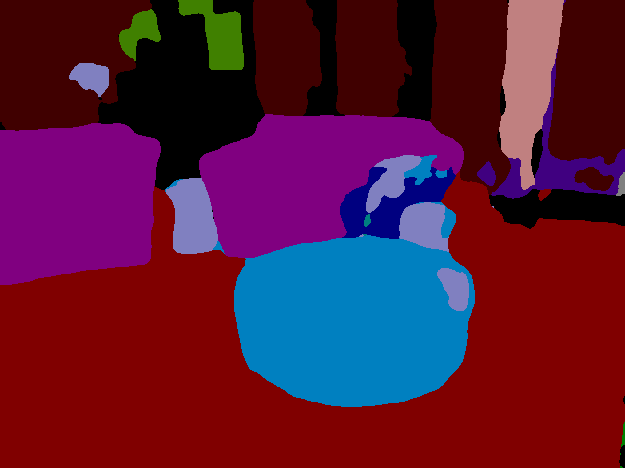}} &
						\subfloat{\includegraphics[width = 0.19\linewidth]{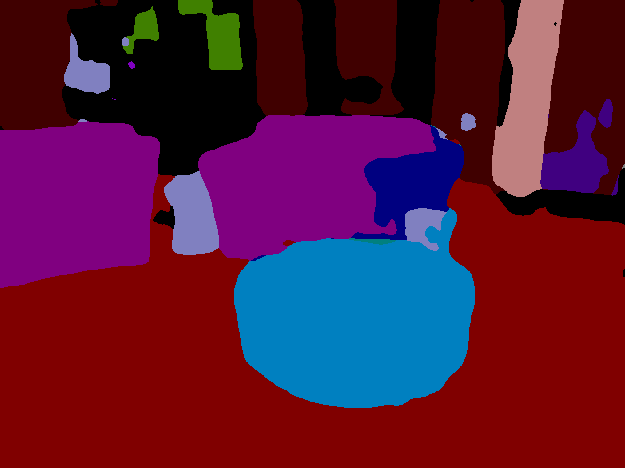}}\\[-0.15in]
						\subfloat{\includegraphics[width = 0.19\linewidth]{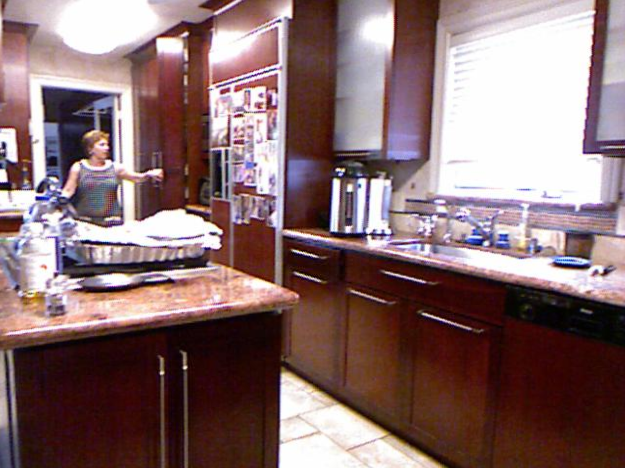}} &
						\subfloat{\includegraphics[width = 0.19\linewidth]{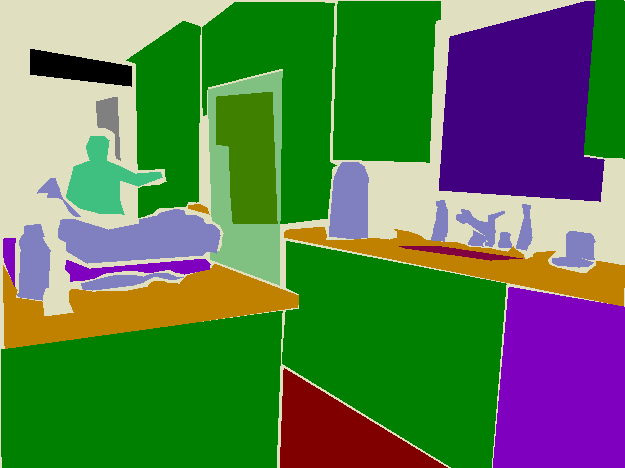}} &
						\subfloat{\includegraphics[width = 0.19\linewidth]{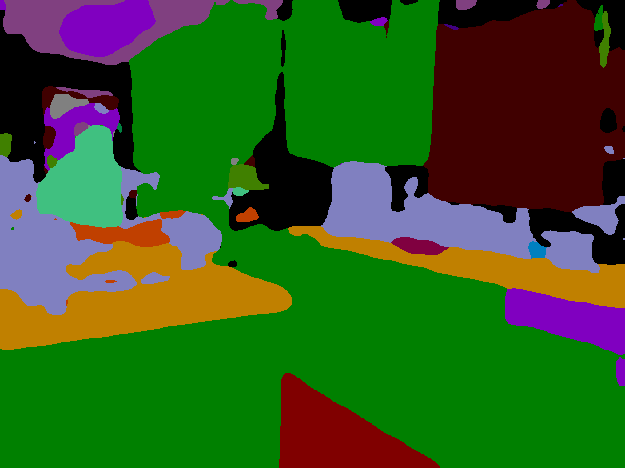}} &
						\subfloat{\includegraphics[width = 0.19\linewidth]{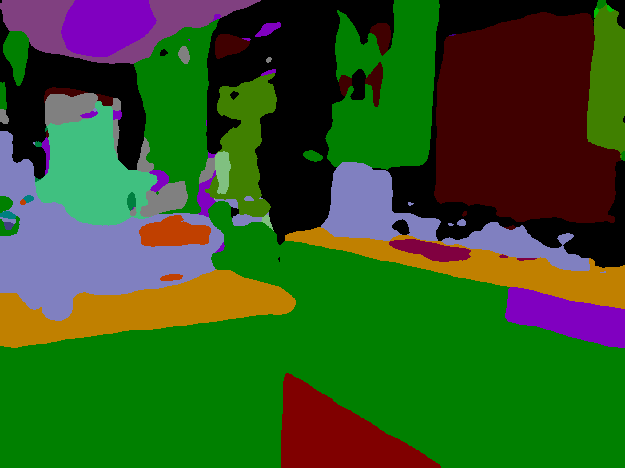}} &
						\subfloat{\includegraphics[width = 0.19\linewidth]{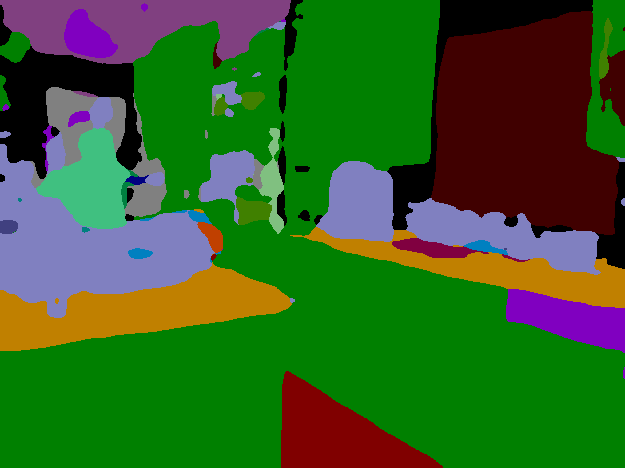}}\\[-0.15in]
						\subfloat{\includegraphics[width = 0.19\linewidth]{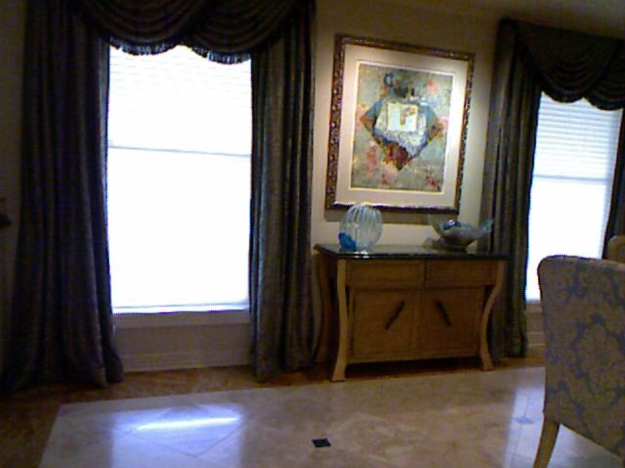}} &
						\subfloat{\includegraphics[width = 0.19\linewidth]{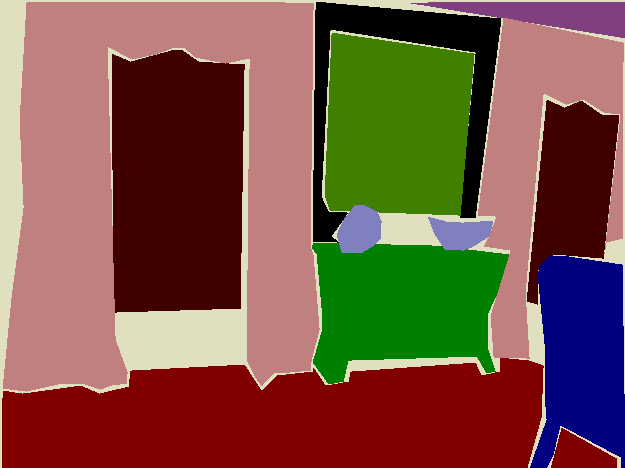}} &
						\subfloat{\includegraphics[width = 0.19\linewidth]{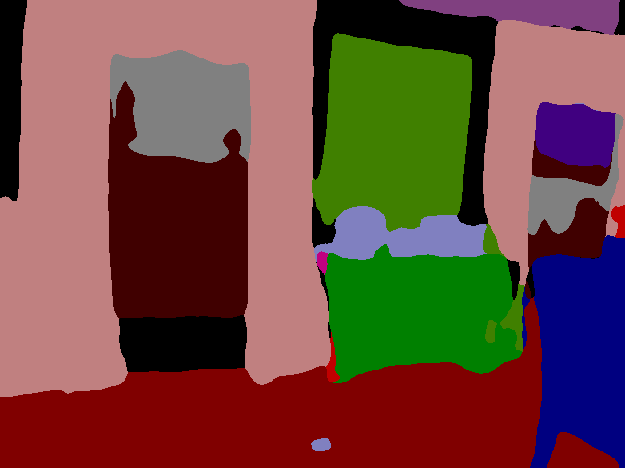}} &
						\subfloat{\includegraphics[width = 0.19\linewidth]{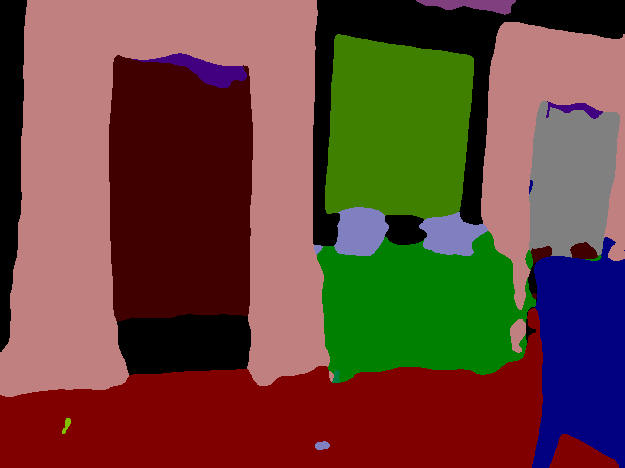}} &
						\subfloat{\includegraphics[width = 0.19\linewidth]{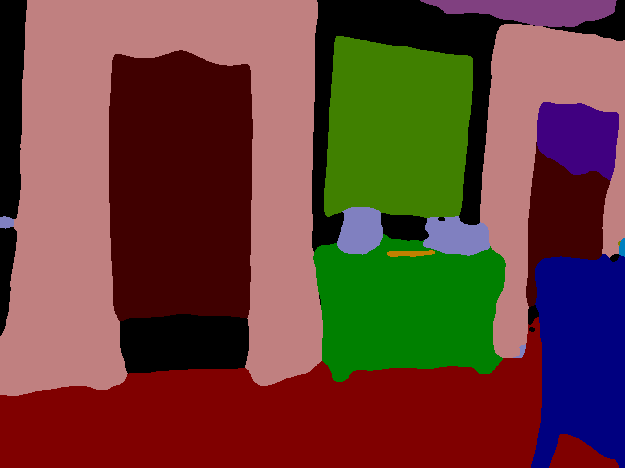}}\\[-0.15in]
						\subfloat{\includegraphics[width = 0.19\linewidth]{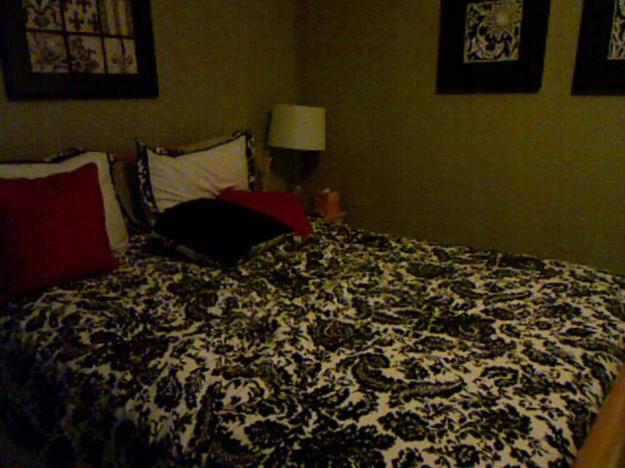}} &
						\subfloat{\includegraphics[width = 0.19\linewidth]{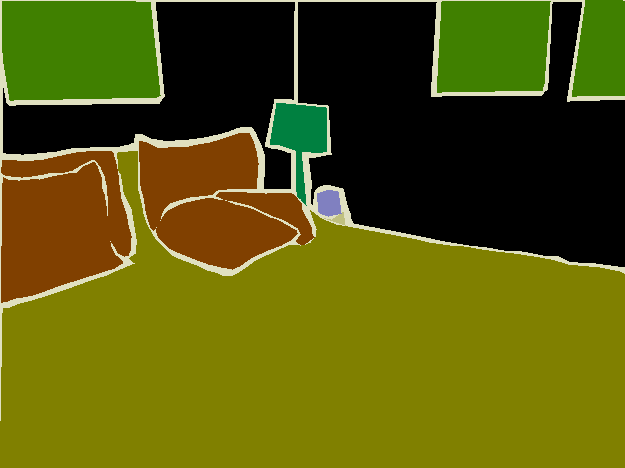}} &
						\subfloat{\includegraphics[width = 0.19\linewidth]{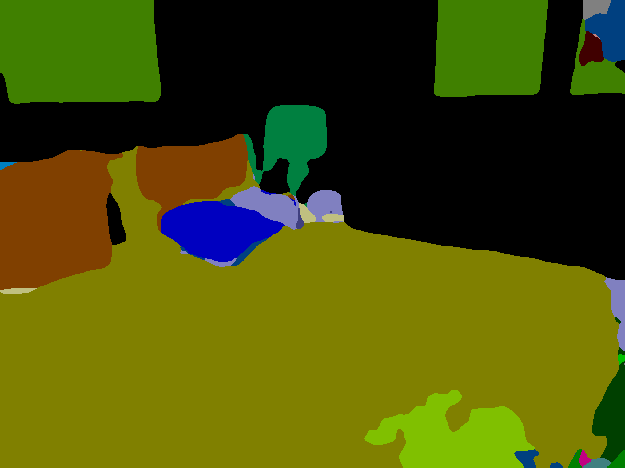}} &
						\subfloat{\includegraphics[width = 0.19\linewidth]{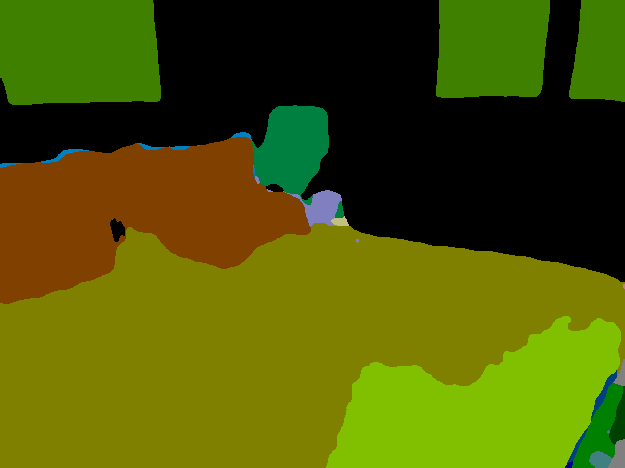}} &
						\subfloat{\includegraphics[width = 0.19\linewidth]{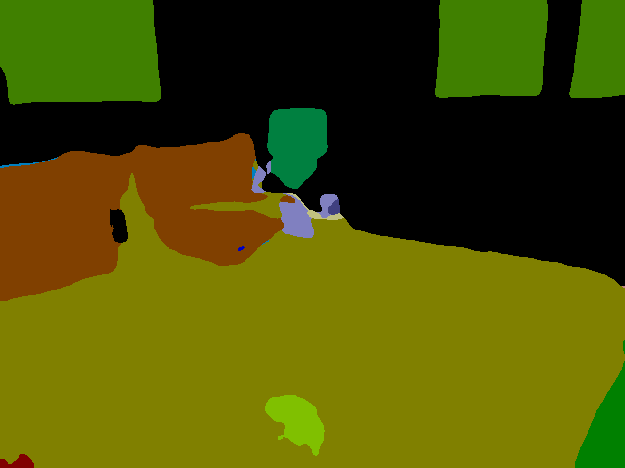}}\\
						Image&GT&RF-50-LW&RF-101-LW&RF-152-LW
					\end{tabular}}
					\vskip 0.1in
					\caption{Visual results on validation set of NYUDv2 with residual models.}
					\label{fig:nyud-res}
				\end{figure}

\noindent \textbf{PASCAL Person-Part.}
Please refer to the main text for quantitative outputs. We provide qualitative results on Figure~\ref{fig:person}.\\
\\
\textbf{NYUD.}
Please refer to the main text for quantitative outputs. We provide qualitative results on Figure~\ref{fig:nyud-res}.\\
\vskip -0.35in

				\begin{figure}[hbt]
					\centering
					\resizebox{\textwidth}{!}{\begin{tabular}{cccc}
							\raisebox{6.\height}{RefineNet-101}&\subfloat{\includegraphics[width = 0.25\linewidth]{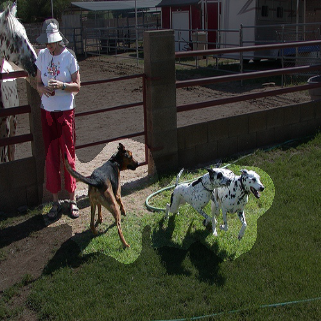}} &
							\subfloat{\includegraphics[width = 0.25\linewidth]{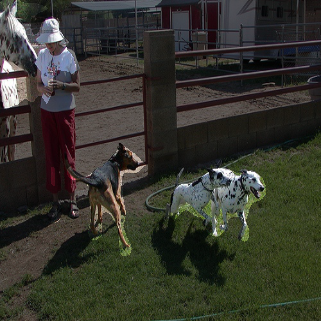}} &
							\subfloat{\includegraphics[width = 0.25\linewidth]{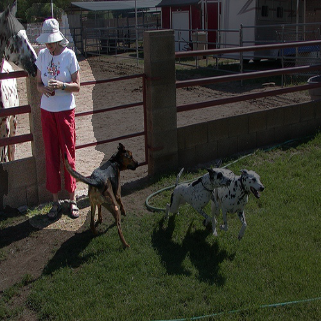}}\\[-0.15in]
							\raisebox{6.\height}{RefineNet-LW-101}&\subfloat{\includegraphics[width = 0.25\linewidth]{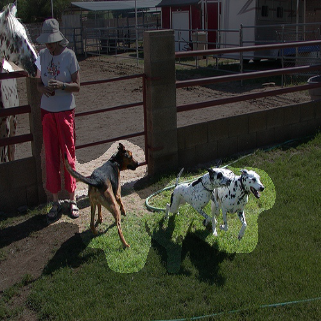}} &
							\subfloat{\includegraphics[width = 0.25\linewidth]{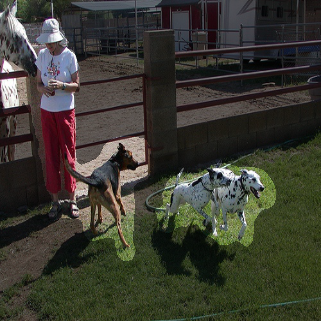}} &
							\subfloat{\includegraphics[width = 0.25\linewidth]{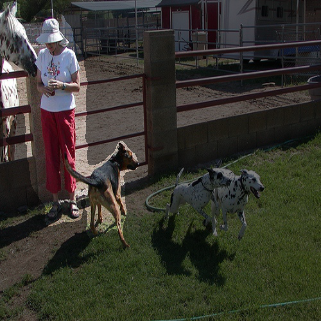}}\\
							&dog&horse&person
						\end{tabular}}
						\vskip 0.1in
						\caption{Comparison of empirical receptive field in the last (classification) layer between RefineNet-101 (top) and RefineNet-LW-101 (bottom). Top activated regions for each unit are shown.}
						\label{fig:erf2}
					\end{figure}
					
					\section{Receptive field size}
					\label{ss:erf}
					To quantify why dropping $3\times3$ convolutions does not result in significant performance drop, we consider the issue of the empirical receptive field (ERF) size~\cite{ZhouKLOT14}. Intuitively, dropping $3\times3$ convolutions should significantly harm the receptive field size of the original architecture. Nevertheless, we note that we do not experience this due to i) the skip-design structure of RefineNet, where low-level features are being summed up with the high-level ones, and ii) keeping CRP blocks.

					Additionally to the results in the main text, we compare ERF of the last (classification) layer between original RefineNet-101 and RefineNet-LW-101, both been pre-trained on PASCAL VOC. From Figure~\ref{fig:erf2}, it can be noticed that both networks exhibit semantically similar activation contours, although the original architecture tends to produce less jagged boundaries.

\end{document}